%% file: main.tex
\documentclass[10pt,twocolumn,letterpaper]{article}

\usepackage{wacv}
\usepackage{times}
\usepackage{epsfig}
\usepackage{graphicx}
\usepackage{amsmath}
\usepackage{amssymb}
\usepackage{booktabs}
\usepackage{color}
\usepackage{xcolor}
\usepackage{tabularx}
\usepackage{multirow}
\usepackage{longtable}
\usepackage{arydshln} 
\usepackage{bm}
\usepackage{algorithm}
\usepackage{algorithmic}
\usepackage{comment}
\usepackage[font={small}]{caption}
\def\correspondingauthor{\footnote{Corresponding author.}}

\usepackage[export]{adjustbox}
\usepackage{lipsum}

\usepackage[accsupp]{axessibility} 
\usepackage[normalem]{ulem}

\DeclareMathOperator*{\argmax}{arg\,max}

\usepackage{pifont}

\newcommand{\cmark}{\ding{51}}%
\newcommand{\xmark}{\ding{55}}%

\def\zhiwu{\textcolor{blue}}

\def\wacvPaperID{306} 

\wacvalgorithmstrack   

\wacvfinalcopy 

\ifwacvfinal
\usepackage[breaklinks=true,bookmarks=false]{hyperref}
\else
\usepackage[pagebackref=true,breaklinks=true,colorlinks,bookmarks=false]{hyperref}
\fi

\begin{document}

\title{A Continual Deepfake Detection Benchmark: Dataset, Methods, and Essentials}

\author{Chuqiao Li\textsuperscript{\rm 1}, Zhiwu Huang\textsuperscript{\rm 2,}\correspondingauthor , Danda Pani Paudel\textsuperscript{\rm 1}, Yabin Wang\textsuperscript{\rm 3,2}, 
\\ Mohamad Shahbazi\textsuperscript{\rm 1}, Xiaopeng Hong\textsuperscript{\rm 4}, Luc Van Gool\textsuperscript{\rm 1,5} \\
\textsuperscript{\rm 1}ETH Z\"urich, Switzerland \quad
\textsuperscript{\rm 2}Singapore Management University, Singapore \\
\textsuperscript{\rm 3}Xi'an Jiaotong University, China 
\quad
\textsuperscript{\rm 4}Harbin Institute of Technology, China 
\quad
\textsuperscript{\rm 5}KU Leuven, Belgium \\
\texttt{chuqli@student.ethz.ch, zzhiwu.huang@gmail.com}, \\ \texttt{\{paudel, mshahbazi, vangool\}@vision.ee.ethz.ch}, \\ \texttt{iamwangyabin@stu.xjtu.edu.cn, hongxiaopeng@ieee.org}  \\
}



\twocolumn[{%
\renewcommand\twocolumn[1][]{#1}%
\maketitle

\begin{center}

\Large
\resizebox{0.8\linewidth}{!}
{
\begin{tabular}{cccccccccccc}
\colorbox{green}{\includegraphics[width=.24\textwidth,valign=c]{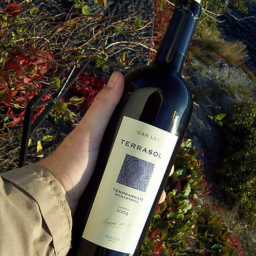}} &
\colorbox{green}{\includegraphics[width=.24\textwidth,valign=c]{figures/cddb/gau/r/000000329827.png}} &
\colorbox{green}{\includegraphics[width=.24\textwidth,valign=c]{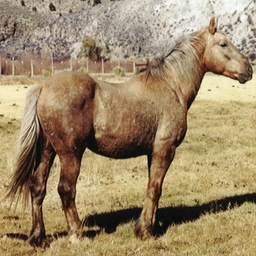}} &
\colorbox{green}{\includegraphics[width=.24\textwidth,valign=c]{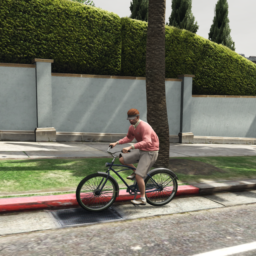}} &
\colorbox{green}{\includegraphics[width=.21\textwidth,valign=c]{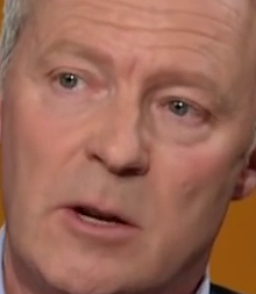}} &
\colorbox{green}{\includegraphics[width=.24\textwidth,valign=c]{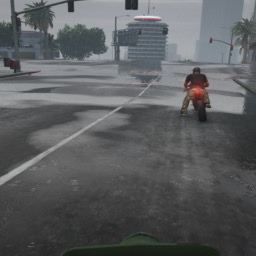}} &
\colorbox{green}{\includegraphics[width=.24\textwidth,valign=c]{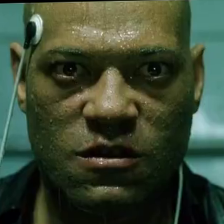}} &
\colorbox{green}{\includegraphics[width=.24\textwidth,valign=c]{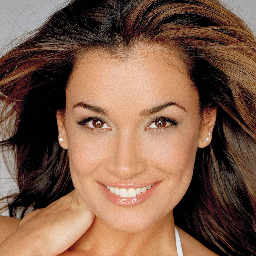}} &
\colorbox{green}{\includegraphics[width=.24\textwidth,valign=c]{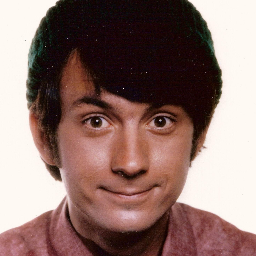}} &
\colorbox{green}{\includegraphics[width=.24\textwidth,valign=c]{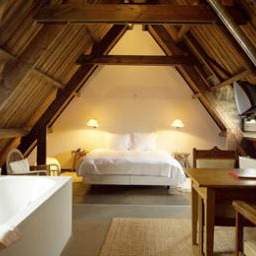}} &
\colorbox{green}{\includegraphics[width=.24\textwidth,valign=c]{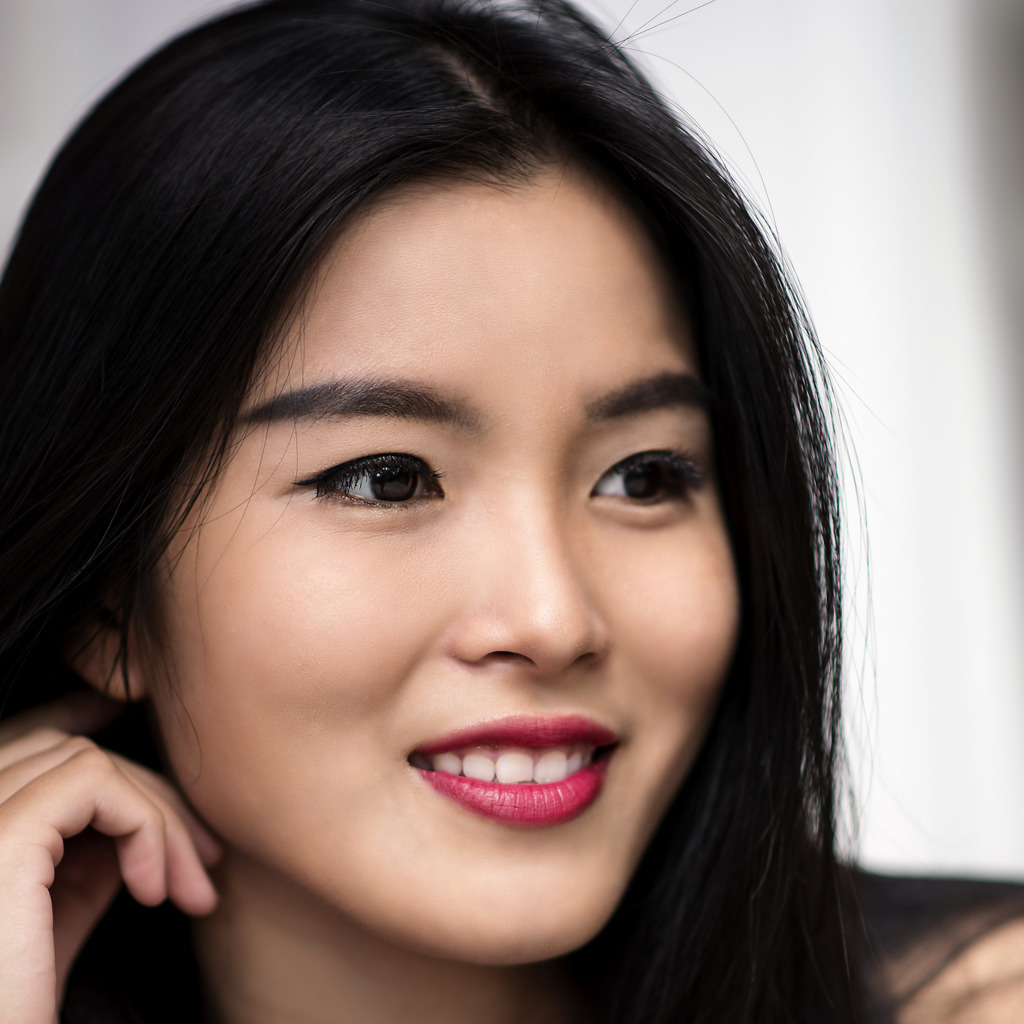}} &
\colorbox{green}{\includegraphics[width=.24\textwidth,valign=c]{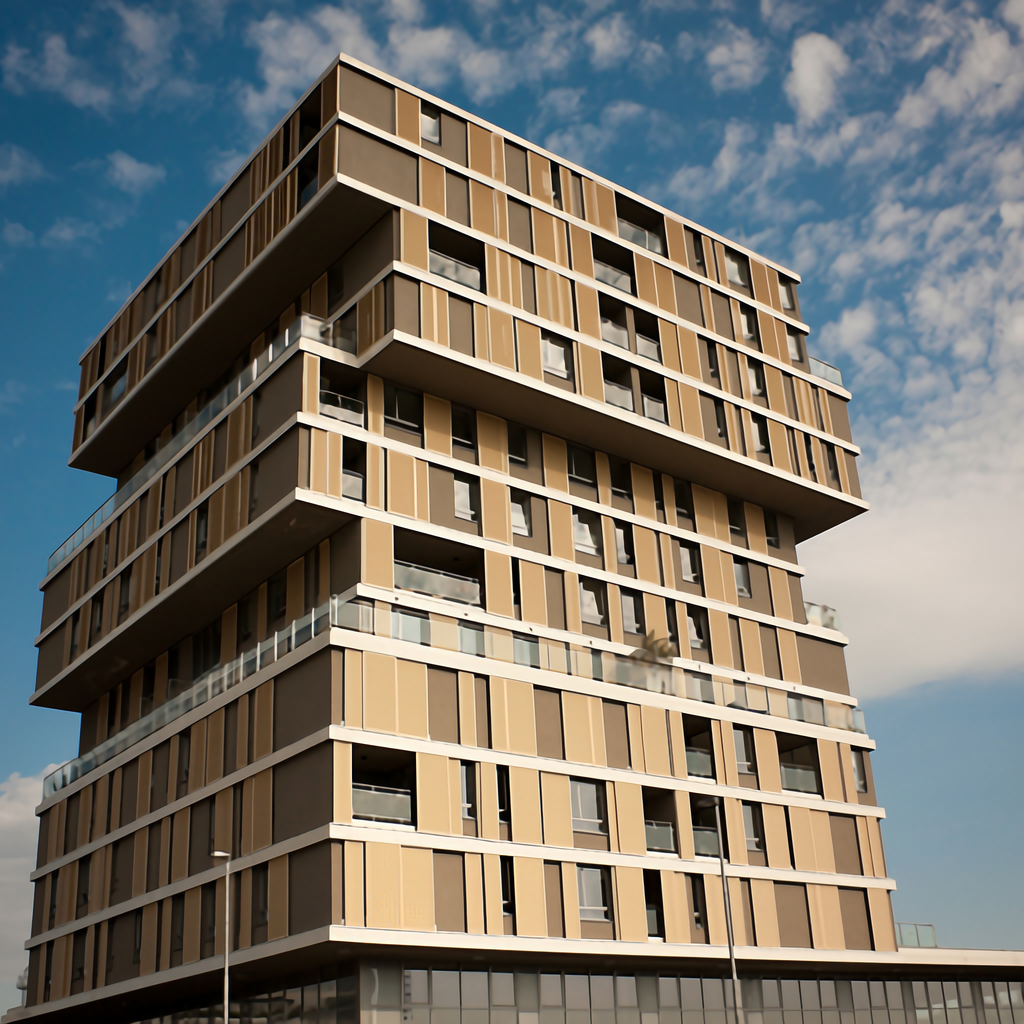}}
\\
\colorbox{red}{\includegraphics[width=.24\textwidth,valign=c]{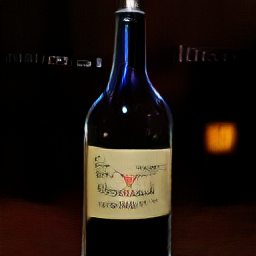}} &
\colorbox{red}{\includegraphics[width=.24\textwidth,valign=c]{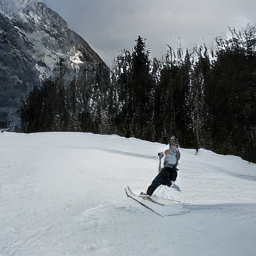}} &
\colorbox{red}{\includegraphics[width=.24\textwidth,valign=c]{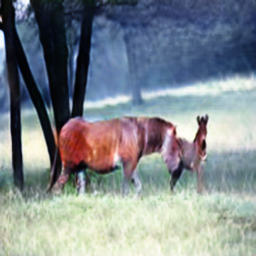}} &
\colorbox{red}{\includegraphics[width=.24\textwidth,valign=c]{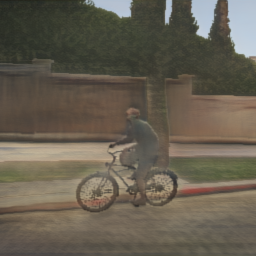}} &
\colorbox{red}{\includegraphics[width=.21\textwidth,valign=c]{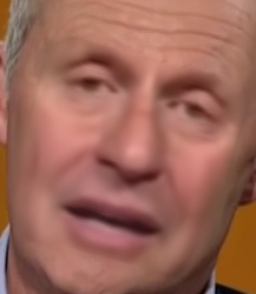}} &
\colorbox{red}{\includegraphics[width=.24\textwidth,valign=c]{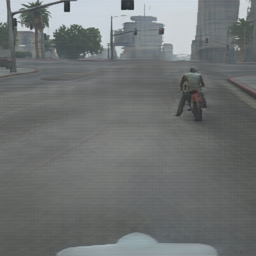}} &
\colorbox{red}{\includegraphics[width=.24\textwidth,valign=c]{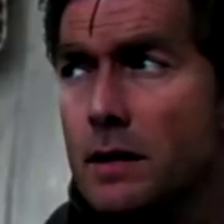}}&
\colorbox{red}{\includegraphics[width=.24\textwidth,valign=c]{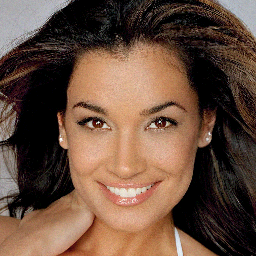}}&
\colorbox{red}{\includegraphics[width=.24\textwidth,valign=c]{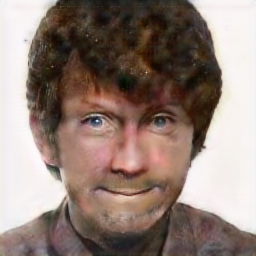}}&
\colorbox{red}{\includegraphics[width=.24\textwidth,valign=c]{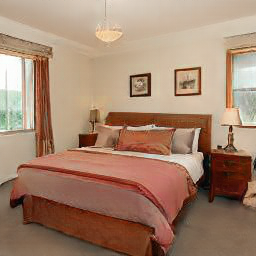}}&
\colorbox{red}{\includegraphics[width=.24\textwidth,valign=c]{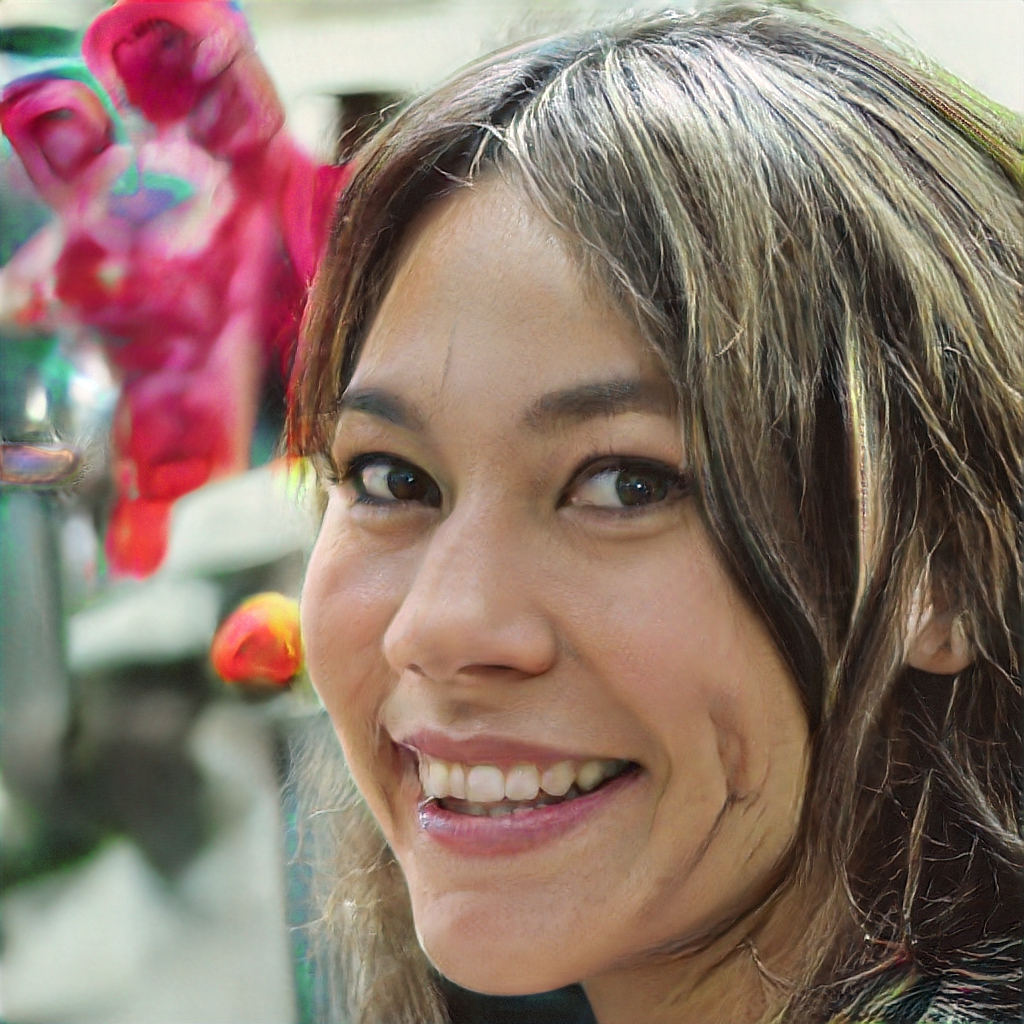}} &
\colorbox{red}{\includegraphics[width=.24\textwidth,valign=c]{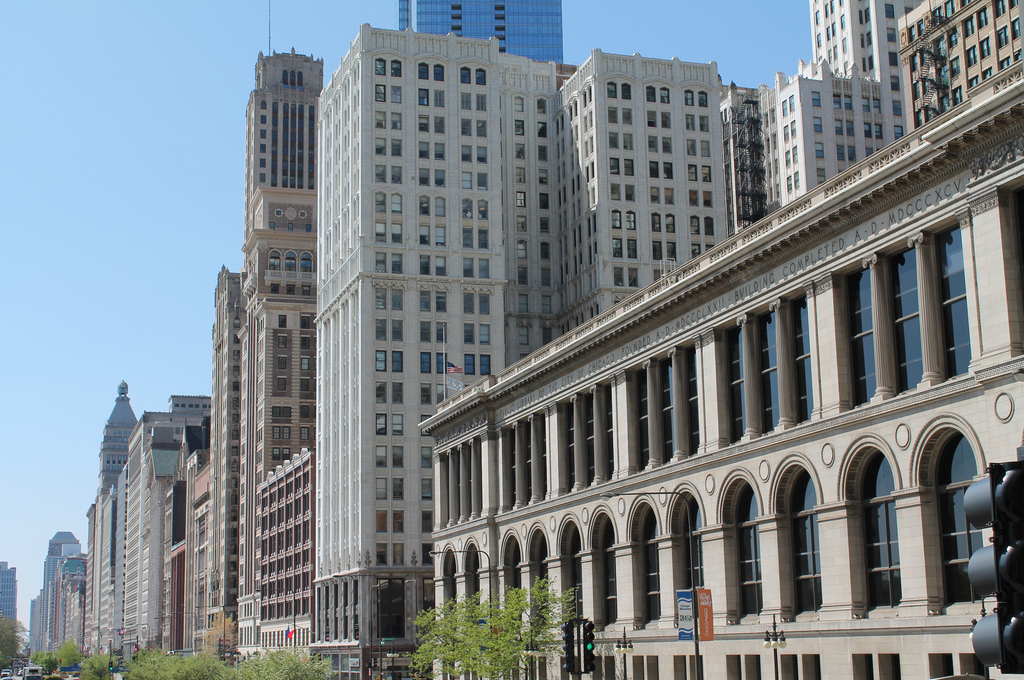}}

\\
\colorbox{green}{\includegraphics[width=.24\textwidth,valign=c]{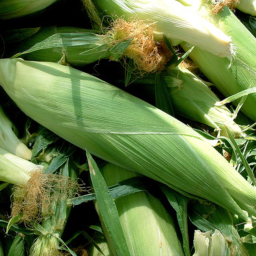}} &
\colorbox{green}{\includegraphics[width=.24\textwidth,valign=c]{figures/cddb/gau/r/000000333745.png}} &
\colorbox{green}{\includegraphics[width=.24\textwidth,valign=c]{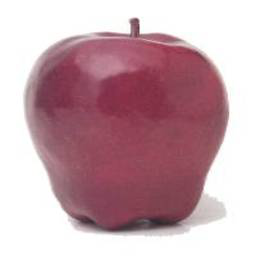}} &
\colorbox{green}{\includegraphics[width=.24\textwidth,valign=c]{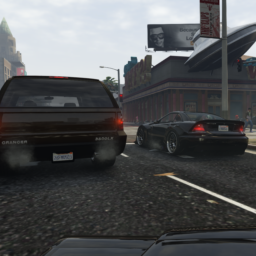}} &
\colorbox{green}{\includegraphics[width=.21\textwidth,valign=c]{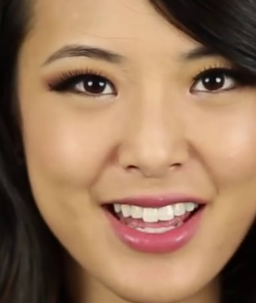}} &
\colorbox{green}{\includegraphics[width=.24\textwidth,valign=c]{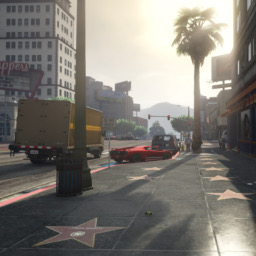}} &
\colorbox{green}{\includegraphics[width=.24\textwidth,valign=c]{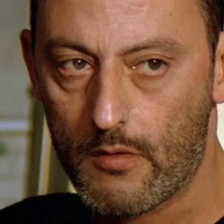}}&
\colorbox{green}{\includegraphics[width=.24\textwidth,valign=c]{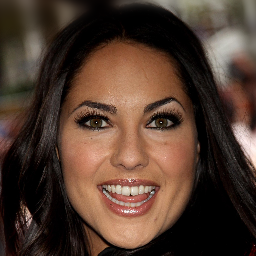}}&
\colorbox{green}{\includegraphics[width=.24\textwidth,valign=c]{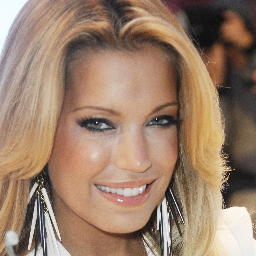}}&
\colorbox{green}{\includegraphics[width=.24\textwidth,valign=c]{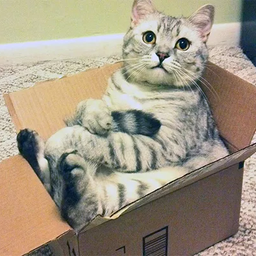}}&
\colorbox{green}{\includegraphics[width=.24\textwidth,valign=c]{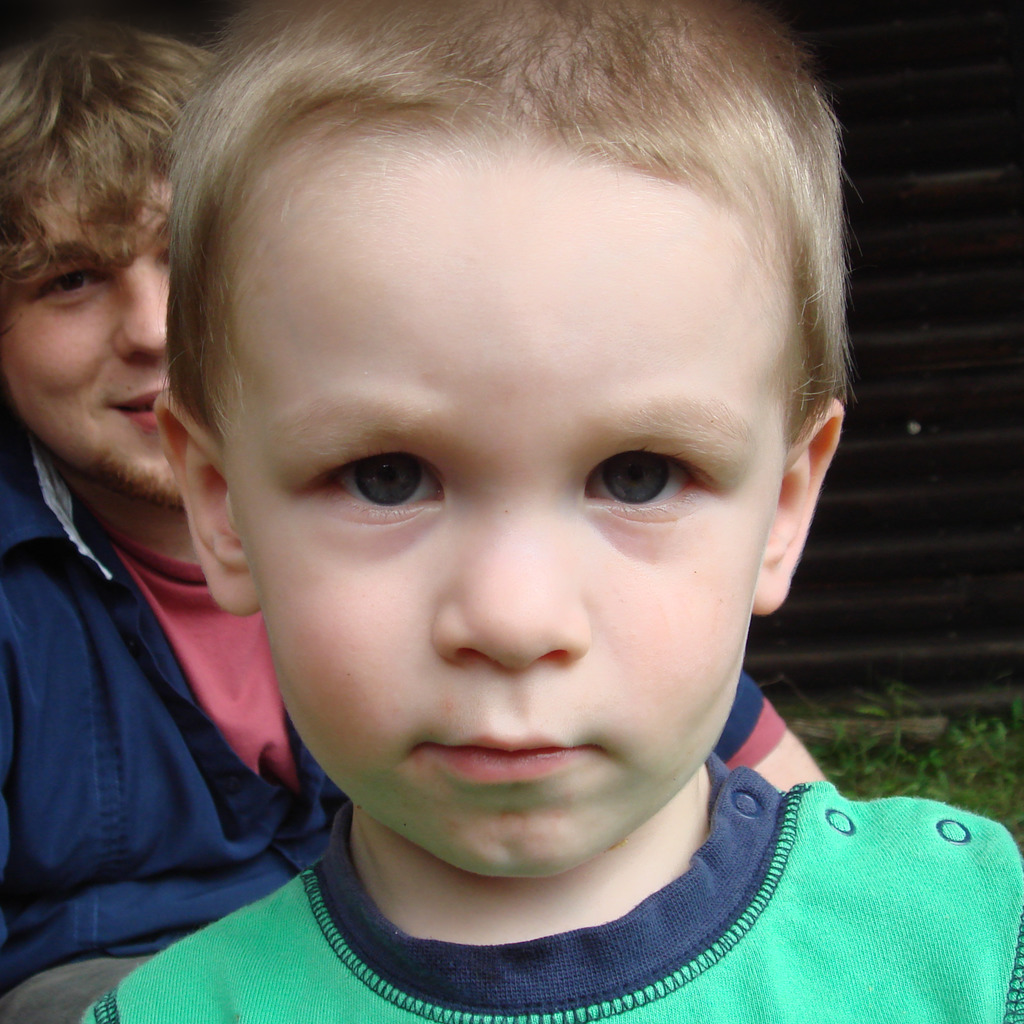}} &
\colorbox{green}{\includegraphics[width=.24\textwidth,valign=c]{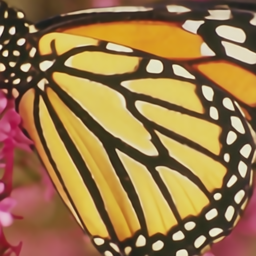}}
\\
\colorbox{red}{\includegraphics[width=.24\textwidth,valign=c]{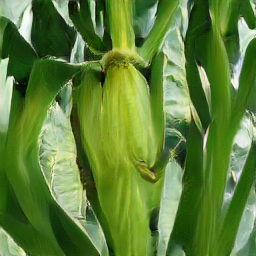}} &
\colorbox{red}{\includegraphics[width=.24\textwidth,valign=c]{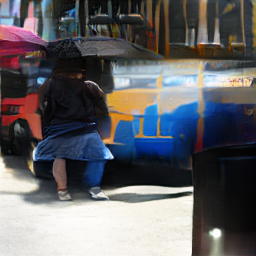}} &
\colorbox{red}{\includegraphics[width=.24\textwidth,valign=c]{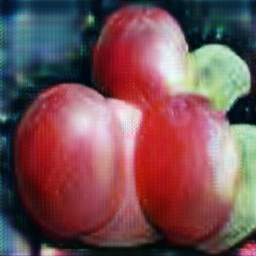}} &
\colorbox{red}{\includegraphics[width=.24\textwidth,valign=c]{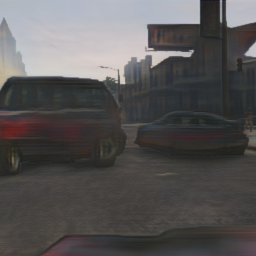}} &
\colorbox{red}{\includegraphics[width=.21\textwidth,valign=c]{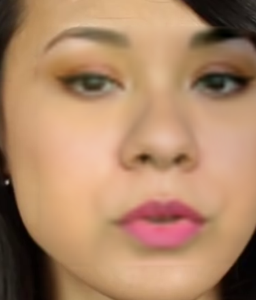}} &
\colorbox{red}{\includegraphics[width=.24\textwidth,valign=c]{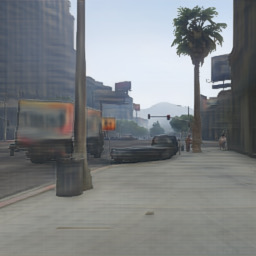}} &
\colorbox{red}{\includegraphics[width=.24\textwidth,valign=c]{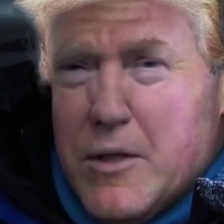}} &
\colorbox{red}{\includegraphics[width=.24\textwidth,valign=c]{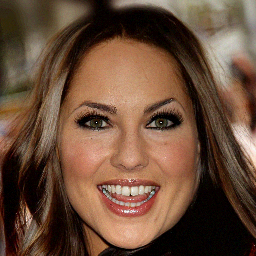}} &
\colorbox{red}{\includegraphics[width=.24\textwidth,valign=c]{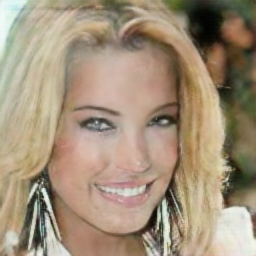}} &
\colorbox{red}{\includegraphics[width=.24\textwidth,valign=c]{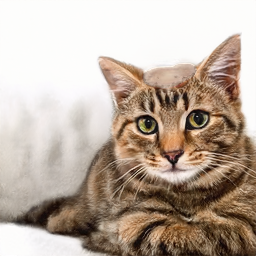}} &
\colorbox{red}{\includegraphics[width=.24\textwidth,valign=c]{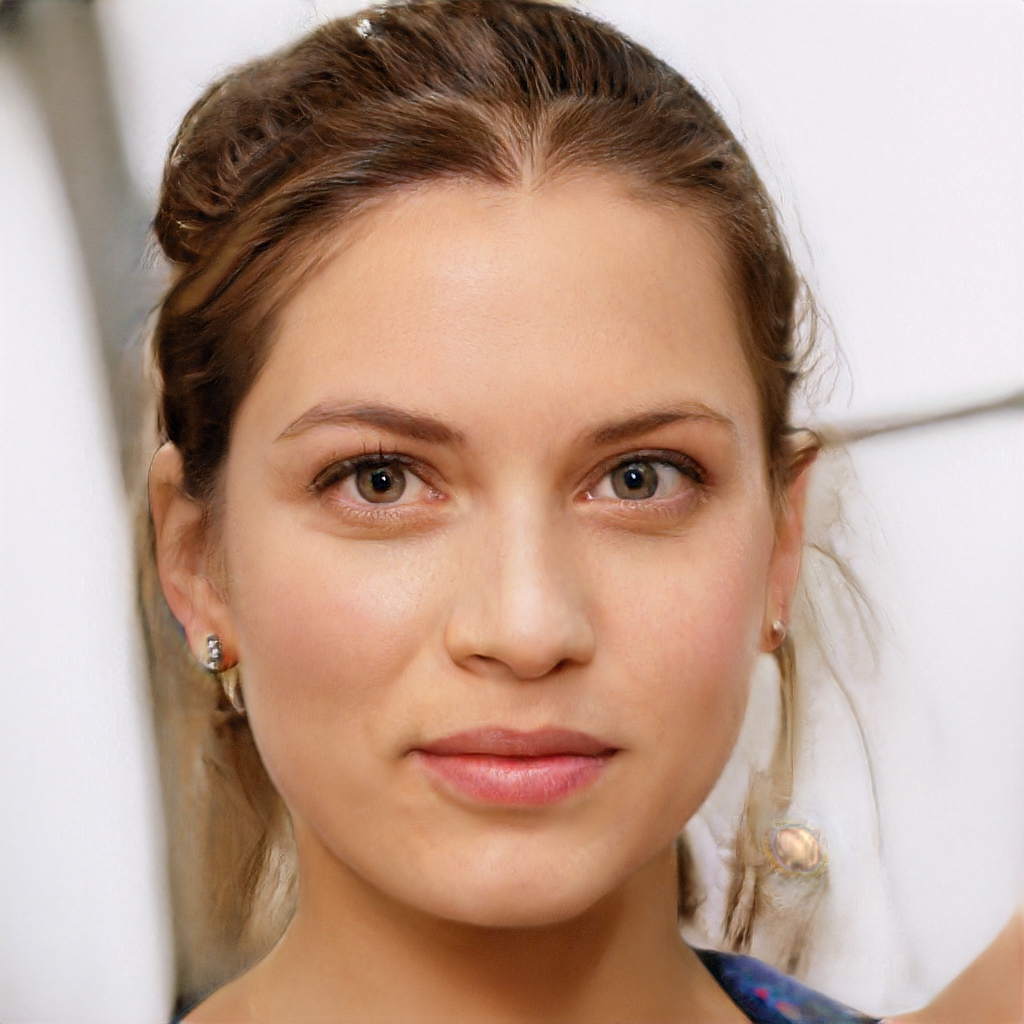}} &
\colorbox{red}{\includegraphics[width=.24\textwidth,valign=c]{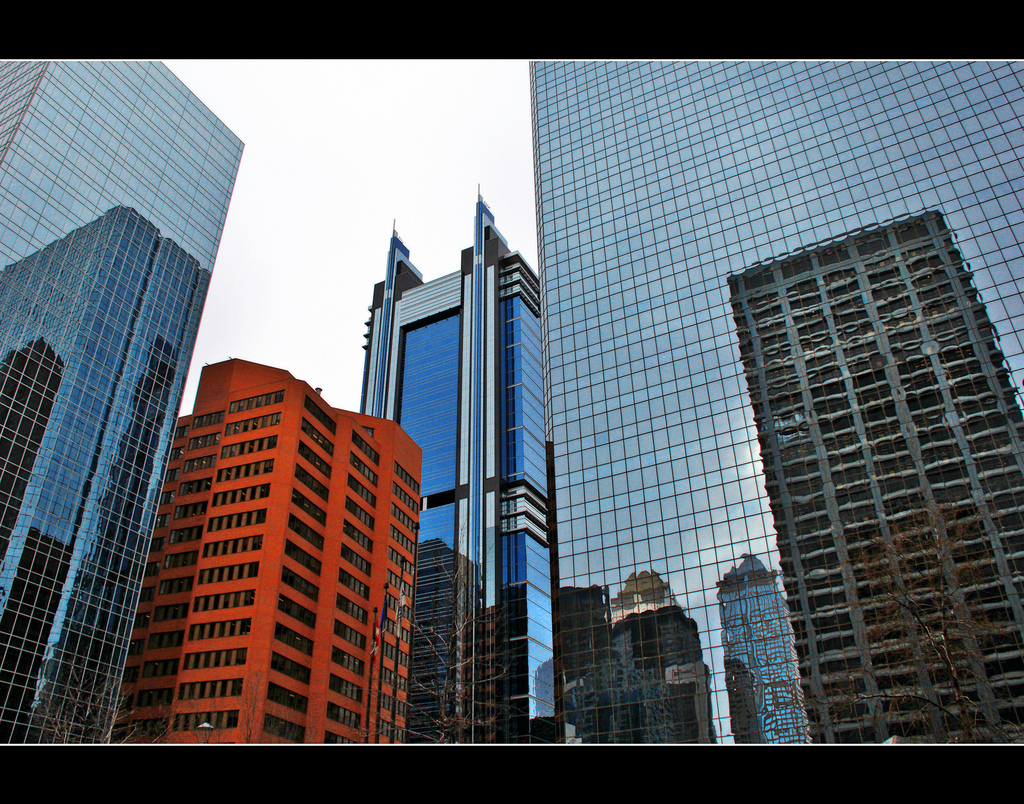}}
\\
GauGAN~\cite{park2019SPADE} & BigGAN~\cite{brock2018large} & CycleGAN~\cite{zhu2017unpaired} & IMLE~\cite{li2019diverse} & FaceForensic++~\cite{rossler2019faceforensics++} & CRN~\cite{chen2017photographic} & WildDeepfake~\cite{zi2020wilddeepfake} & Glow~\cite{kingma2018glow} & StarGAN~\cite{choi2018stargan} & StyleGAN~\cite{karras2019style} & WhichFaceReal~\cite{whichface} & SAN~\cite{dai2019second}
\\
\end{tabular}
}
\vspace{-0.2cm}
\captionof{figure}{The suggested continual deepfake detection benchmark (CDDB) aims to promote the research of learning a unified model over a  stream of likely heterogeneous deepfakes sequentially. The longest CDDB stream consists of the above 12 types of deepfake sources (Reals: \textcolor{green}{green} boundary, Fakes: \textcolor{red}{red} boundary). }
\label{fig:cdeepfake_ill}
\end{center}
}] 
{
  \renewcommand{\thefootnote}%
    {\fnsymbol{footnote}}
  \footnotetext[1]{Corresponding Author}
}

\input{00_abstract}

\input{01_introduction}

\input{02_relatedwork}

\input{03_benchmark}

\input{04_approach}

\input{05_experiment}

\input{06_conclusion}

\input{08_acknowledge}

{\small
\bibliographystyle{ieee_fullname}
\bibliography{egbib}
}

\input{07_supp}

\end{document}

%% file: 00_abstract.tex
\begin{abstract}
There have been emerging a number of benchmarks and techniques for the detection of deepfakes. However, very few works study the detection of incrementally appearing deepfakes in the real-world scenarios. To simulate the wild scenes, this paper suggests a continual deepfake detection benchmark (CDDB) over a new collection of deepfakes from both known and unknown generative models. The suggested CDDB designs multiple evaluations on the detection over easy, hard, and long sequence of deepfake tasks, with a set of appropriate measures. In addition, we exploit multiple approaches to adapt multiclass incremental learning methods, commonly used in the continual visual recognition, to the continual deepfake detection problem. We evaluate existing methods, including their adapted ones, on the proposed CDDB. Within the proposed benchmark, we explore some commonly known essentials of standard continual learning. 
Our study provides new insights on these essentials in the context of continual deepfake detection. 
The suggested CDDB is clearly more challenging than the existing benchmarks, which thus offers a suitable evaluation avenue to the future research. Both data and code are available at \url{https://github.com/Coral79/CDDB}.
    \vspace{-0.3cm}
\end{abstract}

%% file: 01_introduction.tex
\section{Introduction}
\label{sec:intro}
Deepfakes (deep learning generated fake images/videos) have become ubiquitous with the advent of increasingly improved deep generative models, such as autoencoders \cite{kingma2013auto}, generative adversarial nets (GANs) \cite{goodfellow2014generative} and generative normalizing flows (Glows) \cite{dinh2017density}. As a result, there is a growing threat of ``weaponizating'' deepfakes for malicious purposes, which is potentially detrimental to privacy, society security and democracy \cite{chesney2019deep}. To address this issue, many deepfake detection datasets (e.g.,\cite{korshunov2018deepfakes,li2018ictu,li2019celeb,dufour2019contributing,rossler2019faceforensics++,dolhansky2020deepfake,jiang2020deeperforensics}) and techniques (e.g., \cite{zhou2017two,afchar2018mesonet,yang2019exposing,nguyen2019use,nguyen2019multi,agarwal2019protecting}) are proposed.

State-of-the-art deep neural networks have made tremendous progress for
deepfake detection tasks in a stationary setup, where a large amount of relatively homogeneous deepfakes are provided all at once.
In this paper, we study a natural extension from this stationary deepfake detection scenario to a dynamic (continual) setting (Fig.\ref{fig:cdeepfake_ill}):
a stream of likely heterogeneous deepfakes appear time by time rather than at once, and the early appeared deepfakes cannot be fully accessed due to the streaming nature of data, privacy concerns, or storage constraints. In such a scenario, at each learning session when trained on a new deepfake detection task, standard neural networks typically forget most of the knowledge related to previously learned deepfake detection tasks \cite{marra2019incremental,kim2021cored}. This is essentially one of the most typical continual learning problems that are well-known to result in catastrophic forgetting \cite{robins1995catastrophic,rebuffi2017icarl,lopez2017gradient,chaudhry2018efficient}. Nevertheless, the study on the specific continual deepfake detection (CDD) problem, benchmarks as well as its particular nature remains fairly limited.

In this paper, we establish a challenging continual deepfake detection benchmark (CDDB) by gathering publicly available deepfakes, from both known and unknown generative models. The CDDB gradually introduces deepfakes to simulate the real-world deepfakes' evolution, as shown in Fig.\ref{fig:cdeepfake_ill}. The key benchmarking task is to measure whether the detectors are able to incrementally learn deepfake detection tasks without catastrophic forgetting. Up to our knowledge, there exist only two similar benchmarks~\cite{marra2019incremental,kim2021cored}. Both of these benchmarks are limited as they merely perform CDD on only one  deepfake type of known generative models (e.g., GANs or face-swap like deepfake models). As mentioned, the source of deepfakes may not only be unknown but also be of diverse types, in practice. Therefore, we proposed a new CDDB which better resembles the real-world scenarios. Furthermore, our CDDB also categorizes the evaluation protocol into different cases: from easy to hard and short to long CDD sequences. Such categorization allows us to better probe the CDD methods.

We evaluate a set of well-known and most promising existing continual learning methods on the established benchmark. In this process, we first evaluate the popular multi-class incremental learning methods in the CDD settings. Furthermore, we develop multiple approaches to adapt these continual learning methods to the binary CDD problem. Our evaluations also include several other variants which are evaluated for easy/hard and short/long sequences. These exhaustive evaluations offer two major benefits: (a) suitable baselines for the future CDD research; (b) new insights on the established  essentials in the context of CDD. In the latter case, we explore particular essentials including (i) knowledge distillation; (ii) class imbalance issue; (iii)  memory budget. Notably, our empirical evidence suggests that the existing consideration for class imbalance issues can be clearly hurtful to the CDD performance.

In summary, this paper makes three-fold contributions:
\begin{itemize}
    \vspace{-0.2cm}
    \item We propose a realistic and challenging continual deepfake detection benchmark over a collection of public datasets to thoroughly probe CDD methods.
    \vspace{-0.3cm}
    \item We comprehensively evaluate existing and adapted methods on the proposed benchmark. These evaluations serve as lock, stock, and barrel of CDD baselines. 
    \vspace{-0.7cm}
    \item Using the proposed dataset and conducted evaluations, we study several aspects of CDD problem. Our study offers new insights on the CDD essentials.
\end{itemize}

%% file: 02_relatedwork.tex
\section{Related Work}
\label{sec:relat}

\noindent\textbf{Datasets and Benchmarks for Deepfake Detection.} To evaluate deepfake detection methods, many datasets and benchmarks have been proposed. For instance, FaceForensic++ \cite{rossler2019faceforensics++} contains face videos collected from YouTube and manipulated using deepfake \cite{deepfakesgit}, Face2Face \cite{thies2016face2face}, Faceswap \cite{faceswap2009} and Neural Texture \cite{thies2019deferred}.
WildDeepfake \cite{zi2020wilddeepfake} aims at real-world deepfakes detection by collecting real and fake videos with unknown source model directly from the internet.
CNNfake \cite{wang2020cnn} proposes a diverse dataset obtained from various image synthesis methods, including GAN-based techniques (e.g., \cite{karras2018progressive}, \cite{brock2018large}) and traditional deepfake methods. 
Table \ref{tab:deepfake_datastats} summarizes the prominent benchmark datasets for deepfake detection.
The majority of the proposed benchmarks do not include incremental detection in their experimental setups. 
While few works, namely GANfake \cite{marra2019incremental} and CoReD \cite{kim2021cored}, have addressed the CDD setting, they either only address known GAN-based deepfakes \cite{marra2019incremental} or only treat known GAN fakes and known traditional deepfakes separately \cite{kim2021cored}. Furthermore, their studied task sequences are generally short (e.g. they consist of 4 or 5 tasks). However, in the real-world scenario, deepfakes might come from known or unknown source models. These models might be based on GANs or traditional methods, and finally, they form a long sequence of tasks evolving through time.
To bridge the gap between the current benchmarks and the real-world scenario, our suggested dataset includes a collection of deepfakes from both known or unknown sources. In addition, the suggested benchmark provides three different experimental setups (Easy, Hard, and Long) for a thorough evaluation of CDD methods.

\begin{table}[t]
    \centering 
    \resizebox{0.9\linewidth}{!}{
    {\def\arraystretch{1.2}
     \begin{tabular}{lllc@{}}
    \toprule Dataset & Real Source & Deepfake Source  & Continual \\
    \midrule
    Deepfake-TIMIT~\cite{korshunov2018deepfakes} & VidTIMIT dataset \cite{Sanderson2009lncs} & Known Deepfake tech & \xmark \\
    UADFW \cite{yang2019exposing} & EBV dataset \cite{li2018ictu}  & Known Deepfake tech & \xmark \\
    FaceForensics++~\cite{rossler2019faceforensics++} & YouTube & Known Deepfake tech & \xmark \\ 
    Celab-DF v2~\cite{li2020celeb} & YouTube & Known Deepfake tech & \xmark\\
    DFDC~\cite{dolhansky2020deepfake} & Actors & Known Deepfake tech & \xmark \\ 
    WildDeepfake~\cite{zi2020wilddeepfake} & Internet & Unknown Deepfake tech &
    \xmark \\
    WhichFaceReal~\cite{whichface} & Internet & Unknown Deepfake tech &
    \xmark \\
    CNNfake~\cite{wang2020cnn} & Multi-Datasets & Known Deepfake tech & \xmark \\
    GANfake~\cite{marra2019incremental} & Multi-Datasets & Known Deepfake tech & \cmark\\
    CoReD~\cite{kim2021cored} & Multi-Datasets & Known Deepfake tech & \cmark \\
    CDDB (ours) & \textbf{Multi-Datasets\&Internet} & \textbf{Known\&unknown tech} & \cmark\\
    \bottomrule
    \end{tabular}}
    }
    \vspace{-0.2cm}
    \caption{\scriptsize{Comparison of prominent deepfake datasets. Only CoReD \cite{kim2021cored}, GANfake \cite{marra2019incremental} and our CDDB study continual fake detection benchmarks. However, both CoReD and GANfake merely detect pure GAN-generated images (or pure deepfake-generated videos), while ours studies a high mixture of deepfake sources, which are from either known generative models or unknown ones (i.e., directly from internet).}}
    \label{tab:deepfake_datastats}
    \vspace{-0.5cm}
\end{table} 

\noindent\textbf{Deepfake Detection Methods.} Along with the discussed benchmarks, many approaches have been proposed for deepfake detection (e.g., \cite{rossler2019faceforensics++,li2018exposing,nguyen2019use,afchar2018mesonet,matern2019exploiting,nguyen2019multi,marra2019incremental,wang2020cnn,gerstner2022detecting,wang2022deepfake}). These approaches mainly aim at finding generalizable features from a set of available samples that can be used to detect deepfakes at test time. For instance,  \cite{rossler2019faceforensics++} employs XceptionNet \cite{chollet2017xception}, a CNN with separable convolutions and residual connections, pre-trained on ImageNet \cite{deng2009imagenet} and fine-tuned for deepfake detection. Similarly, \cite{wang2020cnn} uses ResNet-50 \cite{he2016deep} pretrained with ImageNet, and further train it in a binary classification setting for deepfake detection. 
Different to the aforementioned methods, \cite{marra2019incremental} and \cite{kim2021cored} address the CDD problem.
\cite{marra2019incremental} adapts one of the traditional CIL methods, i.e., incremental classifier and representation learning (iCaRL) \cite{rebuffi2017icarl}, through a multi-task learning scheme over both deepfake recognition and detection tasks. To mitigate the catastrophic forgetting, \cite{marra2019incremental} keeps using the original iCaRL\cite{rebuffi2017icarl}'s knowledge distillation loss, which enforces the newly updated network to output close prediction results to those of the network trained on the previous task given the same samples from an exemplar set of old samples. 
Similarly, CoReD \cite{kim2021cored} tackles forward learning and backward forgetting with a student-teacher learning paradigm, where the teacher is the model trained for the previous tasks, and the student is the new model being adapted to also include the current task. \cite{kim2021cored} only uses samples from the current task for the teacher-to-student knowlege distillation. To further alleviate  the forgetting, \cite{kim2021cored} adds a feature-level knowledge distillation loss (i.e., representation loss).

\noindent\textbf{Class-incremental Learning (CIL).}
In this paper, we focus on studying three prominent categories of CIL methods: \emph{gradient-based}, \emph{memory-based} and \emph{distillation-based}.\footnote{\scriptsize{Other CILs are based on regularization or expansion \cite{kirkpatrick2017overcoming,zenke2017continual,aljundi2018memory,yoon2017lifelong,lee2017overcoming,hung2019compacting,shahbazi2021efficient}.}}

\noindent\emph{Gradient-based methods} (e.g., \cite{riemer2018learning,zeng2019continual,farajtabar2020orthogonal,saha2020gradient,tang2021layerwise,wang2021training}) overcome catastrophic forgetting by minimizing the interference among task-wise gradients when updating network weights.
For instance, \cite{wang2021training} proposes a null space CIL (NSCIL) method to 
train the network on the new task by projecting its gradient updates to the null space of the approximated covariance matrix on all the past task data.

\noindent\emph{Memory-based methods} (e.g., \cite{robins1995catastrophic,rebuffi2017icarl,lopez2017gradient,chaudhry2018efficient,hu2018overcoming,kemker2018fearnet,chaudhry2019continual,shin2017continual,pellegrini2020latent,van2021class})  generally mitigate forgetting by replaying a small set of examples from past tasks stored in a memory.
For example, 
on the selected exemplars from previous tasks, latent replay CIL (LRCIL) \cite{pellegrini2020latent} suggests to replay their latent feature maps from intermediate layers of the model in order to reduce the required memory and computation.

\noindent\emph{Distillation-based methods} (e.g., \cite{li2017learning,rebuffi2017icarl,hou2019learning,castro2018end,wu2019large,yu2020semantic,tao2020topology,liu2020mnemonics,mittal2021essentials}) apply knowledge distillation \cite{hinton2015distilling} between a network trained on previous tasks and a network being trained on current task to alleviate the performance degradation on previous tasks. iCaRL \cite{rebuffi2017icarl} applies the distillation on a exemplar set that is selected by  using a herding method, which chooses samples closest to the sample means. \cite{rebuffi2017icarl} also identifies \emph{class imbalance} as a crucial challenge for continual multi-class classification (CMC). To address this, \cite{rebuffi2017icarl} suggests a classification strategy named nearest-mean-of-exemplars. 
Following the same motivation, LUCIR \cite{hou2019learning} applies cosine normalization to the magnitude of the parameters in fully-connected (FC) layers.
Based on the distillation idea, DyTox \cite{douillard2022dytox}\footnote{\scriptsize{Other transformer-based CIL methods (e.g. \cite{wang2021learning,deng2022incremental,yu2021improving,kahardipraja2021towards,li2022technical}) are emerging.}} applies the transformer ConViT \cite{d2021convit} to achieve the state-of-the-art CIL.

\noindent\textbf{Discussion.} The discussed CIL methods are mostly designed for CMC, which aims to learn a unified classifier for a set of sequentially-encountered classes. As one of the continual binary classification (CBC) problems, CDD can be regarded as a binary task or a set of binary tasks \cite{marra2019incremental}, and therefore we further investigate three general approaches of adapting these CIL methods to the CDD problem.

%% file: 03_benchmark.tex
\section{Suggested CDD Benchmark}
\label{sec:ben}

\begin{table}[t]
    \centering 
    \scriptsize
    \resizebox{0.9\linewidth}{!}{
    {\def\arraystretch{1.2}
     \begin{tabular}{@{}lllr@{}}
    \toprule Family & Deepfake Source & Real Source  & \# Images \\
    \midrule
    \multirow{6}{*}{\shortstack[l]{GAN Model}} & ProGAN~\cite{karras2018progressive} & LSUN & 736.0k  \\
    & StyleGAN~\cite{karras2019style} & LSUN & 12.0k  \\
    & BigGAN~\cite{brock2018large} & ImageNet & 4.0k  \\ 
    & CycleGAN~\cite{zhu2017unpaired} & Style/object transfer & 2.6k \\
    & GauGAN~\cite{park2019SPADE} & COCO & 10.0k \\
    & StarGAN~\cite{choi2018stargan} & CelebA & 16.8k  \\
\cdashline{1-4}
    \multirow{5}{*}{\shortstack[l]{Non-GAN Model}} 
    & Glow~\cite{kingma2018glow} & CelebA & 16.8k \\
    & CRN~\cite{chen2017photographic} & GTA & 12.8k  \\
    & IMLE~\cite{li2019diverse} & GTA & 12.8k \\  
    & SAN~\cite{dai2019second} & Standard SR benchmark & 440  \\  
   
    & FaceForensics++~\cite{rossler2019faceforensics++} & YouTube & 5.4k \\
      \cdashline{1-4}
    \multirow{2}{*}{\shortstack[l]{Unknown Model}} 
    & WhichFaceReal~\cite{whichface} & Internet & 2.0k \\

    & WildDeepfake~\cite{zi2020wilddeepfake} & Internet & 10.5k \\
    \bottomrule
    \end{tabular}}
    }
    \vspace{-0.2cm}
    \caption{\scriptsize{A new collection of mixed deepfake sources for the suggested CDDB.}}
    \label{tab:cdeepfake}
    \vspace{-5mm}
\end{table}

For a more real-world CDD benchmark, we suggest enforcing high heterogeneity over the deepfake data stream. In particular, we construct a new collection of deepfakes by gathering highly heterogeneous deepfakes collected by \cite{wang2020cnn,marra2019incremental,whichface,zi2020wilddeepfake}, which are from remarkably diverse resources. Moreover, the deepfakes from \cite{whichface,zi2020wilddeepfake}
has no information about their source generative models, and thus the new data collection reaches a more real-world scenario, which is always full of arbitrary deepfakes from either known or unknown sources.

\subsection{Data Collection}
\label{sec:coll}

The new data collection comprises 3 groups of deepfake sources: 1) GAN models, 2) non-GAN models, and 3) unknown models. 
Below details the deepfake sources and their associated real sources, which are listed in Table \ref{tab:cdeepfake}.

\noindent\textbf{GAN Models.} This group consists of fake images synthesized by \emph{6 GAN models}. ProGAN \cite{karras2018progressive} and StyleGAN \cite{karras2019style} are two of the most popular unconditional GANs. They were trained on each category of the dataset LSUN \cite{yu2015lsun} and thus they can produce realistic looking LSUN images. BigGAN \cite{brock2018large} is one of the state-of-the-art class-conditional GAN models trained on ImageNet \cite{deng2009imagenet}. Moreover, we include three image-conditional GAN models for image-to-image translation, namely CycleGAN \cite{zhu2017unpaired}, GauGAN \cite{park2019SPADE} and StarGAN \cite{choi2018stargan}. These models were trained on one style/object transfer task selected from the dataset collected by \cite{zhu2017unpaired}, COCO \cite{lin2014microsoft}, and CelebA \cite{liu2015deep}, respectively.

\noindent\textbf{Non-GAN Models.} This set contains deepfakes generated by \emph{8 non-GAN models}, including Generative Flow (Glow) \cite{kingma2018glow}, Cascaded Refinement Networks (CRN) \cite{chen2017photographic}, Implicit Maximum Likelihood Estimation (IMLE) \cite{li2019diverse}, second-order attention network (SAN) \cite{dai2019second} and 4 other deepfake models (Deepfake \cite{deepfakesgit}, Face2Face \cite{thies2016face2face}, Faceswap \cite{faceswap2009} and Neural Texture \cite{thies2019deferred}) from  \cite{rossler2019faceforensics++}. These models were trained on CelebA \cite{liu2015deep}, GTA \cite{richter2016playing}, a super-resolution dataset and YouTube videos, respectively, for image synthesis. 

\noindent\textbf{Unknown Models.} This group includes deepfake images from 2 \emph{unknown generative models}, one collected by WildDeepfake \cite{zi2020wilddeepfake} and one by WhichFaceReal \cite{whichface}. They both collect deepfakes and real images/videos directly from the internet. WildDeepfake \cite{zi2020wilddeepfake} originally contains deepfake/real videos. As our focus is on the detection of deepfake images, we randomly select a number of frames from each video. This group of models are included to further simulate the real-world, where the source model of the encountered deepfakes might not be known. 

\subsection{Evaluation Scenarios}
\label{sec:scene}

From the new collection, a large number of differently-ordered sequences of tasks can be produced to study CDD. In our benchmark, we suggest three different evaluation scenarios: an easy task sequence (\emph{EASY}), a hard task sequence (\emph{HARD}), and a long task sequence (\emph{LONG}). The \emph{EASY} setup is used to study the basic behavior of evaluated methods when they address an easy CDD problem. The \emph{HARD} setup aims to evaluate the performance of competing methods when facing a more challenging CDD problem. 
The \emph{LONG} setup is designed to encourage methods to better handle long sequences of deepfake detection tasks, where the catastrophic forgetting might become more serious. 

The three evaluation sequences are detailed as follows:
\begin{enumerate}
\small
\vspace{-0.2cm}
    \item \emph{EASY}: \{GauGAN, BigGAN, CycleGAN, IMLE, FaceForensic++, CRN, WildDeepfake\}
    \vspace{-0.3cm}
    \item \emph{HARD}: \{GauGAN, BigGAN, WildDeepfake, WhichFaceReal, SAN\}
    \vspace{-0.3cm}
    \item \emph{LONG}: \{GauGAN, BigGAN, CycleGAN, IMLE, FaceForensic++, CRN, WildDeepfake, Glow, StarGAN, StyleGAN, WhichFaceReal, SAN\}
    \vspace{-0.2cm}
\end{enumerate}
More concretely, the procedure on each sequence is to train a given model through the stream of involved training data sequentially, followed by an evaluation using the set of associated test data.
Following the common practice in CIL  \cite{rebuffi2017icarl,hou2019learning,tao2020topology,mittal2021essentials} and the suggestion in \cite{wang2020cnn}, we allow pre-training evaluated methods over the fake images of ProGAN and the corresponding real samples as a warm-up step.

\subsection{Evaluation Metrics}
\label{sec:metric}

CDD is a continual learning problem, and thus it should study the performance of the evaluated methods in terms of forward learning of each new task, as well as backward forgetting of previous ones. Accordingly, we suggest using the average detection accuracy (\emph{AA}) and the average forgetting degree (\emph{AF}), i.e., mean of backward transfer degradation (BWT) \cite{lopez2017gradient}, as the evaluation metrics. 
Formally, we can obtain a test accuracy matrix $B \in \mathbb{R}^{n\times n}$ (i.e., upper triangular matrix), where each entry $B_{i,j}$ indicates the test accuracy of the $i$-th task after training the $j$-th task and $n$ is the total number of involved tasks. \emph{AA} and \emph{AF} can be calculated as $AA=\frac{1}{n}\sum_{i=1}^{n}B_{i,n}$,  $AF = \frac{1}{n-1} \sum_{i=1}^{n-1} BWT_{i}$, where $BWT_{i}= \frac{1}{n-i-1}\sum_{j=i+1}^{n} (B_{i,j}-B_{i,i}) $. 

As CDD is also a detection problem, we suggest using mean average precision (\emph{mAP}) to measure the performance of evaluated methods in terms of the trade-off between precision and recall, where we regard real samples as negatives and fake samples as positive. Each AP is defined as the area under the precision-recall (PR) curve for deepfake detection over one single deepfake task in the sequence \cite{wang2020cnn}. \emph{mAP} is the average of all the APs calculated over all detection tasks.

As different tasks may contain the same or similar real samples, and some fake samples could be from unknown generative models, we additionally employ the average deepfake recognition accuracy (\emph{AA-M}) to study the challenge for identifying the specific deepfake and real resources. This metric is mainly used for understanding the gap between CMC and CBC.

%% file: 04_approach.tex
\section{Proposed Benchmarking Methods}
\label{sec:app}

The studied CDD problem requires to distinguish real and fake samples from the sequentially occurring sources of real/fake pairs. 
This section focuses on studying three main adaptions of CIL to better address the CDD problem.

\subsection{Problem Definition and Overview}

In the CDD problem, deepfakes and their corresponding real images appear sequentially in time, forming the sequence $\mathcal{X} = \{\mathcal{X}_1, \mathcal{X}_2 \ldots, \mathcal{X}_t\}$, where $\mathcal{X}_i=(X_i^R, X_i^F)$ represents the paired set of real and fake images corresponding to source $i$. 
At each incremental step $t$, complete data is only available for the new pair of real and fake image sets $\mathcal{X}_t=(X_t^R, X_t^F)$. For memory-based and distillation-based methods, we additionally use a small amount of exemplar data $\mathcal{P}=\{\mathcal{P}_1, \mathcal{P}_2 \ldots, \mathcal{P}_{t-1}\}$, which is selected from previous data $\mathcal{X} = \{\mathcal{X}_1, \mathcal{X}_2 \ldots, \mathcal{X}_{t-1}\}$ for rehearsal or distillation. The trained models at step $t$ are finally expected to distinguish all the fake and real images corresponding to the sources observed up to and including step $t$. 

The CDD problem can be relaxed to a CMC problem, once each real/fake source is regarded as an independent class. In this case, traditional CIL methods can be applied to CDD. A \emph{basic CIL system} is to train a network model $\Theta$ that consists of a deep feature extractor $\phi(\cdot)$ and a fully-connected (FC) layer $f_c(x)$. Like a standard multi-class classifier, the output logits are processed through an activation function $\varphi(\cdot)$ before classification loss $\ell_{\rm class}$ is evaluated corresponding to the correct class. To address catastrophic forgetting, many of the state-of-the-art CIL methods apply distillation loss $\ell_{\rm distill}$ over the exemplar set $\mathcal{P}$ of old samples from previous tasks. Besides, a supplementary loss $\ell_{\rm supp}$ is often applied to strengthen the classification over multiple classes at $\mathcal{P}$. In general, the loss function for CIL systems can be formulated as:
\vspace{-0.3cm}
\begin{equation}
    \ell_{\rm CIL}(\Theta) =  \ell_{\rm class}(\mathcal{X}_t,\Theta) +\gamma_d \ell_{\rm distill}(\mathcal{P},\Theta) + \gamma_m \ell_{\rm supp}(\mathcal{P},\Theta)
\label{eq:cilLoss}
\end{equation}
where $\gamma_d, \gamma_m$ are hyperparameters for the trade-off among the three losses. The classification loss is the usual cross-entropy loss calculated on the new data 
\vspace{-0.3cm}
\begin{equation}
    \ell_{\rm class}(\mathcal{X}_t,\Theta) = -\sum_{x_i \in \mathcal{X}_t} \sum_{y_j=1}^{k} \delta_{y_j=y_i} \log g_{y_j}(x_i)
    \label{eq:mutil_class}
    \vspace{-0.3cm}
\end{equation}
where $\delta_{y_j=y_i}$ is an indicator to check whether the prediction $y_j$ is in line with the ground truth $y_i$, $k$ is the total class number,
$g_{y_j}(\cdot)=\varphi(\cdot) \circ f_c(\cdot) \circ \phi(\cdot)$ computes the probability of the class $y_j$, 
and $\circ$ is function composition. The distillation term is a \textit{KL-divergence} loss \cite{hinton2015distilling} with temperature $T$:    
\vspace{-0.3cm}
\begin{equation}
    \ell_{\rm distill}(\mathcal{P},\Theta) = -\sum_{x_i \in \mathcal{P}} T^2 D_{KL}(g^T(x_i)||\tilde{g}^T(x_i) )
\label{eq:distill_loss}
\vspace{-0.3cm}
\end{equation}
where $\tilde{g}(\cdot)$ is the old classifier, which is the one before the current updating phase.
 
As done by \cite{hou2019learning}, the distillation can be also performed over the feature level, i.e. $\ell_{\rm distill}(\mathcal{P},\Theta) = -\sum_{x_i \in \mathcal{P}} 1-\langle\tilde{\phi}(x_i),\phi(x_i)\rangle$, where $\tilde{\phi}(\cdot),\phi(\cdot)$ are the feature extractors of the old and new networks respectively, and $\langle \cdot \rangle$ denotes cosine similarity. 
The additional term could be a margin ranking loss from \cite{hou2019learning}:
\vspace{-0.3cm}
\begin{equation}
    \small
    \ell_{\rm supp}(\mathcal{P},\Theta) = \sum_{x_i \in \mathcal{P}} \sum_{y_j=1}^J  \max(\tau-\langle\theta^{y_i},\phi(x_i)\rangle + \langle\theta^{y_j},\phi(x_i)\rangle,0)
\label{eq:supp_loss}
\vspace{-0.1cm}
\end{equation}
where $\langle \cdot \rangle$ indicates cosine similarity, $\tau$ is the margin threshold, $\theta$ is class embeddings (i.e., weight parameterization of the FC layer), $\theta^{y_i}$ is the ground truth class embedding of $x_i$, $\theta^{y_j}$ is one of the nearest-$J$ class embeddings. The  loss pushes nearest classes away from the given class. $\ell_{\rm supp}$ could be also a divergence loss \cite{douillard2022dytox}, which acts as an auxiliary classifier encouraging a better diversity among new classes and old ones. More details are presented in \cite{douillard2022dytox}.

The three losses can be also applied to different data. For example, $\ell_{\rm class}$ and $\ell_{\rm distill}$ can be also used to simultaneously learn over both the new and old data, i.e., $\mathcal{P} \cup \mathcal{X}_t$. In this paper, we mainly study four representative methods LRCIL \cite{pellegrini2020latent}, iCaRL \cite{rebuffi2017icarl}, LUCIR \cite{pellegrini2020latent} and DyTox \cite{douillard2022dytox}, all of which can be formulated by Eqn.\ref{eq:cilLoss}. For LRCIL, $\gamma_d$ and $\gamma_m$ are zeros. For iCaRL, $\gamma_m$ is zero. By comparison, LUCIR and DyTox applies non-zero $\gamma_d$ and $\gamma_m$.

\subsection{Adapting CIL to CDD}

We study three main adaptations of CIL methods for CDD. 
The key idea for the adaption is to enforce the classification loss $\ell_{\rm class}$ and the distillation loss $\ell_{\rm distill}$ to better fit the binary classification (i.e., detection) task, which is formulated as 
\vspace{-0.3cm}
\begin{equation}
    \ell_{\rm BIL}(\Theta) =  \hat{\ell}_{\rm class}(\mathcal{X}_t,\Theta) +\gamma_d \hat{\ell}_{\rm distill}(\mathcal{P},\Theta) + \gamma_m \ell_{\rm supp}(\mathcal{P},\Theta)
\label{eq:bilLoss}
\end{equation}
where $\hat{\ell}_{\rm class}$ and  $\hat{\ell}_{\rm distill}$ are the main adapted components. Below details three main ways for the adaption on them.

\noindent\textbf{Binary-class (BC) Learning.} One of the most straightforward solutions 
is to change the categorical cross-entropy loss with the binary one:
\vspace{-0.3cm}
\begin{equation}
\small
    \hat{\ell}_{\rm class}(\mathcal{X}_t,\Theta) = -\sum_{x_i \in \mathcal{X}_t} \delta_{y=y_i} \log g_y(x_i) + (1-\delta_{y=y_i}) \log(1-g_y(x_i))
    \vspace{-0.2cm}
\end{equation}
where $\delta_{y=y_i}$ is an indicator for the ground-truth label $y_i$, $g_y(x_i)$ applies Sigmoid function $\varphi(x_i)$ instead, calculating the probability of the given sample $x_i$ being a fake sample. In addition, the distillation loss $\hat{\ell}_{\rm distill}$ is based on the binary prediction. For the final classification, we apply the Sigmoid function based results. As $\ell_{\rm supp}$ is originally designed for better multi-class classification, it can be ignored in the binary adaptation.

\noindent\textbf{Multi-class (MC) Learning.} For this approach,
 we use the original classification, distillation, and supplementary losses, i.e. $\hat{\ell}_{\rm class}=\ell_{\rm class}$, $\hat{\ell}_{\rm distill}=\ell_{\rm distill}$, $\hat{\ell}_{\rm supp}=\ell_{\rm supp}$. We regard each real/fake class from different tasks as an independent class. For classification, we apply
$\hat{y_i} = \argmax_{y_j} g_{y_j} (x_i)$,  given a sample $x_i$. If $\hat{y_i}$ is one of the fake/real classes, we will predict $x_i$ to fake/real.

\noindent\textbf{Multi-task (MT) Learning.} Another adaption
is to apply a multi-task learning formulation. In particular, both the multi-class classification and the binary-class classification (i.e., detection) tasks are managed by the same classifier $g(\cdot)$. To this end, we adapt the classification loss by adding a binary cross-entropy term to the categorical cross-entropy
\vspace{-0.3cm}
\begin{equation}
\label{eq:mutlitask_loss}
    \small{
    \begin{aligned}
    \hat{\ell}_{\rm class}(\mathcal{X}_t,\Theta) & = (1-\lambda) \ell_{\rm class}(\mathcal{X}_t,\Theta) + \lambda \ell'_{\rm class}(\mathcal{X}_t,\Theta) \\
    \end{aligned}
    }
 \vspace{-0.1cm}
\end{equation}
where $\ell_{\rm class}$ is the regular multi-class classification task loss Eqn.\ref{eq:mutil_class}, the binary-class classification task loss $\ell'_{\rm class}$ is computed by taking into account the activations, $g(\cdot)$, of all the classes, separately for the fake and real classes, and the hyperparameter $\lambda$ is for the balance between these two task-based losses.
Formally, $\ell'_{\rm class}$ is computed by
\vspace{-0.3cm}
\begin{equation}
\small{
    \begin{aligned}
    \ell'_{\rm class}(\mathcal{X}_t,\Theta) = \sum_{x_i \in (\mathcal{X}_t)} \delta_{\scriptscriptstyle\mathrm{Y=F}} d_F(x_i) +  \delta_{Y=R} d_R(x_i) \\
    \end{aligned}
    }
        \vspace{-0.2cm}
\end{equation}
where $d_{F}(x_i)$ and $d_{R}(x_i)$ are designed for the aggregation over all fake classes and all real classes respectively. In this paper, we study the following four aggregation approaches: (1) \emph{SumLog} (Eqn.\ref{eqn:agg_bil1}) that is proposed in \cite{marra2019incremental}, (2) \emph{SumLogit} (Eqn.\ref{eqn:agg_bil2}), (3) \emph{SumFeat} (Eqn.\ref{eqn:agg_bil3}), and (4) \emph{Max} (Eqn.\ref{eqn:agg_bil4}). 

\vspace{-0.2cm}
\begin{subequations}
\small
\begin{align}
d_{F}(x_i) &= \hspace{-3mm} \sum_{y \in \{fake\}} \hspace{-3mm}\log g_y(x_i), \quad \hspace{4mm}
 d_{R}(x_i)  = \hspace{-3mm} \sum_{y \in \{real\}} \hspace{-3mm}\log g_y(x_i) \label{eqn:agg_bil1}\\
  d_{F}(x_i) & =  \log \hspace{-3mm} \sum_{y \in \{fake\}} \hspace{-3mm} g_y(x_i), \quad  \hspace{4mm}
   d_{R}(x_i)  =  \log \hspace{-3mm} \sum_{y \in \{real\}} \hspace{-3mm} g_y(x_i) \label{eqn:agg_bil2} \\
 d_{F}(x_i) & =  \log  g_y(\hspace{-3mm} \sum_{y \in \{fake\}} \hspace{-3mm} x_i), \quad \hspace{4mm}
  d_{R}(x_i) =  \log  g_y(\hspace{-3mm} \sum_{y \in \{real\}} \hspace{-3mm} x_i) \label{eqn:agg_bil3} \\
  d_{F}(x_i) & = \hspace{-2mm} \max_{y \in \{fake\}} \hspace{-2mm}(\log g_y(x_i)), \quad 
    d_{R}(x_i) = \hspace{-2mm} \max_{y \in \{real\}} \hspace{-2mm}(\log g_y(x_i)) \label{eqn:agg_bil4}
\end{align}
\label{eqn:agg_bil}%
\end{subequations}
where $y \in \{fake\}$ indicates all the associated fake classes, and $y \in \{real\}$ corresponds to all the real classes. 

For the MT case, we use the original distillation and supplementary losses, i.e. $\hat{\ell}_{\rm distill}=\ell_{\rm distill}$, $\hat{\ell}_{\rm supp}=\ell_{\rm supp}$, where $\ell_{\rm distill}$ and $\ell_{\rm supp}$ are computed using Eqn.\ref{eq:distill_loss} and Eqn.\ref{eq:supp_loss}, respectively. As done in the MC case, we use $\hat{y_i} = \argmax_{y_j} g_{y_j} (x_i)$ for the final classification.

%% file: 05_experiment.tex
\section{Benchmarking Results}
\label{sec:exp}

\begin{table*}[t]
 \tiny
  \centering
   \resizebox{0.85\linewidth}{!}{%
    \begin{tabular}{p{1.5cm}|p{2.43cm}||c|c|c|c|c|c|c||c|c|c|c}
& & \multicolumn{6}{c}{\textbf{CDDB-EASY1500} } & &  & \\ \hline
Learning Sys. & Evaluated Method  & Task1  & Task2  & Task3   & Task4   & Task5 & Task6 & Task7 & AA & AF & AA-M & mAP\\ \hline

 \multirow{4}[0]{*}{Baseline} & CNNDet\cite{wang2020cnn}-Zeroshot & 63.85 & 68.75 & 71.95 & 60.93 & 91.31 & 60.93 & 49.01 & 66.68 & NA & NA & 79.02 \\

 & CNNDet\cite{wang2020cnn}-Finetune & 57.25 & 51.00 & 57.63 & 47.81 & 80.41 & 48.00 & 78.77 & 60.28 & -14.29 & NA & 61.67\\
 
 & ConViT\cite{d2021convit}-Zeroshot & 89.55  & 75.38 & 94.66 & 98.51 & 80.22 & 96.67 & 49.25 & 83.46 & NA & NA & 61.04\\
 
 & ConViT\cite{d2021convit}-Finetune\dag & 51.45  & 49.25 & 52.86 & 51.49 & 77.63 & 55.49 & 86.28 & 60.64 & -42.75 & NA & 56.92\\
 
 \hline
 
 \multirow{4}[0]{*}{\shortstack[l]{Binary-class  \\ (BC) learning}} & NSCIL\cite{wang2021training}-\textbf{Sigmoid*}  & 48.35 & 50.25 & 49.43 & 56.58 & 56.56 & 70.73 & 57.63 &  55.65 & -42.04 & NA & 63.50 \\
 & LRCIL\cite{pellegrini2020latent}-\textbf{Sigmoid*}  &  83.00 & 88.00 & 82.82 & 96.20 & 79.02 & 97.14 & 62.82  & \textcolor{green}{84.14} & -9.15  & NA & \textcolor{green}{91.37}\\
 & iCaRL\cite{rebuffi2017icarl}-\textbf{Sigmoid*} & 76.90 & 80.00 & 88.93 & 99.41 & 85.03 & 99.45 & 76.64  & \textcolor{blue}{87.05} & -10.64  & NA & \textcolor{blue}{92.25}\\
 & LUCIR\cite{hou2019learning}-\textbf{Sigmoid*} & 90.60  & 91.05 & 90.46 & 99.80 & 91.04 & 99.80 & 75.38 & \textcolor{red}{91.16} & -4.76 & NA & \textcolor{red}{\textbf{95.94}}\\
 
 \hline
 \multirow{5}[0]{*}{\shortstack[l]{Multi-class  \\ (MC) learning}}  & NSCIL\cite{wang2021training} & 46.80  & 52.50 & 47.90 & 45.26 & 35.21 & 51.33 & 58.85 & 48.26 & -50.88  & 8.41 & 44.26\\
 
 & LRCIL\cite{pellegrini2020latent}  & 83.50  & 77.88 & 90.84 & 98.90 & 84.75 & 98.86 & 65.92  & \textcolor{green}{85.81} & -5.88 & 67.11 & \textcolor{green}{92.63}\\

 & iCaRL\cite{rebuffi2017icarl}  & 77.50 & 71.38 & 91.22 & 99.57 & 95.66 & 99.92 & 78.28 & \textcolor{blue}{87.65} & -9.41 & 65.39 & \textcolor{blue}{94.12}\\

 & LUCIR\cite{hou2019learning}-\textbf{LinFC}  & 91.60  & 89.12 & 92.56 & 99.76 & 94.45 & 99.80 & 71.21  & \textcolor{red}{91.21} & -2.88 & 74.62 & \textcolor{red}{94.75}\\

 & DyTox\cite{douillard2022dytox} & 98.30  & 94.25 & 98.85 & 100.00 & 95.66 & 100.00 & 85.94 & \textcolor{cyan}{96.14} & -1.24 & 83.66 & \textcolor{cyan}{93.90}\\

 \hline

 \multirow{16}[0]{*}{\shortstack[l]{Multi-task  \\ (MT) learning}} & LRCIL\cite{pellegrini2020latent}-SumLog\cite{marra2019incremental}  & 86.65 & 85.63 &  91.79 & 99.22 & 88.72 & 99.57 & 68.27 & \textcolor{green}{88.76} & -3.99 & 66.03 & \textcolor{green}{93.95}\\

 & iCaRL\cite{rebuffi2017icarl}-SumLog\cite{marra2019incremental}  & 74.40  & 78.38 & 88.36 & 99.65 & 92.24 & 99.69 & 79.84 & \textcolor{blue}{87.05} & -10.72 & 71.41 & \textcolor{blue}{93.02}\\

 & LUCIR\cite{hou2019learning}-SumLog\cite{marra2019incremental}  & 88.70  & 88.62 & 93.89 & 99.37 & 94.82 & 99.80 & 75.71 & \textcolor{red}{91.56} & -3.65  & 74.47 & \textcolor{red}{95.47}\\

 & DyTox\cite{douillard2022dytox}-SumLog\cite{marra2019incremental}\ddag &  98.30 & 95.00 & 99.43 & 100.00 & 96.30 & 100.00 & 85.89 & \textcolor{cyan}{\underline{96.42}} & -0.72 & 83.94 & \textcolor{cyan}{94.01}\\

  & LRCIL\cite{pellegrini2020latent}-\textbf{SumLogit} & 86.95 & 88.12 & 92.75 & 99.45 & 88.35 & 99.45 & 70.24 & \textcolor{green}{\textbf{89.33}} & -4.27 & 68.00 & \textcolor{green}{\textbf{94.84}}\\

  & iCaRL\cite{rebuffi2017icarl}-\textbf{SumLogit}  & 85.25 & 86.12 & 88.93 & 99.65 & 92.98 & 99.80 & 80.27 & \textcolor{blue}{\textbf{90.43}} & -6.12 & 74.18 & \textcolor{blue}{\textbf{95.27}}\\

 & LUCIR\cite{hou2019learning}-\textbf{SumLogit}  & 89.95  & 89.62 & 94.47 & 99.65 & 95.75 & 99.80 & 74.79 & \textcolor{red}{\textbf{92.00}} & -3.13 & 73.55 & \textcolor{red}{95.26} \\

 & DyTox\cite{douillard2022dytox}-\textbf{SumLogit} & 98.30  & 94.75 & 99.05 & 100.00 & 96.86 & 100.00 & 86.82 & \textcolor{cyan}{\textbf{96.54}} & -0.82 & 84.42 & \textcolor{cyan}{\textbf{94.23}}\\

  & LRCIL\cite{pellegrini2020latent}-\textbf{SumFeat} & 84.85 & 85.13 & 92.56 & 99.26 & 87.15 & 99.45 & 69.90 & \textcolor{green}{87.81} & -5.30 & 66.40 & \textcolor{green}{93.73}\\

  & iCaRL\cite{rebuffi2017icarl}-\textbf{SumFeat} & 77.90  & 84.25 & 90.84 & 99.65 & 82.72 & 99.69 & 79.11 & \textcolor{blue}{87.74} & -9.52 & 68.15 & \textcolor{blue}{93.88} \\
 & LUCIR\cite{hou2019learning}-\textbf{SumFeat} & 90.05  & 89.38 & 94.85 & 99.92 & 95.01 & 99.92 & 73.53 & \textcolor{red}{\underline{91.81}} & -3.12  & 74.23 & \textcolor{red}{\underline{95.66}}\\
 
 & DyTox\cite{douillard2022dytox}-\textbf{SumFeat} & 98.05  & 93.63 & 99.24 & 100.00 & 96.12 & 100.00 & 86.67 & \textcolor{cyan}{96.24} & -1.13 & 83.98 & \textcolor{cyan}{\underline{94.20}}\\

  & LRCIL\cite{pellegrini2020latent}-\textbf{Max} & 87.80  & 89.00 & 92.18 & 67.96 & 87.15 & 99.22 & 69.85 & \textcolor{green}{\underline{89.16}} & -4.43 & 69.38 & \textcolor{green}{\underline{94.60}}\\
  & iCaRL\cite{rebuffi2017icarl}-\textbf{Max}  & 82.35  & 87.00 & 92.94 & 99.76 & 91.77 & 99.73 & 79.74 & \textcolor{blue}{\underline{89.92}} & -6.79  & 73.47 & \textcolor{blue}{\underline{94.87}}\\

 & LUCIR\cite{hou2019learning}-\textbf{Max}  & 89.85  & 91.25 & 94.08  & 99.53 & 92.51 & 99.65 & 72.61 & \textcolor{red}{91.21} & -4.00  & 74.06 & \textcolor{red}{95.41}\\

 & DyTox\cite{douillard2022dytox}-\textbf{Max} & 98.80  & 92.63 & 99.05 & 100.00 & 96.12 & 100.00 & 86.23 & \textcolor{cyan}{96.12} & -1.19 & 84.14 & \textcolor{cyan}{94.08}\\

 \hline

 \multirow{3}[0]{*}{Joint Training} & CNNDet \cite{wang2020cnn}-Binary  & 98.65  & 98.38 & 96.37 & 100.00 & 95.19 & 100.00 & 79.59  & 95.20 & NA & NA & 98.36\\
 
  & CNNDet \cite{wang2020cnn}-Multi & 95.70 & 97.00 & 95.80 & 99.96 & 96.30 & 99.96 & 75.28 & 94.29 & NA & 79.28 & 97.25\\
 & DyTox\cite{douillard2022dytox}-Multi & 99.40  & 96.37 & 98.28 & 100.00 & 94.64 & 100.00 & 80.22 & 95.53 & NA & 81.87 & 95.34\\
    \end{tabular}
    }
    
    \vspace{-0.3cm}
\caption{\scriptsize{ Benchmarking results on the suggested CDDB's \emph{EASY} evaluation. CNNDet/ConViT-Zeroshot is merely trained on ProGAN, and CNNDet/ConViT-Finetune is tuned over the 7 tasks. ConViT\cite{d2021convit}-Finetune\dag: low AA/mAP seems attributed to huge forgetting. Sigmoid*: applying Sigmoid function based classification loss. SumLog\cite{marra2019incremental}\ddag: most cases fail, only $\lambda=0.0001$ works. AA: Average Accuracy for deepfake detection, AF: Average Forgetting degree, AA-M: Average Accuracy for deepfake recognition, mAP: mean Average Precision. \textcolor{green}{green}: LRCIL, 
\textcolor{blue}{blue}: iCaRL,
  \textcolor{red}{red}: LUCIR, \textcolor{cyan}{cyan}: DyTox. \textbf{Bold}: best \textcolor{green}{\textbf{green}}/\textcolor{blue}{\textbf{blue}}/\textcolor{red}{\textbf{red}}/\textcolor{cyan}{\textbf{cyan}} results, \underline{Underline}: second best \textcolor{green}{\underline{green}}/\textcolor{blue}{\underline{blue}}/\textcolor{red}{\underline{red}}/\textcolor{cyan}{\underline{cyan}} results.}}
  \label{tab:cddb_easy}
  \vspace{-0.5cm}
\end{table*}

We evaluate three families of CIL methods with our exploited variants (using BC, MC, MT) on the suggested CDD benchmark for the three scenes (Sec.\ref{sec:scene}): 1) \emph{EASY}, 2) \emph{HARD}, and 3) \emph{LONG}, with the introduced measures (Sec.\ref{sec:metric}). 
The three sets of state-of-the-art CIL methods are: 1) Gradient-based: NSCIL \cite{wang2021training}, 2) Memory-based: LRCIL \cite{pellegrini2020latent}, and 3) Distillation-based: iCaRL \cite{rebuffi2017icarl}, LUCIR \cite{hou2019learning}
and DyTox \cite{douillard2022dytox}.
Besides, we evaluated the multi-task variant of iCaRL (iCaRL-SumLog)\footnote{\scriptsize{We overlooked the method \cite{kim2021cored}, as its code is not publicly available.}} \cite{marra2019incremental}.
We adapted the top-4 best CIL methods (i.e., LRCIL, iCaRL, LUCIR, DyTox) using BC, MC, MT. Note that it is non-trivial to adapt DyTox with BC, as the loss is tightly constrained for multiple-class classification. Thus, we do not evaluate its BC variant. For a fair comparison, except for DyTox that is based on an ImageNet pre-trained transformer ConViT \cite{d2021convit}, we employ the state-of-the-art deepfake CNN detector (CNNDet) \cite{wang2020cnn} that applies a ResNet-50 pretrained on ImageNet \cite{deng2009imagenet} and ProGAN \cite{karras2018progressive} as the  backbone for all the rest methods over the proposed CDDB. 
For most methods, we used their official codes, and tuned their hyperparameters for better performances. For a consistency, we empirically set the MT learning hyperparameter as $\lambda=0.3$ for all the MT methods. For \emph{EASY, HARD} and \emph{LONG}, we assign the same memory budget (i.e., 1500) for all those methods that need a memory to save exemplars. Additionally, we study three reduced memory budgets (i.e., 1000, 500, 100) for \emph{HARD}. For all the evaluations, we evaluate the joint training methods using CNNDet/ConViT with either binary classification loss (CNNDet-Binary) or  multiclass classification loss (CNNDet/ConViT-Multi) to study the approximated upper bound for the incremental learning methods. The detailed settings, the parameter setups, and more empirical studies are presented in Suppl. Material\footnote{\scriptsize{The supp. metrial also studies GANfake \cite{marra2019incremental}. As it does not release the detailed train/val/test splits, we can only empirically study our own splits following the description in the oiriginal paper.}}.

\subsection{EASY  Evaluation}

Table \ref{tab:cddb_easy} reports the benchmarking results of the evaluated methods for the CDDB \emph{EASY}. Table \ref{tab:cddb_easy_essentials} studies other essential components for the CDD problem.

\begin{table}[t]
 \scriptsize
  \centering
  \resizebox{0.85\linewidth}{!}{%

    \begin{tabular}{p{1.9cm}|p{2.43cm}||c|c|c}
& & \multicolumn{3}{c}{\textbf{CDDB-EASY1500} } \\ \hline
Learning System & Evaluated Method  & AA & AF & AA-M \\ \hline
\multirow{4}[0]{*}{\shortstack[l]{Multi-class}}  &  LRCIL\cite{pellegrini2020latent}   & \textcolor{green}{85.81} & -5.88 & 67.11\\

 & LUCIR(CosFC)\cite{hou2019learning}   & \textcolor{red}{87.24} & -10.32 & 71.42\\
 
 & LRCIL\cite{pellegrini2020latent}-\textbf{KD} & \textcolor{green}{86.85} & -5.50 & 66.57 \\
 
 & LUCIR\cite{hou2019learning}-\textbf{LinFC}   & \textcolor{red}{91.21} & -2.88 & 95.41\\

    \end{tabular}
    }
 \vspace{-0.3cm}
\caption{\scriptsize{Evaluation results on essentials of CILs on the \emph{EASY} evaluation. AA:  Average Accuracy for detection, AF: Average Forgetting degree, AA-M: Average recognition Accuracy. Results of LRCIL and LUCIR are in \textcolor{green}{green},  \textcolor{red}{red} respectively.}}
  \label{tab:cddb_easy_essentials}
  \vspace{-0.5cm}
\end{table}

\noindent\textbf{Vanilla Deepfake Detection vs. Continual one (CDD).} As discussed in Sec.\ref{sec:intro} and Sec.\ref{sec:relat}, most of deepfake detection techniques are designed to single/stationary deepfake detection tasks.
We follow CNNDet\cite{wang2020cnn}'s suggestion to train it on the ProGAN dataset, and applied the pre-trained model to the seven tasks directly in the EASY evaluation. Thus, we further name this method as CNNDet\cite{wang2020cnn}-Zeroshot. Besides, we also compare the one that finetunes the pre-trained CNNDet on each new task, due to the access limit to the old task data. The same training strategies are also applied to one of the state-of-the-art transformers, i.e., ConViT \cite{d2021convit}. By comparing the continual learning setups (BC, MC, MT), we can find that all the zeroshot and finetune setups perform clearly worse, showing that the necessity of applying the continual learning approaches to the CDD problem.

\noindent\textbf{Basic Findings on Essentials for CDD.} 1) Due to the serious \emph{forgetting} problem, \emph{finetuning} the CNNDet/ConViT methods works worse than the pretrained CNNDet/ConViT models over ProGAN. 2) The clearly inferior performances of the \emph{state-of-the-art gradient-based method NSCIL} might be because its null space cannot be approximated well on the highly heterogeneous fakes and reals. 3) LRCIL merely performs a rehearsal on the data to address forgetting, and thus we added a \emph{knowledge distillation} (KD) term to LRCIL (i.e., LRCIL-KD), which further improves AA with a marginally worse AA-M. (see Table \ref{tab:cddb_easy_essentials}). 
4) \emph{class imbalance issue} is not so important to CDD. From Table \ref{tab:cddb_easy_essentials}, we discover that its used cosine normalization based fully connected layer (CosFC) \cite{hou2019learning} clearly hurts the performance (see the comparison LUCIR-CosFC vs. LUCIR-LinFC in Table \ref{tab:cddb_easy_essentials}). CosFC is proposed to address the \emph{class imbalance issue}, with which the test samples are often predicted into new classes in the CMC context. However, predicting the test fakes into the new fake class is acceptable for our studied CDD problem. Therefore, the normalization techniques is very likely to hurt the CDD performance. Based on the observation, we replace CosFC with regular FC (LinFC) in all LUCIR based variants for better performances.

\noindent\textbf{CNN vs. ViT.} The CNN-based methods are mostly outperformed by the ViT-based methods (ConViT and DyTox), as we can see in Table \ref{tab:cddb_easy}. Nevertheless, we should notice that the parameter size ($\approx$86M) of the used ConViT is about 3 times larger than that ($\approx$25M) of CNNDet (ResNet50). The study provides two choices when we address the CDD problem. One is the family of light CNNDet models, and the other one is the group of heavy ConViT mdoels.

\noindent\textbf{BC vs. MC vs. MT Learning Systems.} Table \ref{tab:cddb_easy} reflects that DyTox gets almost saturated performances on the EASY eveluation (even higher than its upperbound DyTox-Multi). Also we discover that its variants performs almost the same in terms of AA, while DyTox-MC works generally worse than DyTox-MT in the reduced memory case (see Tabel \ref{tab:dytox_memory_main}). Hence, we turn to concentrate on the comparison over the lighter CNN-based methods. Except for the case on LRCIL, Table \ref{tab:cddb_easy} shows that the BC variants of the rest models (like iCaRL and LUCIR) perform very comparably with their corresponding MC models in terms of AA and AF, showing that the fine-grained classification benefits the detection. It is also good for the MC method to eventually label an image as a fake if the classifier decides for any of the fake classes. By comparison, most of the MT variants of LRCIL, iCaRL and LUCIR work better than (or at least comparable with) their corresponding BC and MC models in terms of AA, and some of MTs perform clearly better than their corresponding MCs in terms of AA-M. This mainly stems from the natural complementary properties between the fine-grained multi-class separation and the coarse-grained binary-class cohesion, and most of the suggested MT methods  balance them well.

\noindent\textbf{{SumLog vs. SumLogit vs. SumFeat vs. Max.}} From  Table \ref{tab:cddb_easy}, we can see that the \emph{SumLogit} and \emph{Max} variants mostly work better than the original \emph{SumLog} \cite{marra2019incremental} for the two main measures, i.e., AA and mAP. This is mainly because of consistency with final classifier's operation, which applies $argmax$ to the resulting logits. By comparison, the \emph{SumFeat} variants merely perform better than the original \emph{SumLog} for LUCIR in terms of AA and mAP. This might be because, different from LRCIL and iCaRL, it additionally applies the metric learning like loss (i.e., margin ranking loss) and thus the resulting features might be more discriminative for the aggregation to address the binary classification.  

\subsection{HARD and LONG Evaluations}

We select the top-2 BC, MC and MT models in terms of AA for LRCIL, iCaRL and LUCIR from the \emph{EASY} evaluation for bechmarking \emph{HARD} and \emph{LONG}. 

\noindent\textbf{EASY vs. HARD vs. LONG.} As the LONG evaluation includes both easy and hard tasks, the evaluated methods generally get higher scores than those in HARD. Besides, LONG's overall AAs and mAPs are visibly worse than those on EASY, meaning that incremental learning over a longer sequence is more challenging. Interestingly, the AFs are not clearly worse, showing the forgetting issue is not serious in this context. By comparison, HARD's overall AAs, mAPs and AA-Ms are clearly lower than those of EASY and LONG, because it contains more challenging tasks such as WildDeepfake, WhichFaceReal that are from the wild scenes and SAN that merely has a small data for training.

\begin{table}[t]
  \centering
  \scriptsize
    \resizebox{1\linewidth}{!}{%

    \begin{tabular}{l|l||c|c|c|c||c|c|c|c}
 &  & \multicolumn{4}{c||}{\textbf{CDDB-HARD1500} } & \multicolumn{4}{c}{\textbf{CDDB-LONG1500}}\\ \hline
Learning System  & Evaluated Method & AA & AF & AA-M & mAP & AA & AF & AA-M & mAP \\ \hline
\multirow{3}[0]{*} {Binary-class} & LRCIL \cite{pellegrini2020latent}-\textbf{Sigmoid*} & \textcolor{green}{74.07} & -5.43   & NA  & \textcolor{green}{80.40} & \textcolor{green}{87.06} & -4.79  & NA  & \textcolor{green}{91.81} \\ 

 &  iCaRL  \cite{rebuffi2017icarl}-\textbf{Sigmoid*}   & \textcolor{blue}{80.52}  & -7.89 & NA  & \textcolor{blue}{87.76} & \textcolor{blue}{87.12}  & -6.92 & NA  & \textcolor{blue}{91.68} \\ 
 
&  LUCIR  \cite{hou2019learning}-\textbf{Sigmoid*}    & \textcolor{red}{\underline{83.17}}  & -4.45  & NA  & \textcolor{red}{\textbf{89.72}}   & \textcolor{red}{87.19}  & -6.91  & NA  & \textcolor{red}{92.11}  \\

 \hline

\multirow{4}[0]{*} {Multi-class} &  LRCIL \cite{pellegrini2020latent}   &  \textcolor{green}{74.22}  & -5.41 & 57.74  & \textcolor{green}{80.21} &  \textcolor{green}{86.40}  & -4.65  & 62.32 & \textcolor{green}{91.28} \\  

&  iCaRL  \cite{rebuffi2017icarl} & \textcolor{blue}{76.72} & -7.79  & 67.46  &  \textcolor{blue}{85.05} & \textcolor{blue}{82.50} & -10.76  & 53.49  & \textcolor{blue}{89.63}  \\ 

&  LUCIR \cite{hou2019learning}-\textbf{LinFC}  & \textcolor{red}{82.33}   & -3.12  & 66.27  & \textcolor{red}{86.73} & \textcolor{red}{86.95}   & -6.77  & 69.74  & \textcolor{red}{91.43} \\

&  DyTox \cite{douillard2022dytox}  & \textcolor{cyan}{88.46}  & -0.72  & 79.68  & \textcolor{cyan}{88.06} & \textcolor{cyan}{93.34}  & -1.67  & 78.62 & \textcolor{cyan}{\textbf{94.08}} \\

 \hline

\multirow{8}[0]{*} {Multi-task} & LRCIL \cite{pellegrini2020latent}-\textbf{SumLogit}  & \textcolor{green}{\textbf{77.28}} & -2.70   & 60.22  & \textcolor{green}{\textbf{81.38}}  &  \textcolor{green}{\underline{87.93}}  & -2.99  &  65.60 & \textcolor{green}{\underline{92.45}}  \\

& iCaRL \cite{rebuffi2017icarl}-\textbf{SumLogit}   & \textcolor{blue}{\underline{81.16}} & -7.84  & 69.00  & \textcolor{blue}{\textbf{90.22}} & \textcolor{blue}{\underline{88.68}} & -5.12  & 70.51  & \textcolor{blue}{\underline{93.10}} \\

&  LUCIR \cite{hou2019learning}-\textbf{SumLogit}      & \textcolor{red}{\textbf{83.42}} & -3.28  & 65.31  & \textcolor{red}{\underline{87.89}}  & 
  \textcolor{red}{\textbf{88.57}} & -5.78  & 71.55  & \textcolor{red}{\textbf{92.83}} \\

&  DyTox \cite{douillard2022dytox}-\textbf{SumLogit}  & \textcolor{cyan}{\underline{88.60}}  & -0.62  & 80.59  & \textcolor{cyan}{\underline{89.15}} & \textcolor{cyan}{\underline{93.80}}  & -1.41  & 78.86  & \textcolor{cyan}{\underline{92.47}} \\

&  LRCIL \cite{pellegrini2020latent}-\textbf{Max}  &  \textcolor{green}{\underline{76.93}} & -2.55  & 59.20  & \textcolor{green}{\underline{81.20}}  &   \textcolor{green}{\textbf{88.49}}  & -2.64  & 65.67 & \textcolor{green}{\textbf{92.49}}  \\

&   iCaRL \cite{rebuffi2017icarl}-\textbf{Max}    &  \textcolor{blue}{\textbf{81.28}} & -8.95   & 60.64  & \textcolor{blue}{\underline{88.61}}  &  \textcolor{blue}{\textbf{89.05}}  & -5.24  & 71.35 & \textcolor{blue}{\textbf{94.03}}   \\

&  LUCIR \cite{hou2019learning}-\textbf{SumFeat}      & \textcolor{red}{82.14} & -2.90  & 65.59  & \textcolor{red}{87.39}  & \textcolor{red}{\underline{88.40}} & -5.30  & 70.99  & \textcolor{red}{\underline{92.70}} \\ 

&  DyTox \cite{douillard2022dytox}-\textbf{SumFeat}  & \textcolor{cyan}{\textbf{88.84}}   & -0.76  & 80.57  & \textcolor{cyan}{\textbf{89.45}} & \textcolor{cyan}{\textbf{94.04}}  & -1.20  & 79.62  & \textcolor{cyan}{\textbf{94.08}} \\

\hline
 \multirow{3}[0]{*} {Joint Training} &  CNNDet \cite{wang2020cnn}-Binary & 85.29  & NA  & NA  & 91.34 & 93.17  & NA  & NA  & 94.69 \\
&   CNNDet \cite{wang2020cnn}-Multi  &  84.63 & NA  & 70.59 & 90.10 & 92.30  & NA  & 78.71  & 94.82 \\
&  DyTox \cite{douillard2022dytox}-Multi  & 87.93   & NA  & 75.46  & 87.53 & 95.31  & NA  & 79.98  & 93.81 \\
    \end{tabular}
    }
        \vspace{-0.3cm}
      \caption{\scriptsize{Benchmarking results on CDDB's \emph{HARD} and \emph{LONG}. AA: Average detection Accuracy  , AF: Average Forgetting, AA-M: Average Accuracy for recognition.  mAP: mean Average Precision. \textcolor{green}{green}: LRCIL, 
      \textcolor{blue}{blue}: iCaRL,
      \textcolor{red}{red}: LUCIR, \textcolor{cyan}{cyan}: DyTox. \textbf{Bold}: best results, \underline{Underline}: second best results.}}
  \label{tab:cddb_hard_long}
    \vspace{-0.3cm}
\end{table}
\begin{table}[t]
  \centering
  \scriptsize
\resizebox{1\linewidth}{!}{%
    \begin{tabular}{l|l||c|c|c|c|c|c|c|c}
            & & \multicolumn{8}{c}{Reduced Memory Budgets for \textbf{CDDB-HARD}} \\ \hline
  &  & \multicolumn{4}{c|}{1000}  & \multicolumn{4}{c}{500}  \\ \hline
Learning System  & Evaluated Method & AA & AF & AA-M & mAP  & AA & AF & AA-M & mAP \\\hline
\multirow{3}[0]{*} {Binary-class} 
 &  LRCIL \cite{pellegrini2020latent}-\textbf{Sigmoid*}   &  \textcolor{green}{\underline{75.61}} &  -6.35 & NA & \textcolor{green}{80.30}  &  \textcolor{green}{\underline{73.51}} &  -8.96 & NA & \textcolor{green}{\underline{80.36}} \\ 
 &  iCaRL  \cite{rebuffi2017icarl}-\textbf{Sigmoid*}  &  \textcolor{blue}{76.99} &  -10.28 & NA & \textcolor{blue}{85.46} &  \textcolor{blue}{72.91} &  -16.01 & NA & \textcolor{blue}{\underline{81.34}} \\ 

u &  LUCIR  \cite{hou2019learning}-\textbf{Sigmoid*}   &  \textcolor{red}{81.75} &  -6.12 & NA & \textcolor{red}{\underline{88.46}}  &  \textcolor{red}{78.99} & -9.43 & NA & \textcolor{red}{85.81}\\ 
 \hline
\multirow{4}[0]{*} {Multi-class} &  LRCIL \cite{pellegrini2020latent}   &  \textcolor{green}{\textbf{76.39}} &  -4.39 & 55.10 & \textcolor{green}{80.45}  &  \textcolor{green}{73.18} &  -9.01 & 46.69 & \textcolor{green}{79.90} \\ 
 &  iCaRL  \cite{rebuffi2017icarl}  &  \textcolor{blue}{72.37}  & -13.04  & 41.43 & \textcolor{blue}{85.25}  & \textcolor{blue}{71.75} &  -13.52 &42.07 & \textcolor{blue}{\textbf{83.61}}\\
 &  LUCIR \cite{hou2019learning}-\textbf{LinFC}   &  \textcolor{red}{81.57} & -3.09 & 65.96 & \textcolor{red}{85.97}  &  \textcolor{red}{78.57} & -6.97 & 59.87 & \textcolor{red}{84.87}\\ 
 &  DyTox \cite{douillard2022dytox}  & \textcolor{cyan}{\textbf{88.64}}   & -0.86  & 80.02  & \textcolor{cyan}{\underline{89.41}} & \textcolor{cyan}{87.27}  & -2.02  & 76.54  & \textcolor{cyan}{87.77} \\
 \hline
\multirow{8}[0]{*} {Multi-task} &  LRCIL \cite{pellegrini2020latent}-\textbf{SumLogit}  &  \textcolor{green}{75.14} &  -3.53 & 56.31 & \textcolor{green}{\underline{81.05}}  & \textcolor{green}{73.05} &  -9.08 & 49.99 & \textcolor{green}{\textbf{80.71}}\\ 
 &  iCaRL \cite{marra2019incremental}-\textbf{SumLogit}  &  \textcolor{blue}{\textbf{79.76}} &  -8.73 & 66.66 & \textcolor{blue}{\textbf{87.75}} &  \textcolor{blue}{\textbf{73.98}} & -14.50 & 58.44 & \textcolor{blue}{81.00}\\
 &  LUCIR \cite{hou2019learning}-\textbf{SumLogit}  &  \textcolor{red}{\textbf{82.53}} &  -5.34 & 72.00 & \textcolor{red}{88.14}  &  \textcolor{red}{\underline{79.70}} & -8.18 & 65.86 & \textcolor{red}{\textbf{88.08}}\\ 
  
 &  DyTox \cite{douillard2022dytox}-\textbf{SumLogit}  & \textcolor{cyan}{87.39}  & -1.375  & 78.60  & \textcolor{cyan}{88.77} & \textcolor{cyan}{\underline{87.64}}  & -1.92  & 77.65  & \textcolor{cyan}{\textbf{88.14}} \\
  
 &  LRCIL \cite{pellegrini2020latent}-\textbf{Max}  & \textcolor{green}{74.52} & -5.18 & 43.65 & \textcolor{green}{\textbf{81.39}} & \textcolor{green}{\textbf{74.01}} & -8.62 & 49.59 & \textcolor{green}{78.72} \\ 

 &  iCaRL \cite{marra2019incremental}-\textbf{Max}   &  \textcolor{blue}{\underline{78.55}}  &  -8.69  & 65.34 & \textcolor{blue}{\underline{86.78}} & \textcolor{blue}{\underline{73.70}} & -14.65 &58.85 & \textcolor{blue}{81.06}\\ 
 &  LUCIR \cite{hou2019learning}-\textbf{SumFeat}    &  \textcolor{red}{\underline{82.35}} &  -4.76 & 71.22 & \textcolor{red}{\textbf{88.93}}  &  \textcolor{red}{\textbf{80.77}} &  -7.85 & 70.03 & \textcolor{red}{\underline{87.84}}\\ 
 & DyToX \cite{douillard2022dytox}-\textbf{SumFeat}  & \textcolor{cyan}{\underline{88.59}} & -1.27  & 80.57  & \textcolor{cyan}{\textbf{89.70}} & \textcolor{cyan}{\textbf{87.73}}  & -1.80  & 77.88  & \textcolor{cyan}{\underline{87.89}} \\
  
  \end{tabular}
  }
    \vspace{-0.3cm}

    \caption{\scriptsize{ Benchmarking results on reduced memories for the suggested CDDB's \emph{HARD} evaluation. \textcolor{green}{green}: LRCIL, 
      \textcolor{blue}{blue}: iCaRL,
      \textcolor{red}{red}: LUCIR, \textcolor{cyan}{cyan}: DyTox. \textbf{Bold}: best results, \underline{Underline}: second best results.}}
  \label{tab:cddb_memory_main}
    \vspace{-0.5cm}
\end{table}

\begin{table}[t]
  \centering
   \resizebox{1\linewidth}{!}{%

    \begin{tabular}{l|l||c|c|c|c|c|c|c|c}
            & & \multicolumn{8}{c}{100 Memory Budgets for \textbf{DyTox}} \\ \hline
  &  & \multicolumn{4}{c|}{EASY}  & \multicolumn{4}{c}{HARD}  \\ \hline
Learning System  & Evaluated Method & AA & AF & AA-M & mAP  & AA & AF & AA-M & mAP \\\hline
Multi-class  &  DyTox \cite{douillard2022dytox}   & 91.99  & -6.33  & 74.22 & 89.41 & 84.53  & -4.26  & 70.08 & 85.45\\
 \hline
\multirow{3}[0]{*} {Multi-task} &  
  DyTox \cite{douillard2022dytox}-\textbf{SumLogit} & \textbf{93.58}  & -4.33  & 76.45   & 90.30 &  \textbf{86.38} & -3.38    & 68.94 & \underline{86.23}  \\
  
 &  DyTox \cite{douillard2022dytox}-\textbf{SumFeat}   &93.32  & -4.61  & 74.67 & \textbf{91.37} & 85.23  & -4.74  & 70.31 & 84.40\\
  
 &  DyTox \cite{douillard2022dytox}-\textbf{Max}  &  \underline{93.47}  & -4.71  & 71.24 &  \underline{90.71}  & \underline{86.35}  & -2.68  & 70.85 & \textbf{86.47} \\
  
  \end{tabular}
  }
      \vspace{-0.3cm}

    \caption{\scriptsize{Benchmarking results on the reduced memory (memory=100) for DyTox's \emph{HARD} and \emph{EASY} evaluation. \textbf{bold}: best results, \underline{underline}: second best results}}
  \label{tab:dytox_memory_main}
    \vspace{-0.5cm}
\end{table}

\noindent\textbf{Memory Budget.} As all the evaluated methods require a memory to utilize an exemplar set of old samples to address the forgetting problem, we evaluate the performance as a function of the memory budget.
Table \ref{tab:cddb_memory_main} summarizes the results for the HARD evaluation with memory budget being 1000 and 500. The results again demonstrates the suggested MTs mostly outperform their competitors in terms of AA, mAP, AA-M. The LUCIR-SumLogit and LUCIR-SumFeat generally performs the best. The memory reduction from 1500 to 500 result in clear drops in the terms of AAs, AA-Ms, AFs and mAPs, showing the memory budget is one of the most essential factors for the CDD problem. Table \ref{tab:dytox_memory_main} studies the performances of DyTox and its variants for the further reduced memory (memory=100). The results demonstrate that most MT variants of DyTox clearly outperform its MC variant when the memory is very limited.

\begin{figure}[t]
    \centering
    \includegraphics[width=0.43\linewidth]{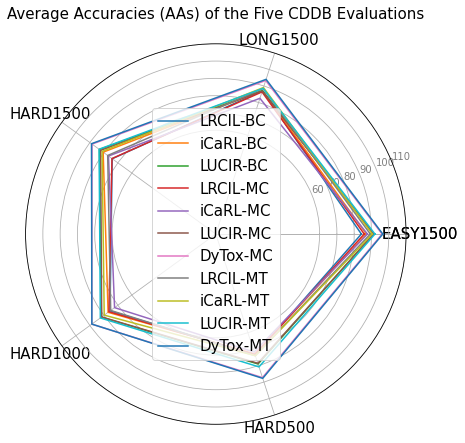}
    \includegraphics[width=0.4\linewidth]{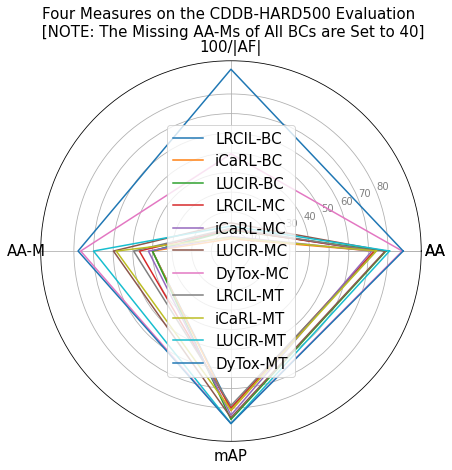}
    \vspace{-0.3cm}
\caption{\scriptsize{ \textbf{Left}: Radar plot (the bigger covering area the better) on AAs of the evaluated methods for CDDB's \emph{EASY1500, LONG1500, HARD1500, HARD1000, HARD500}. \textbf{Right}: Radar plot on AAs, AFs, mAPs and AA-Ms of the evaluated methods for \emph{HARD500}. HARDm: HARD with memory budget$=m$}, BC/MC/MT: binary-class/multi-class/multi-task learning methods that have the highest AAs/mAPs, AA: Average Accuracy  (detection), AF: Average Forgetting degree, AA-M: Average Accuracy (recognition), mAP: mean Average Precision, i.e., the mean of areas under the PR curve.}
  \label{fig:pr_curve}
    \vspace{-0.5cm}
\end{figure}

\noindent\textbf{Overall Spotlight.} Fig.\ref{fig:pr_curve} shows a radar plot of the five CDDB evaluations in terms of AAs, and a radar plot of the HARD500 evaluations in terms of the suggested four measures (i.e., AAs, AFs, mAP, AA-Ms). 
Our CDDB-HARD uses more advanced CIL method (DyTox) to reach the highest AA of around 86\% when memory=100 (/500), while the existing CDD benchmarks (GANfake \cite{marra2019incremental} and CoReD \cite{kim2021cored}) utilizes earlier CIL method (like iCaRL) to get highest AAs of above 96\% (memory=512) and 86\% (memory=0) respectively. The higher challenge is mainly attributed to the suggested CDD benchmark on the collection of known and unknown deepfakes, whereas they proposed performing CDD on either pure GAN-generated fakes or pure traditional deepfake-produced images. Besides, the rest three measures (i.e., AFs, mAP, AA-Ms) are overlooked by the two benchmarks, which however are valuable to study CDD. The low scores on them further imply that the suggested CDD benchmark is challenging and thus will open up promising directions for more solid research on the CDD problem.

%% file: 06_conclusion.tex
\section{Conclusion and Outlook}
\label{sec:con}
The continual deepfake detection benchmark proposed in this paper attempts to bring attention to the real world problem of detecting evolving deepfakes. In this front, difficult cases of deepfakes, long-term aspects of continual learning, and the varieties of deepfake sources are considered. To invite novel solutions,  their evaluation protocols and several baseline methods are established by borrowing the most promising ones from the literature. The proposed dataset, evaluation protocol, and established baselines will allow researchers to quickly test their creative ideas and probe them against  the existing ones. Additionally, through our evaluations, we are able to provide empirical study to analyze and discuss the common practices of continual learning in the context of the CDD problem.

Our experiments show that the proposed  CDDB is clearly more challenging than the existing benchmarks.
Hence we believe it will open up promising new directions of solid research on continual deepfake detection. As future works, other essentials remain to be studied, including 1) exemplar selection, 2) knowledge distillation, and  3) data augmentation. Due to the nature of CDD, we recommend using the new collection of deepfakes as an open set. Accordingly, we welcome external contributions to include any newly appeared deefake resources with the benchmark to simulate the real-world scenarios.

%% file: 08_acknowledge.tex
\smallskip
\noindent\small{\textbf{Acknowledgments.}} \footnotesize{
This work was founded by the Singapore Ministry of Education (MOE) Academic Research Fund (AcRF) Tier 1 grant (MSS21C002). This work was also supported in part by the ETH Z\"urich Fund (OK), an Amazon AWS grant, and an Nvidia GPU grant. The authors thank Ankush Panwar for processing the WildDeepFake dataset.}

%% file: 07_supp.tex
\appendix
\small

\section{Supplementary Material}

In the supplementary material, we further present 
(1) illustration of the suggested approaches for the adaptation of class incremental learning (CIL) methods to the  continual deepfake detection problem (CDD), 
(2) training details of the evaluated methods on the proposed continual deepfake detection benchmark (CDDB),
(3) study on more essential components of CIL methods on the suggested CDDB,
(4) evaluation on the GANFake  \cite{marra2019incremental} dataset, (5) experiments on the generalization capability of the suggested CIL methods to unseen domains and corrupted test images,
and (6) more overviews of the suggested CDDB evaluations.

\subsection{Illustrations of Adapting CIL to CDD}

\begin{figure} 
  \centering
  \includegraphics[width=0.85\linewidth]{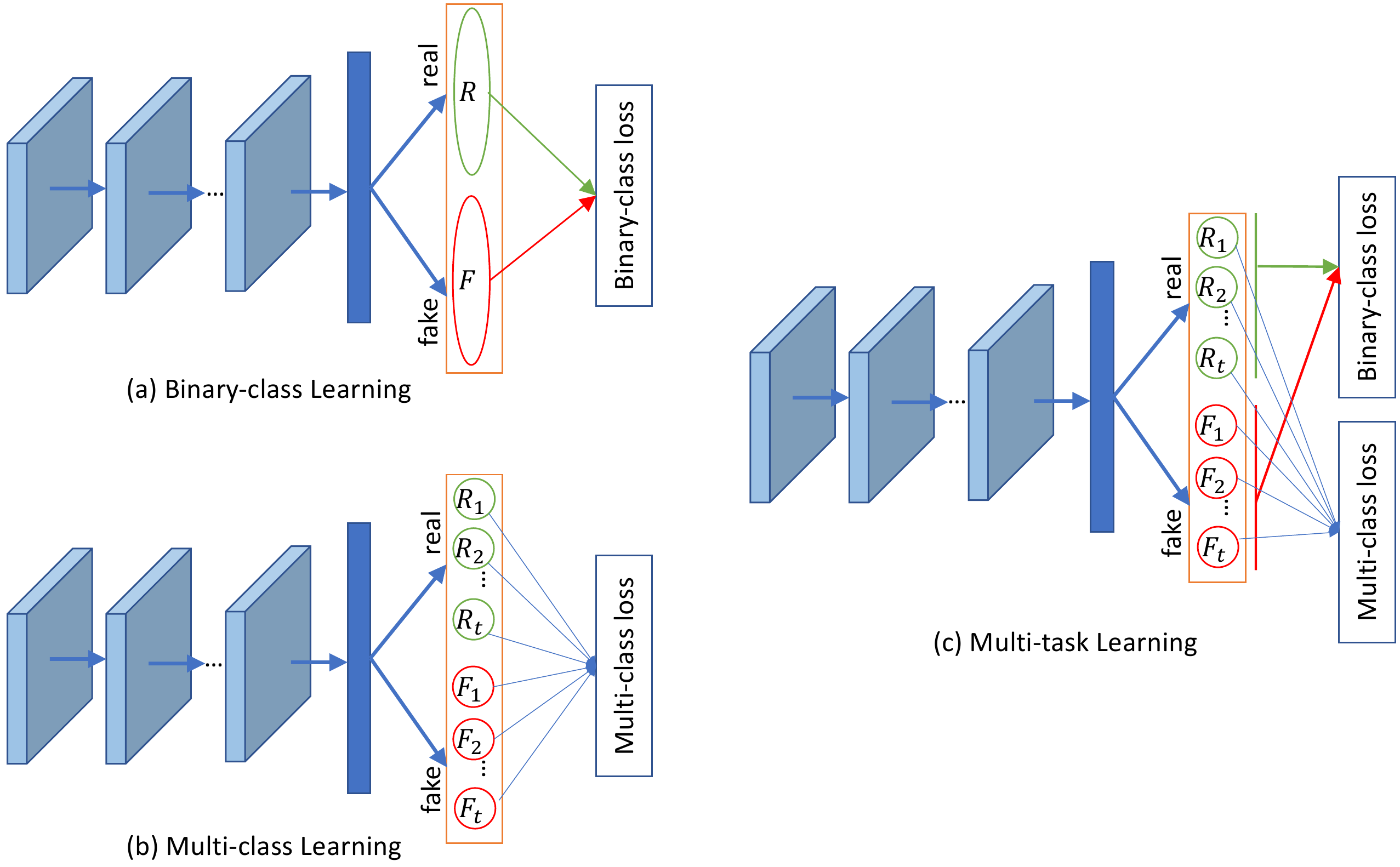}
  \caption{ 
     \scriptsize{Three main adaptions (a), (b), (c) of multi-class incremental learning for CDD. Here $R, F$ indicate the logits/features for reals and fakes respectively, and $t$ is the number of deepfake tasks.}}
  \label{fig:cil2bil}
  \vspace{-0.5cm}
\end{figure}

In the main paper, we study three main adaptations of CIL methods for CDD. Fig.\ref{fig:cil2bil} illustrates the suggested three approaches that adapts the exiting CIL methods to the context of CDD.

\subsection{Training Settings}

We evaluated three types of the state-of-the-art class incremental learning (CIL) methods on the suggested CDDB.

\noindent\textbf{Gradient-based Method.} We used the official code of the null space class incremental learning method (NSCIL) \cite{wang2021training}, which is one of the state-of-the-art gradient-based methods. We followed the official code's setup to tune the hyperparameter $\lambda$ from the default setup \{10, 30\} to the new settings \{100, 1\}, which is used to select the smallest singular values corresponding to the null space. The larger $\lambda$ leads to larger approximate null space, increasing the plasticity to learn new tasks while decreasing the memorization of old tasks \cite{wang2021training}.
We trained the NSCIL model for $12$ epochs in total, as we found that training more epochs for NSCIL resulted in performance degradation. The initial learning rate is $0.001$ for the first task and $0.0001$ for all other tasks and is divided by $2$ after $4$ and $8$ epochs for the  EASY evaluation. For batch normalization layers, the learning rate starts from $5 \times 10^{-5}$. The other parameters are the same as official NSCIL implementation.

\noindent\textbf{Memory-based Method.} We employed latent replay class incremental learning (LRCIL) \cite{pellegrini2020latent}, which is one of the most effective and efficient memory-based methods.
LRCIL was trained in the NSCIL's framework, with $20$ epochs in total, the initial learning rate is $0.001$ for the first task and $0.0001$ for all other tasks and is divided by $2$ after $10$ and $15$ epochs for the EASY, HARD, LONG evaluations. For batch normalization layers, the learning rate starts from $5 \times 10^{-5}$.
For the GANfake evaluation, we used $60$ epochs instead and the initial learning rate is divided by $2$ after $20$ and $40$ epoch.
Following original LRCIL's implementation, there is no knowledge distillation loss. We also tried to add distillation loss with a factor $\gamma_d = 0.3$. And we chose the second layer as latent layer. 
We used Adam with the batch size $32$ to train the network.

\noindent\textbf{Distillation-based Method.} We utilized five state-of-the-art distillation-based methods, i.e., incremental classifier and representation learning (iCaRL) \cite{rebuffi2017icarl}, Learning a Unified Classifier Incrementally via Rebalancing (LUCIR) \cite{hou2019learning}, Dynamic Token Expension (DyTox) \cite{douillard2022dytox}, as well as Compositional Class Incremental Learning (CCIL) \cite{mittal2021essentials}, which is additionally evaluated in the Supp. Material. 
Besides, we also evaluated the multi-task variant of iCaRL (iCaRL-SumLog) \cite{marra2019incremental}.\\
For iCaRL, we trained its model for $30$ epochs in total. The learning rate starts from $0.0001$ and is divided by $\frac{10}{3}$ after $10$ and $20$ epochs for the EASY, HARD and LONG evaluations.
For GANfake sequences, we used $60$ epochs instead and the initial learning rate $0.001$ is divided by $\frac{10}{3}$ after $20$ and $40$ epoch.
Following the original iCaRL's implementation, $\gamma_d$ was set to $1$ for knowledge distillation loss $\ell_{distill}$ with temperature as $T=1$.
We used Adam with the batch size $32$ to train the network. 
\\
For CCIL, we used its official code\footnote{\url{https://github.com/sud0301/essentials_for_CIL}}. The initial learning rate, total epoches and learning rate decay are the same as iCaRL. We tried the method with ($\gamma_d = 0.3$) or without ($\gamma_d = 0$) knowledge distillation $\ell_{distill}$. Also, we tried it with or without applying mixup and label smoothing techniques to the whole training process which is different from the original implementation (only for the first training phase), and we followed the relevant parameters (e.g, mixup weight). We used Adam with the batch size $32$ to train the network.
The other parameters are the same as the  official  implementation of CCIL.
\\ For LUCIR, the learning rate starts from $0.0001$ and is divided by $10$ after $16$ and $33$ epochs ($50$ epochs in total) for the LONG evaluation.
For the EASY, HARD and GANfake evaluations, we used $40$ epochs and the learning rate is divided by $10$ after $13$ and $26$ epoch. $\gamma_d$ is set to $0.5$ for knowledge distillation loss $\ell_{distill}$. $\gamma_m$ was set to $0.1$ for supplementary loss $\ell_{supp}$. 
Within $\ell_{supp}$, the number of closest class embeddings $J$ was set to $2$ and the threshold $\tau$ was set to $0.2$ for all experiments. The other parameters are the same as official LUCIR implementation. We used Adam with the batch size $32$ to train the LUCIR network.
\\ 
For DyTox \cite{douillard2022dytox}, we used its official implementation\footnote{\url{https://github.com/arthurdouillard/dytox}} with the default settings on its hyperparameters.
DyTox \cite{douillard2022dytox} originally suggests training its backbone from scratch. However, this case's performance (about 56\%) is much lower than that of training from the ImageNet pre-trained base-ConViT\footnote{We empirically find that the performance (AA=96.14\%) of ImageNet pre-trained ConViT is very comparable to that (AA=96.29\%) of the one that is further fine-tuned on ProGAN for the EASY1500 evaluation. Therefore, for the transformer-based networks, we suggest only using ImageNet pre-trained models without the further warm-up step on ProGAN.} on CDDB-Hard500. Therefore, we apply the pre-trained base-ConViT for DyTox throughout the paper.

\begin{table*}[t]
 \tiny
  \centering
     \resizebox{0.65\linewidth}{!}{%

    \begin{tabular}{l|l||c|c|c|c|c}
& & \multicolumn{3}{c}{\textbf{CDDB-EASY1500}} & &  \\ \hline
Learning System & Evaluated Method  & 0.1  & 0.3   & 0.5 & 0.7 & 0.9 \\ \hline

\multirow{12}[0]{*}{Multi-task (MT) learning} & LRCIL\cite{pellegrini2020latent}-SumLog\cite{marra2019incremental}  & \textcolor{green}{\textbf{89.54}}  & 88.76  & \textcolor{green}{86.51} & 87.08 & 87.20  \\
 & iCaRL\cite{rebuffi2017icarl}-SumLog\cite{marra2019incremental}  & \textcolor{blue}{88.97}  & 87.05 & \textcolor{blue}{87.51} & 87.39 & 86.29  \\
& LUCIR\cite{hou2019learning}-SumLog\cite{marra2019incremental}  & \textcolor{red}{91.65} & 91.56 & \textcolor{red}{\textbf{92.05}} & 91.39 & 91.08  \\

& LRCIL\cite{pellegrini2020latent}-\textbf{SumLogit}  & \textcolor{green}{\underline{89.38}}  & \textcolor{green}{\underline{89.33}} & \textcolor{green}{88.08} & 88.52 & 87.71  \\

 &iCaRL\cite{rebuffi2017icarl}-\textbf{SumLogit}  & \textcolor{blue}{88.52}  & \textcolor{blue}{\underline{90.43}} & \textcolor{blue}{90.37} & 87.88 & 87.57  \\
& LUCIR\cite{hou2019learning}-\textbf{SumLogit}  & \textcolor{red}{91.48}  & \textcolor{red}{\underline{92.00}} & \textcolor{red}{\underline{91.90}} & 91.11 & 91.67  \\

& LRCIL\cite{pellegrini2020latent}-\textbf{SumFeat}  & 88.23 & 87.81 & 88.33 & 87.68 & 86.27  \\

&iCaRL\cite{rebuffi2017icarl}-\textbf{SumFeat}  & 87.37  & 87.74 & 86.26 & 86.52 & 85.46  \\
& LUCIR\cite{hou2019learning}-\textbf{SumFeat}  & 91.75 & 91.81 & 91.03 & 91.28 & 91.48  \\

 & LRCIL\cite{pellegrini2020latent}-\textbf{Max}  & \textcolor{green}{89.17}  & 89.16 & 87.83 & \textcolor{green}{88.27} & 85.66  \\
 &iCaRL\cite{rebuffi2017icarl}-\textbf{Max}   & \textcolor{blue}{\underline{90.11}} & 89.92 & 89.76 & \textcolor{blue}{\textbf{90.78}} & 87.11  \\
& LUCIR\cite{hou2019learning}-\textbf{Max}  & \textcolor{red}{91.30}  & 91.21 & 91.20 & \textcolor{red}{91.35} & 91.04  \\
    \end{tabular}
    }
 \vspace{0.1cm}
\caption{\scriptsize{Empirical study on different multi-task learning trade-off parameter $\lambda$ values for the suggested CDDB' EASY1500 evaluation. The reported results are Average Accuracies (AA) for continual deepfake detection. \textbf{Bold}: best \textcolor{green}{\textbf{green}}/\textcolor{blue}{\textbf{blue}}/\textcolor{red}{\textbf{red}} results, \underline{Underline}: second/third best \textcolor{green}{\underline{green}}/\textcolor{blue}{\underline{blue}}/\textcolor{red}{\underline{red}} results. The best/second-best/third-best LRCIL, iCaRL, LUCIR results are in \textcolor{green}{green}, \textcolor{blue}{blue}, and \textcolor{red}{red} respectively. \emph{Note that we aim at comparing all the colorized triplets (LRCIL, iCaRL, LUCIR)'s performances to see which triplets consistently perform well. Each triplet should be of the same MT variant and the same $\lambda$ value. }}}
  \label{tab:cddb_hyperpara}
\end{table*}

For a fair comparison, except for DyTox that is based on an ImageNet pre-trained transformer ConViT \cite{d2021convit}, we used the state-of-the-art deepfake CNN detector (CNNDet) \cite{wang2020cnn}, which applies a ResNet-50 pretrained on ImageNet \cite{deng2009imagenet} and ProGAN \cite{karras2018progressive} as the backbone for all the remaining methods over the proposed CDDB. For the multi-task (MT) learning variants of the evaluated CIL methods, the study on the trade-off hyperparameter $\lambda$ is presented in Table \ref{tab:cddb_hyperpara}. In Table \ref{tab:cddb_hyperpara}, we aim to compare all the colorized triplets (LRCIL, iCaRL, LUCIR)'s performances to see which triplets consistently perform well. Each triplet should be of the same MT variant and the same $\lambda$ value (e.g., LRCIL-SumLogit-0.3, iCaRL-SumLogit-0.3, LUCIR-SumLogit-0.3). By comparing these triplets in terms of their average performances on (LRCIL, iCaRL, LUCIR), we can see that the SumLogit case ($\lambda=0.3$) works the most promisingly on the CDDB-EASY1500 evaluation. Furthermore, it is not feasible to tune the hyperparameters on the test data in the real-world scenario. Accordingly, we used the same hyperparameter $\lambda=0.3$ for all the MT variants of LRCIL, iCaRL and LUCIR throughout the main paper.

\subsection{Study on More Essentials}

Table \ref{tab:cddb_easy_essentials2} studies more essentials in CIL on CDDB. They are \emph{cosine normalization} on linear fully connected (FC) layer, \emph{label smoothing} and \emph{mixup} techniques. 

\noindent\textbf{Linear FC (LinFC).} The main paper finds that performing cosine normalization on FC (CosFC) layers result in LUCIR \cite{hou2019learning} and CCIL \cite{mittal2021essentials} is clearly inferior performance than using LinFC. This is because CosFC is originally designed to address the \emph{class imbalance issue}, which however is much less serious in the CDD context. We additionally apply LinFC to CCIL* \cite{mittal2021essentials} that further drops knowledge distillation. We can see that the LinFC version of CCIL* also performs better than its corresponding CosFC version, enhancing the conclusion that CosFC is too hard for class imbalance based normalization so that it really hurts the CDD performance.

\noindent\textbf{Label Smoothing (LabelSM).} We also study \emph{LabelSM} that is commonly used to improve generalization and reduce overconfidence of classification models \cite{szegedy2016rethinking,mittal2021essentials}. This technique affects the cross-entropy loss for classification by interpolating the one-hot labels with a uniform distribution over the possible classes \cite{mittal2021essentials}. From the results in Table \ref{tab:cddb_easy_essentials2}, we can discover that LabelSM slightly improves the AA scores in the cases of iCaRL \cite{rebuffi2017icarl} and LUCIR \cite{hou2019learning}, while hurting the AA scores in the case of LRCIL \cite{pellegrini2020latent} and CCIL \cite{mittal2021essentials}.

\begin{table*}[t]
 \tiny
  \centering
\resizebox{0.95\linewidth}{!}{%
    \begin{tabular}{l|l||c|c|c|c|c|c|c||c|c|c}
& & \multicolumn{6}{c}{\textbf{CDDB-EASY1500} } & &  \\ \hline
Learning System & Evaluated Method  & Task1  & Task2  & Task3   & Task4   & Task5 & Task6 & Task7 & AA & AF & \zhiwu{AA-M} \\ \hline
 \multirow{16}[0]{*}{\shortstack[l]{Multi-class (MC) \\ learning}}  &  LRCIL\cite{pellegrini2020latent}  & 83.50  & 77.88 & 90.84 & 98.90 & 84.75 & 98.86 & 65.92  & \textcolor{green}{85.81} & -5.88 & 67.11\\
 &iCaRL\cite{rebuffi2017icarl}  & 77.50 & 71.38 & 91.22 & 99.57 & 95.66 & 99.92 & 78.28 & \textcolor{blue}{87.65} & -9.41 & 65.39\\
 & LUCIR(CosFC)\cite{hou2019learning}  & 81.05  & 88.25 & 94.47 & 99.73 & 90.30 & 99.73 & 57.15  & \textcolor{red}{87.24} & -6.32 & 63.27\\
& CCIL(CosFC)\cite{mittal2021essentials} & 55.65 & 56.88 & 67.18 & 89.34 & 75.88 & 89.38 & 64.32 & \textcolor{brown}{71.23} & -22.01 & 39.53\\
 
& CCIL*(CosFC)\cite{mittal2021essentials} & 58.35 & 60.00 & 71.37 & 86.25 & 90.48 & 86.21 & 71.01 & \textcolor{brown}{74.81} & -22.56 & 48.66\\

& LUCIR\cite{hou2019learning}-\textbf{LinFC}  & 91.60  & 89.12 & 92.56 & 99.76 & 94.45 & 99.80 & 71.21  & \textcolor{red}{91.21} & -2.88 & 74.62 \\
& CCIL\cite{mittal2021essentials}-\textbf{LinFC} & 60.50  & 66.12 & 81.49 & 96.16 & 73.38 & 98.55 & 60.45 & \textcolor{brown}{76.66} & -16.50  & 41.85 \\
& CCIL*\cite{mittal2021essentials}-\textbf{LinFC} & 62.70  & 70.75 & 82.63 & 98.51 & 90.76 & 99.80 & 75.42 & \textcolor{brown}{82.94} & -13.24 & 61.95\\

& LRCIL\cite{pellegrini2020latent}-\textbf{LabelSM}  & 72.70 & 74.13 & 89.89 & 99.45 & 92.14 & 99.57 & 72.52  & \textcolor{green}{85.77} & -5.93 & 68.17 \\
& iCaRL\cite{rebuffi2017icarl}-\textbf{LabelSM}  & 79.55 & 80.62 & 90.46 & 99.84 & 89.37 & 99.84 & 80.17 & \textcolor{blue}{88.55} & -8.93 & 71.82 \\
& LUCIR\cite{hou2019learning}-\textbf{LabelSM} & 84.15 & 85.75 & 94.47 & 100.00 & 93.25 & 100.00 & 82.36 & \textcolor{red}{91.42} & -5.31 & 77.35 \\
& CCIL*\cite{mittal2021essentials}-\textbf{LabelSM} & 58.60 & 65.25 & 61.83 & 98.43 & 88.54 & 98.43 & 72.52 & \textcolor{brown}{77.66} & -18.16 & 53.85 \\

& LRCIL\cite{pellegrini2020latent}-\textbf{Mixup}  & 52.30 & 52.13 & 62.40 & 85.54 & 78.65 & 96.94 & 70.00 & \textcolor{green}{71.14} & -25.91 & 35.43 \\
&  iCaRL\cite{rebuffi2017icarl}-\textbf{Mixup}  & 78.05 & 84.88 & 94.47 & 100.00 & 94.09 & 100.00 & 80.47 & \textcolor{blue}{90.28} & -6.95 & 75.42 \\
& LUCIR\cite{hou2019learning}-\textbf{Mixup} & 82.31 & 87.22 & 90.70 & 96.32 & 92.98 & 93.21 & 70.59 & \textcolor{red}{87.62} & -6.54 & 70.93 \\
& CCIL*\cite{mittal2021essentials}-\textbf{Mixup} &  57.80 & 66.25 & 80.34 & 98.63 & 72.27 & 98.9 & 73.92 & \textcolor{brown}{78.30} & -18.98 & 55.16 \\

    \end{tabular}
    }
 \vspace{0.1cm}
\caption{\scriptsize{Benchmarking results on essentials of CIS methods on CDDB's \emph{EASY} evaluation. CCIL* indicates the CCIL method without using knowledge distillation loss. AA:  Average Accuracy for deepfake detection, AF: Average Forgetting degree, AA-M: Average Accuracy for deepfake recognition. The AA results of LRCIL, iCaRL, LUCIR, CCIL are in \textcolor{green}{green}, \textcolor{blue}{blue}, \textcolor{red}{red}, and \textcolor{brown}{brown} respectively.}}
  \label{tab:cddb_easy_essentials2}
\end{table*}

\noindent\textbf{Mixup.} We further employ \emph{Mixup} that is used a form of data augmentation for better clarification in general. The mixup technique is to generate training samples by linearly combining pairs of training samples (i.e., images and labels). Following \cite{mittal2021essentials}, we applied the mixup data to the training data and merely used them (no original training data) to train the evaluated methods. It is interesting to see that mixup brings a clear performance degradation in the case of LRCIL. This is because the mixup operations are performed in the space of raw images, which is not consistent with learning mechanism of LRCIL that performs rehearsal in the latent space of feature maps. Except for LRCIL and CCIL*, the other two methods iCaRL, LUCIR favor the mixup technique for further improvement in terms of AA on the suggested CDDB.

\begin{table}[t]
  \centering
\resizebox{0.95\linewidth}{!}{%

    \begin{tabular}{l|l||c|c|c|c}
            & & \multicolumn{4}{c}{ GANfake \cite{marra2019incremental}} \\ \hline
  &  & \multicolumn{2}{c|}{1024}   & \multicolumn{2}{c}{512} \\ \hline
Learning System  & Evaluated Method & AA & AF   & AA & AF \\ \hline
\multirow{3}[0]{*} {Multi-task learning} 
 &  iCaRL \cite{rebuffi2017icarl}-SumLog \cite{marra2019incremental} & 91.72 & -4.28 & 90.32 & -5.71 \\
&  iCaRL \cite{marra2019incremental}-\textbf{SumLogit} & 93.19  & -1.02 & 86.77   & -10.09 \\
&  LUCIR \cite{hou2019learning}-\textbf{SumLogit}   & 92.67  & -2.34  &  91.88  & -2.65 \\ 

    \end{tabular}
    }
    \vspace{0.1cm}
      \caption{\scriptsize{Continual GANfake detection accuracies of the main evaluated methods  on the GANfake dataset \cite{marra2019incremental} after the training on the last GAN using different memory budgets (1024 and 512).}}
  \label{tab:ganfake_memory}
\end{table}

\subsection{Evaluation on the GANfake Dataset}

For the GANfake \cite{marra2019incremental} dataset\footnote{\scriptsize{Since the GANfake \cite{marra2019incremental} dataset does not release the detailed train/val/test splits, we can merely evaluate the methods on our own splits following the description in the oiriginal paper. Accordingly, the reported results are not really comparable with those reported in \cite{marra2019incremental}}.}, Table \ref{tab:ganfake_memory} summarizes the evaluation results of the four main evaluated methods, i.e., the variant of iCaRL method \cite{marra2019incremental}, our suggested SumLogit variants of LRCIL \cite{pellegrini2020latent}, iCaRL \cite{rebuffi2017icarl} and LUCIR \cite{hou2019learning}. The results show that the suggested MT variants mostly work better than the original one (i.e., SumLog). From the overall results, we can find that the highest AA score\footnote{\scriptsize{The reported highest AA is 97.22 in GANfake \cite{marra2019incremental} when the memory budget is 1024. It is even higher than our reported one, while their train/val/test splits in \cite{marra2019incremental} are different from ours.}} (i.e., 93.19) for the case of memory budget=1024 on GANfake is clearly higher than the one (i.e., 82.53) on our suggested CDDB-HARD1000 (memory budget=1000). Besides, the highest AA score (i.e., 91.88) for the case of memory size=512 on GANfake is much higher than the one (i.e., 80.77) on the suggested CDDB-HARD evaluation with memory budget = 500. Moreover, the highest AA score (i.e., 93.19) on GANfake (memory size=1024) is even higher than the one (i.e., 92.00) of our easiest evaluation, i.e., CDDB-EASY1500 (memory size=1500). This implies that our suggested CDDB is clearly more challenging than GANfake, and therefore we believe it will has a high potential to promote more solid research in the domain of continual deepfake detection.

\subsection{Experiments on the Generalization Capability of Suggested Methods}

The knowledge accumulation nature of the suggested CDDB allows for a better generalization to deepfakes from unseen domains.
To verify this, we compare the adapted continual learning methods (e.g., LRCIL \cite{pellegrini2020latent}, iCaRL \cite{rebuffi2017icarl}, LUCIR \cite{hou2019learning}) against CNNDet \cite{wang2020cnn} (pre-trained on ProGAN) and CNNDet \cite{wang2020cnn}-Finetune (starts from ProGAN to the CDDB-Hard data stream). Due to the CDD setup, we cannot compare CNNDet trained directly on the whole data stream. 
Table \ref{tab:ood1} reports the results, showing the clearly better generalization of our adapted ones.
On the other hand, we also test the generalization ability of these methods to corrupted test images using the same strategy, i.e., Blur+JPEG (0.5), from \cite{wang2020cnn}. Note that \cite{wang2020cnn} suggests Blur+JPEG (0.5) for data augmentation during training, and testing on clean images. Despite no Blur+JPEG augmentation for training, our adapted ones (especially iCaRL) generally work better than the competitors (Table \ref{tab:ood2}).

\begin{table}[t]
\centering
\resizebox{1\linewidth}{!}{
\begin{tabular}{l|ccc||c|c}
\hline
& IMLE & CRN   &  Glow  & Average Acc ($\uparrow$) & mean Average Pre ($\uparrow$)  \\ \hline
CNNDet [66]    & 53.64  &   53.64 & 45.56  & 50.95 & 81.61\\
\hline
CNNDet [66]-Finetune    &  50.43 & 50.27  & 50.00  & 50.23 & 52.19 \\
\hline
LRCIL[51]-Adapted & 76.72   &  93.06 &  68.02   &  \textbf{79.28} & \textbf{87.78}\\
\hline
iCaRL[52]-Adapted   & 83.31  & 84.17 & 66.73  & \underline{78.07} &  82.31 \\
\hline
LUCIR[23]-Adapted & 70.02  & 93.30  &  70.90  & \underline{78.07} & \underline{86.39}\\
\hline

\end{tabular}
}
\caption{\scriptsize{Accuracies of generalization of the trained models on CDDB-Hard to 3 unseen domains (IMLE, CRN, Glow), which are of non-GANs and thus highly heterogeneous to the used CDDB-Hard data. \textbf{Bold}: best, \underline{Underline}: second best. \emph{Note that we have different test data on IMLE and CRN from CNNDet [66] that has no Glow data, the reported mean average precisions (mAPs) are somewhat different.}}} 
\label{tab:ood1}
\end{table}

\begin{table}[t]
\resizebox{1\linewidth}{!}{
\begin{tabular}{l|ccccc||c}
\hline
 & GauGAN       & BigGAN       & WildFake      & WhichFace &  SAN  & Average Acc ($\uparrow$)    \\ \hline
CNNDet &  53.80  & 58.00  & 61.50 &  48.96  &  63.33  & 57.12  \\
\hline
CNNDet[66]-Finetune &  50.95  & 50.25  &  52.73   &  48.25  &  71.11  & 54.66 \\
\hline
LRCIL[51]-Adapted &  61.20  & 61.75  & 56.86   &  59.75  & 48.89  & 57.69 \\
\hline
iCaRL[52]-Adapted & 67.35  & 64.38 &  66.36  &  69.75 & 56.67  & \textbf{64.90} \\
\hline
LUCIR[23]-Adapted &  62.45  & 61.00  &  56.42   & 56.75  &57.78  &  \underline{58.88} \\
\hline

\end{tabular}
}
\caption{\scriptsize{Accuracies of generalization to corrupted (Blur+JPEG) test images that are from the five tasks of CDDB-Hard. \textbf{Bold}: best, \underline{Underline}: second best}} 
\label{tab:ood2}
\end{table}

\begin{figure}[t]
    \centering
    
    \includegraphics[width=0.48\linewidth]{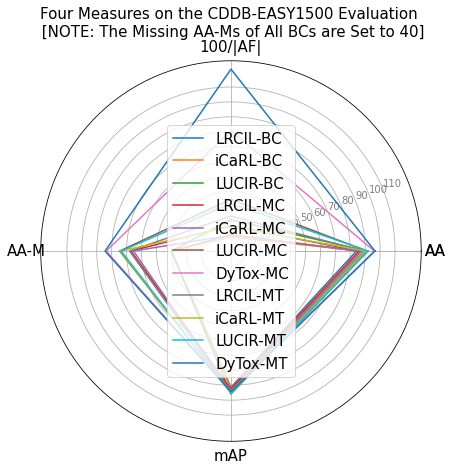}
    \includegraphics[width=0.48\linewidth]{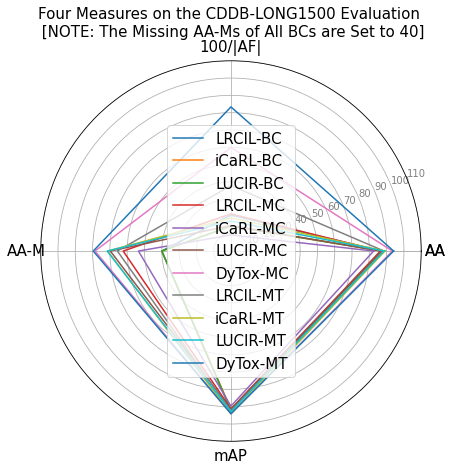}\\
    \includegraphics[width=0.48\linewidth]{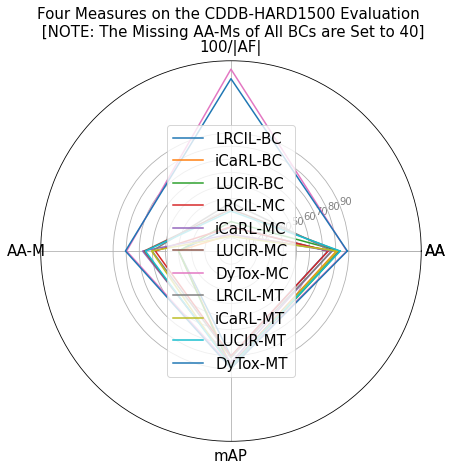}\\
    \includegraphics[width=0.48\linewidth]{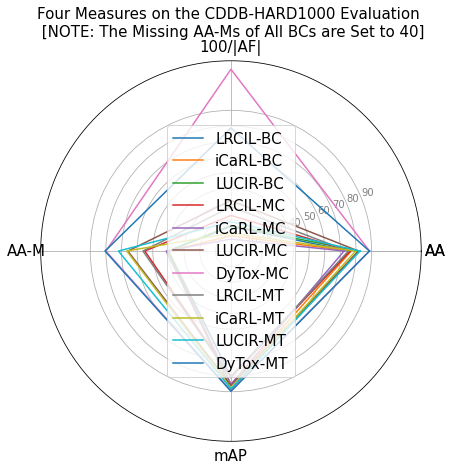}
    \includegraphics[width=0.48\linewidth]{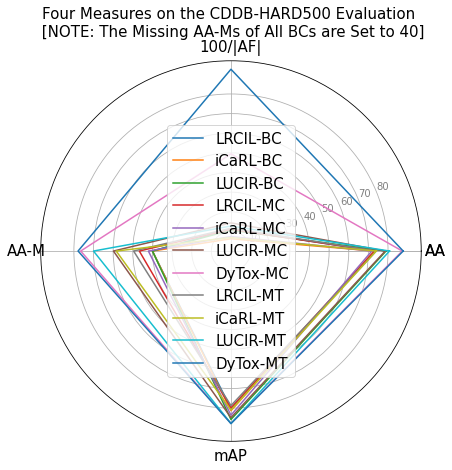}
  \caption{\scriptsize{Radar plots on AAs, AFs, mAPs and AA-Ms of the evaluated methods for \emph{EASY1500}, \emph{LONG1500}, \emph{HARD1500}, \emph{HARD1000}, and \emph{HARD500}, where 1500, 1000, 500 are memory budgets, BC/MC/MT: binary-class/multi-class/multi-task learning methods that have the highest AAs/mAPs, AA: Average Accuracy  (detection), AF: Average Forgetting degree, AA-M: Average Accuracy (recognition), mAP: mean Average Precision, i.e., the mean of areas under the PR curve.}}
  \label{fig:radar_plot2}
    \vspace{-0.2cm}
\end{figure}

\begin{figure}[t]
    \centering
     \includegraphics[width=0.49\linewidth]{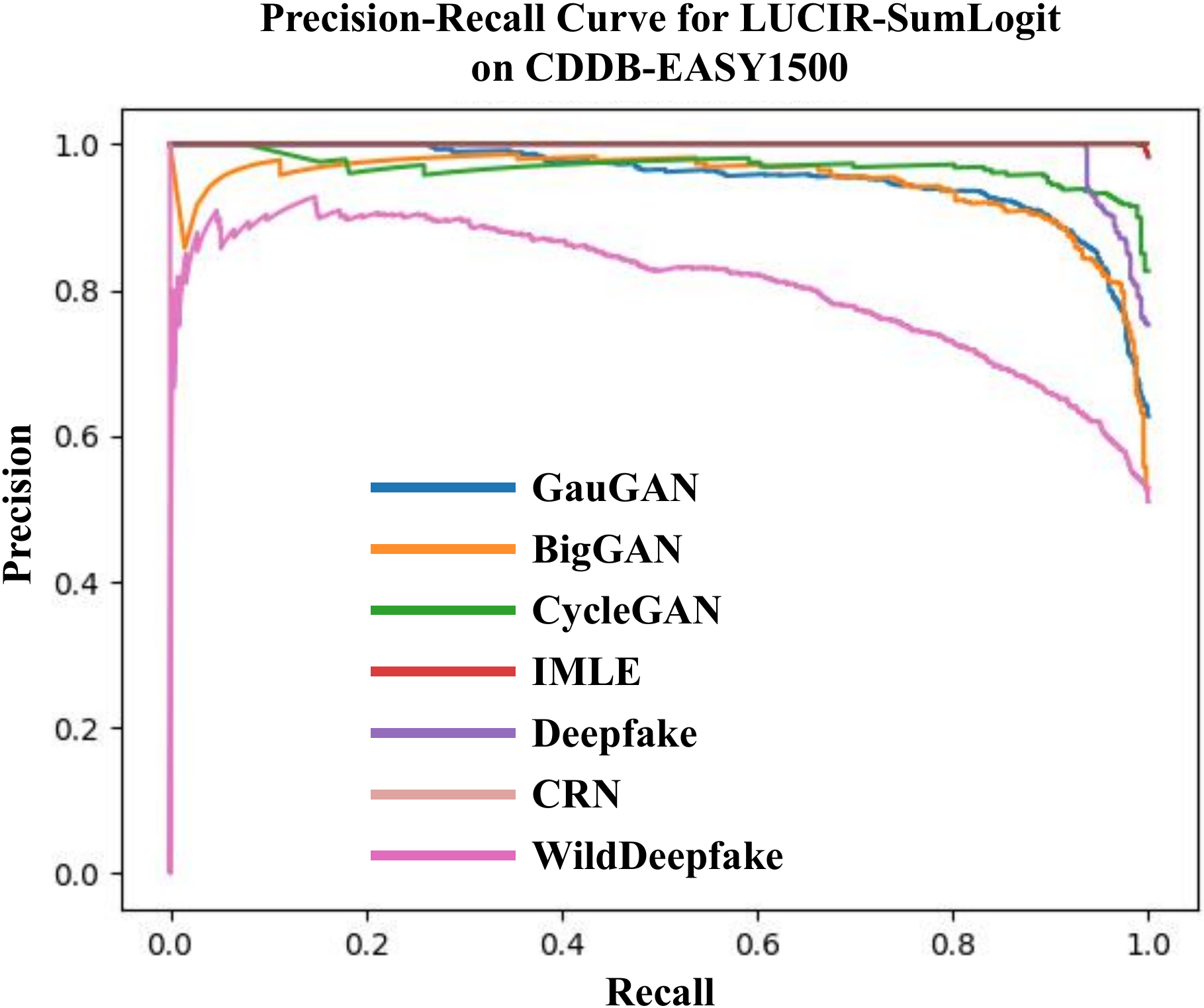}
    \includegraphics[width=0.49\linewidth]{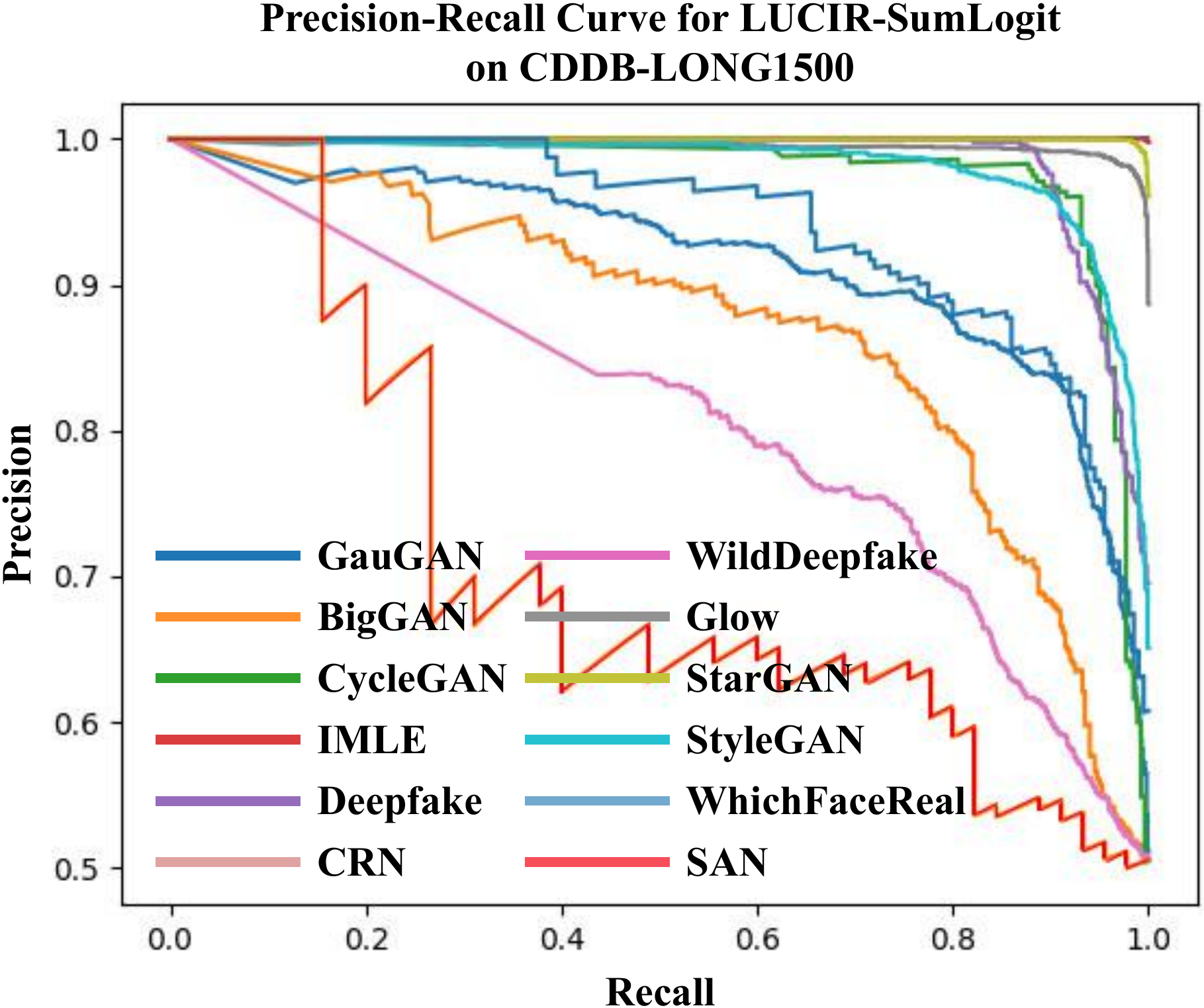}\\
    \includegraphics[width=0.49\linewidth]{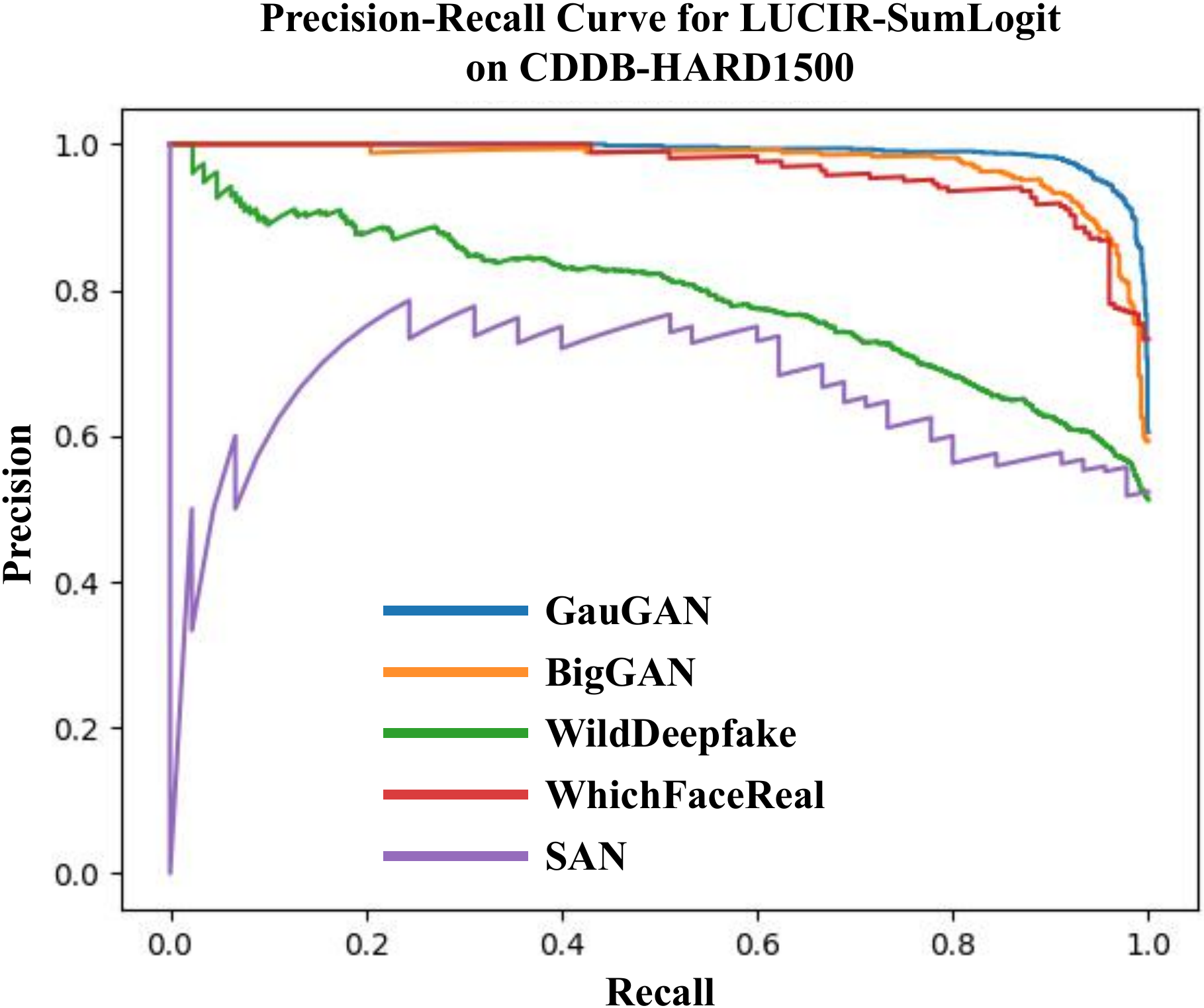} \\
    \includegraphics[width=0.49\linewidth]{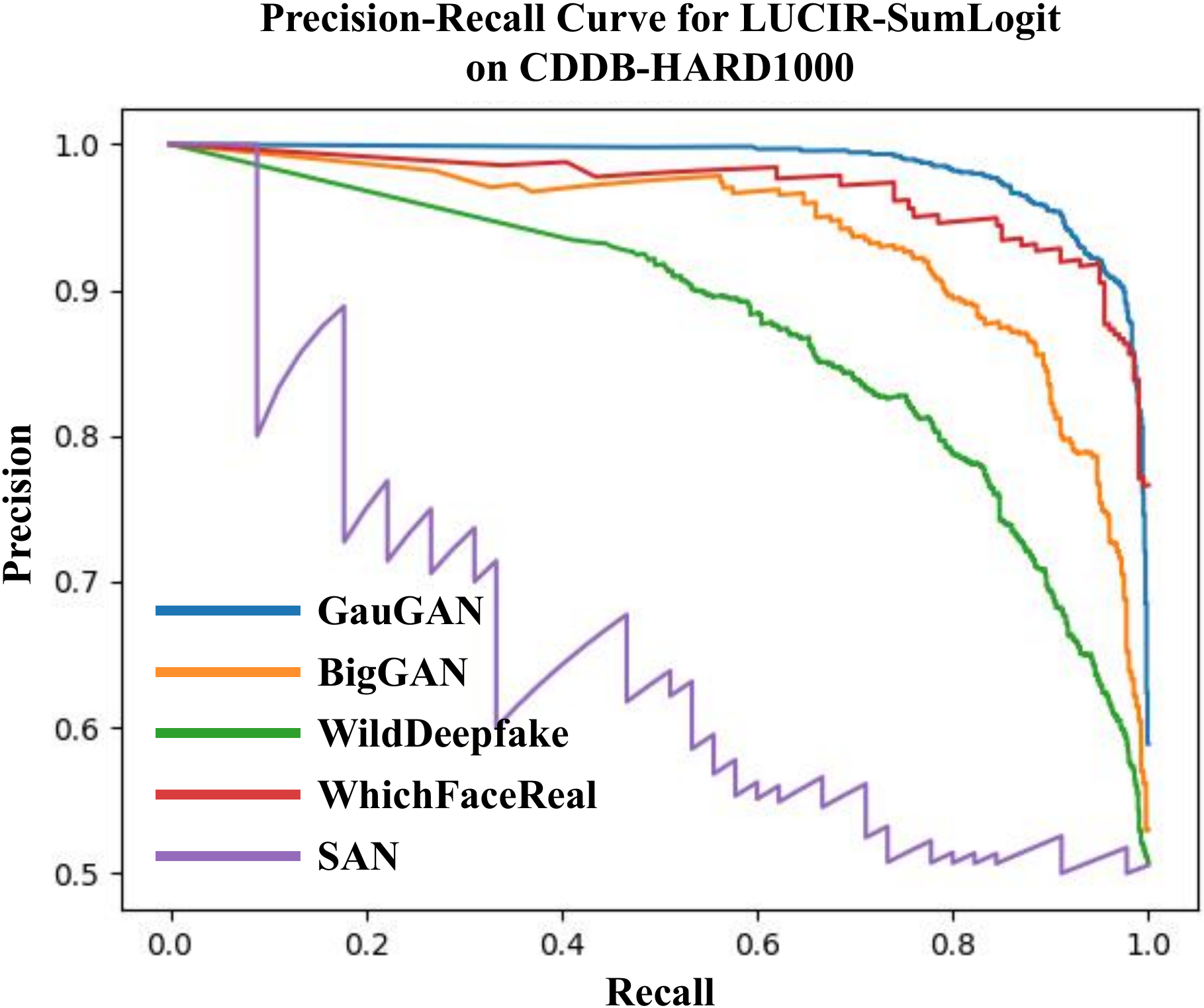} 
    \includegraphics[width=0.49\linewidth]{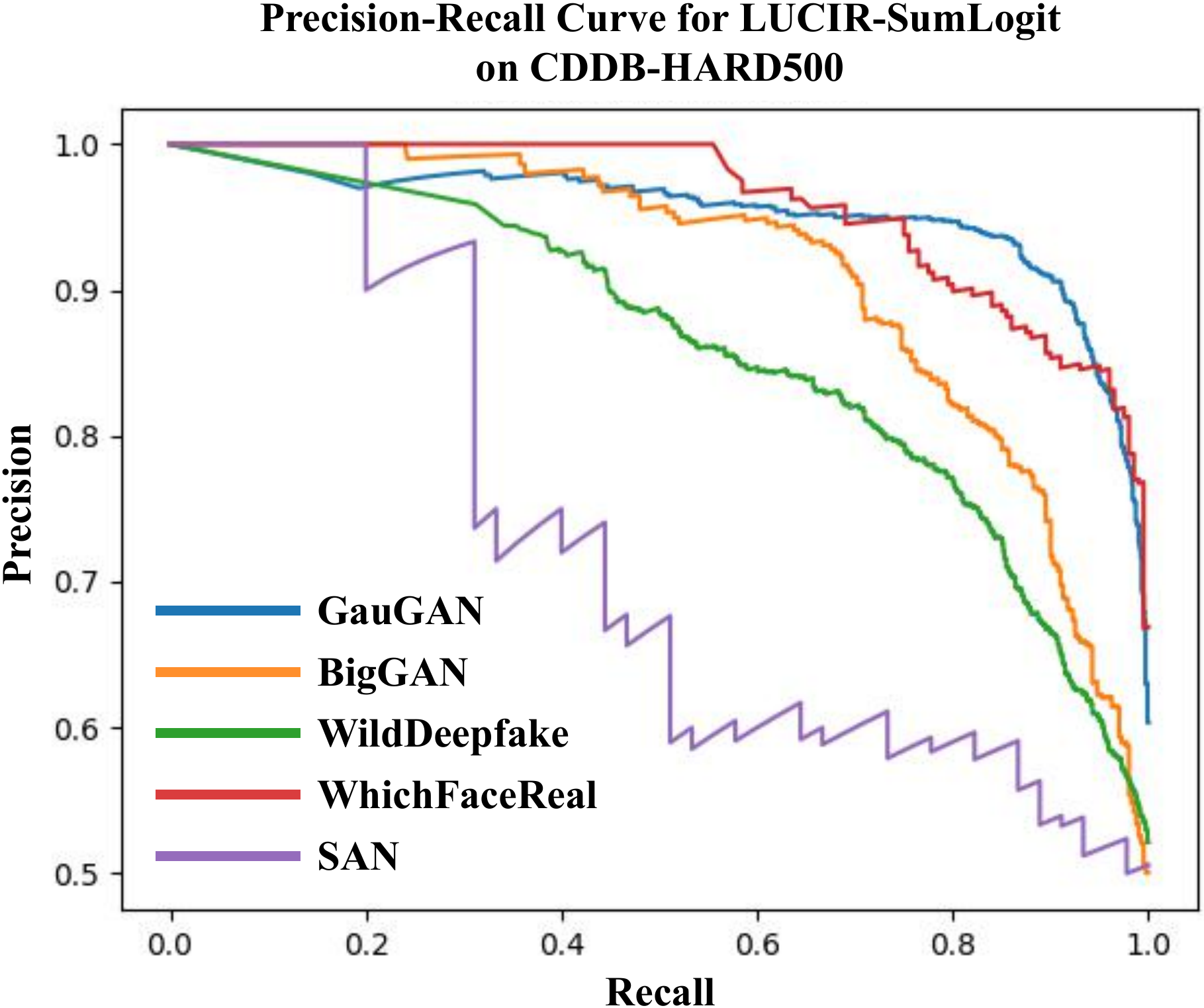} 
  \caption{\scriptsize{  Precision-Recall (PR) curves of LUCIR-SumLogit with the highest mAP for the five CDDB evaluations \emph{EASY1500}, \emph{LONG1500}, \emph{HARD1500}, \emph{HARD1000}, and \emph{HARD500}, where 1500, 1000, 500 are memory budgets, and mAP is mean Average Precision, i.e., the mean of areas under all the PR curves. }}
  \label{fig:pr_curve5}
    \vspace{-0.4cm}
\end{figure}

\begin{figure}[http!]
    \centering
     \includegraphics[width=0.49\linewidth]{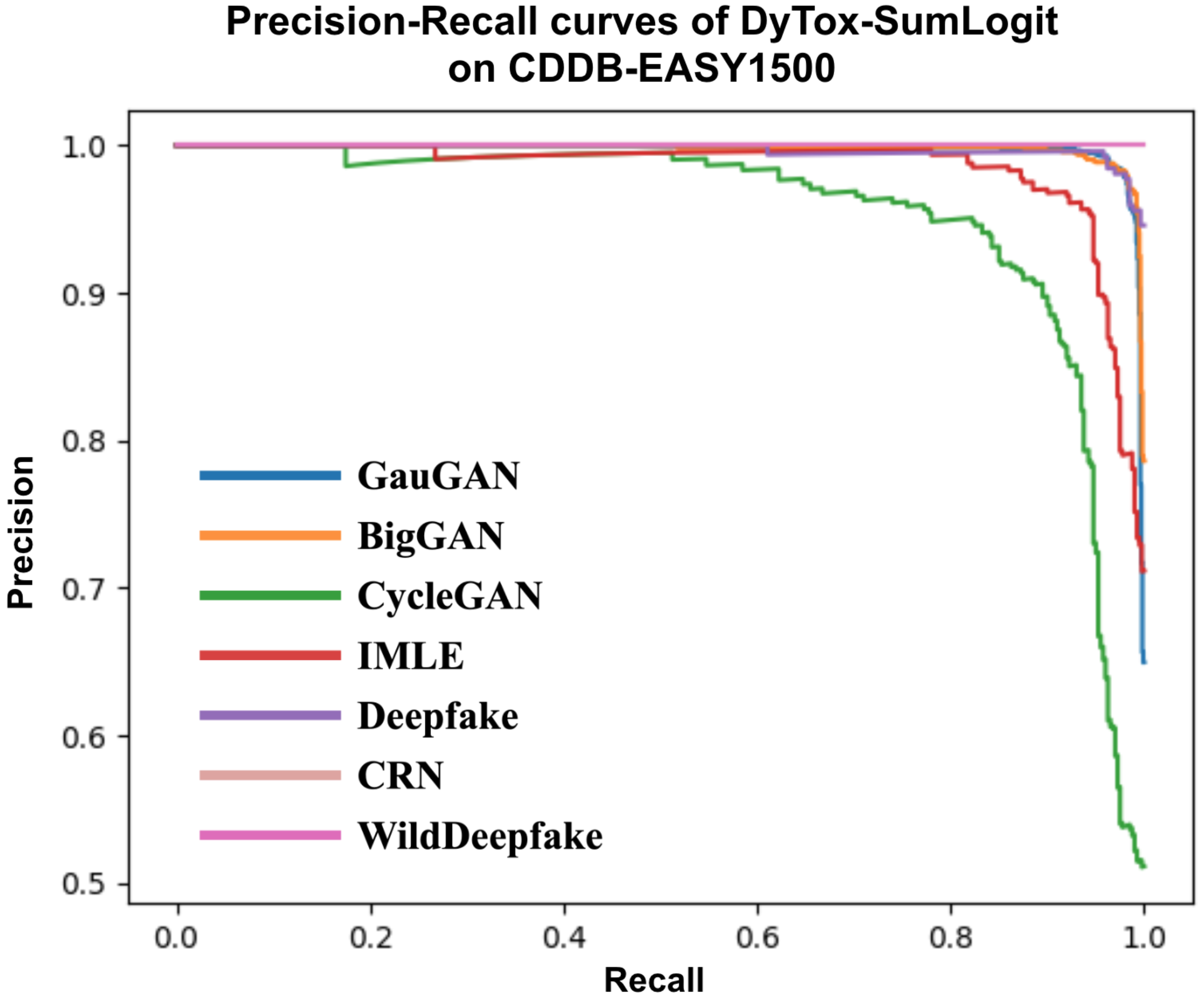}
    \includegraphics[width=0.49\linewidth]{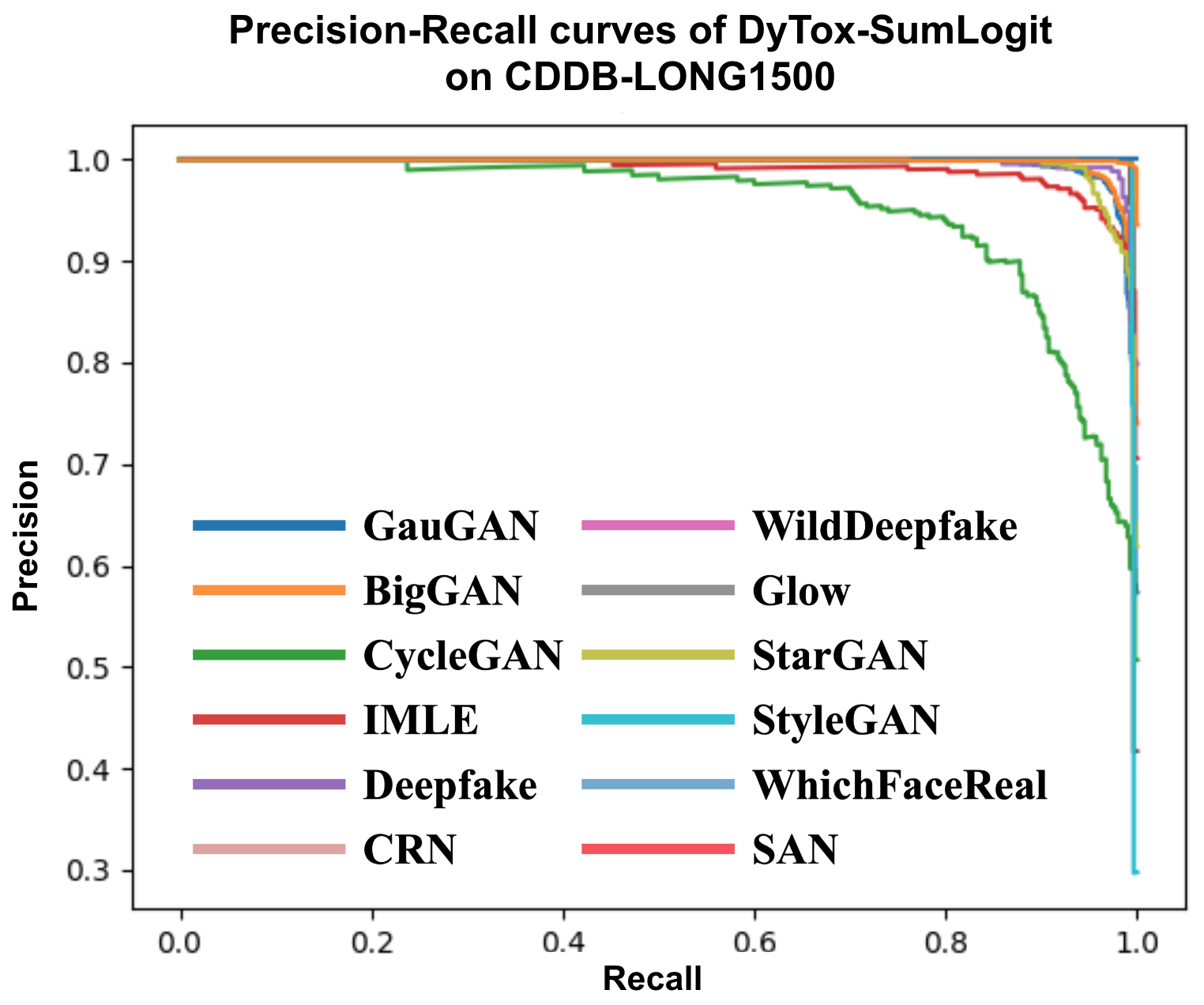}\\
    \includegraphics[width=0.49\linewidth]{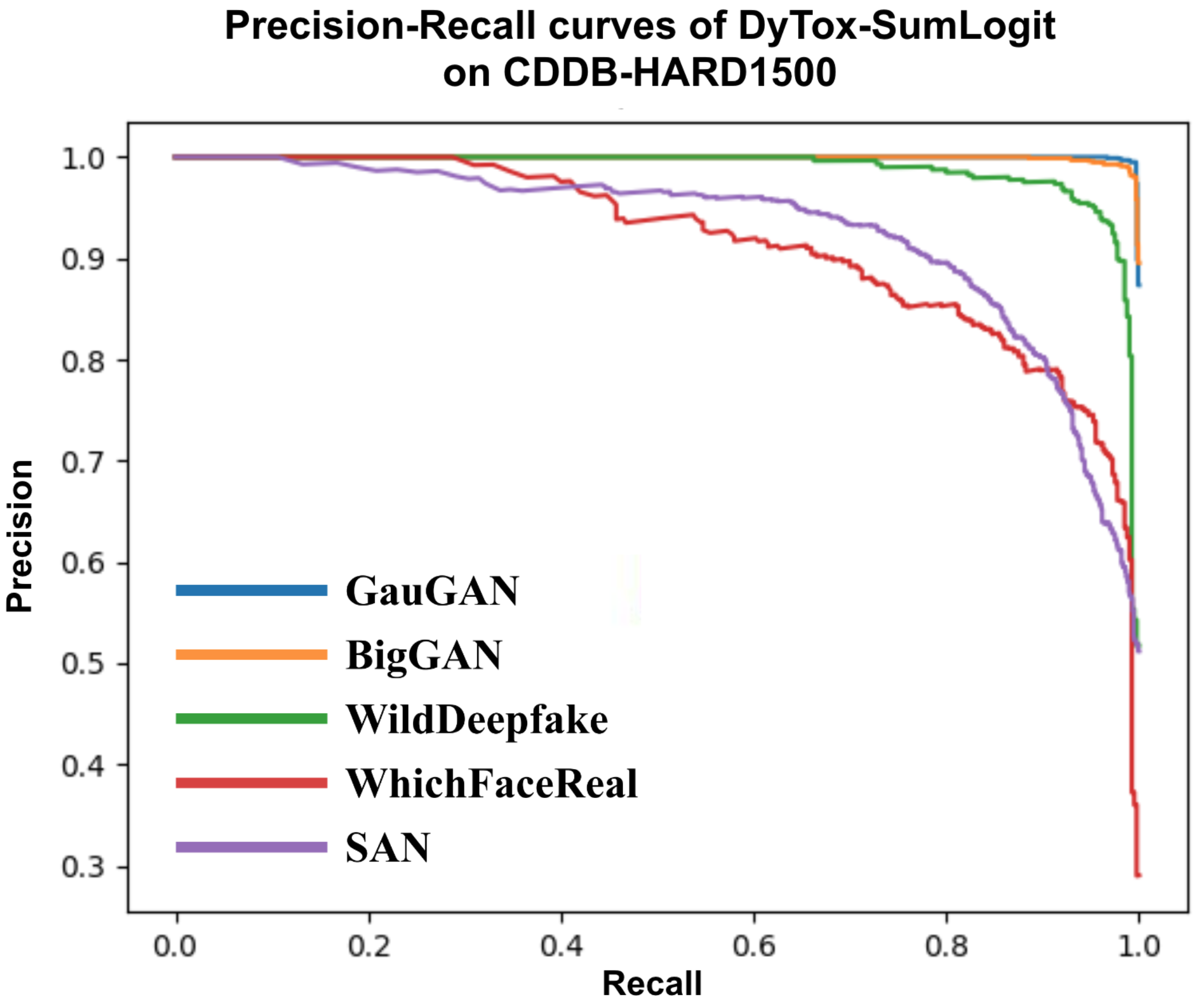} 
    \includegraphics[width=0.49\linewidth]{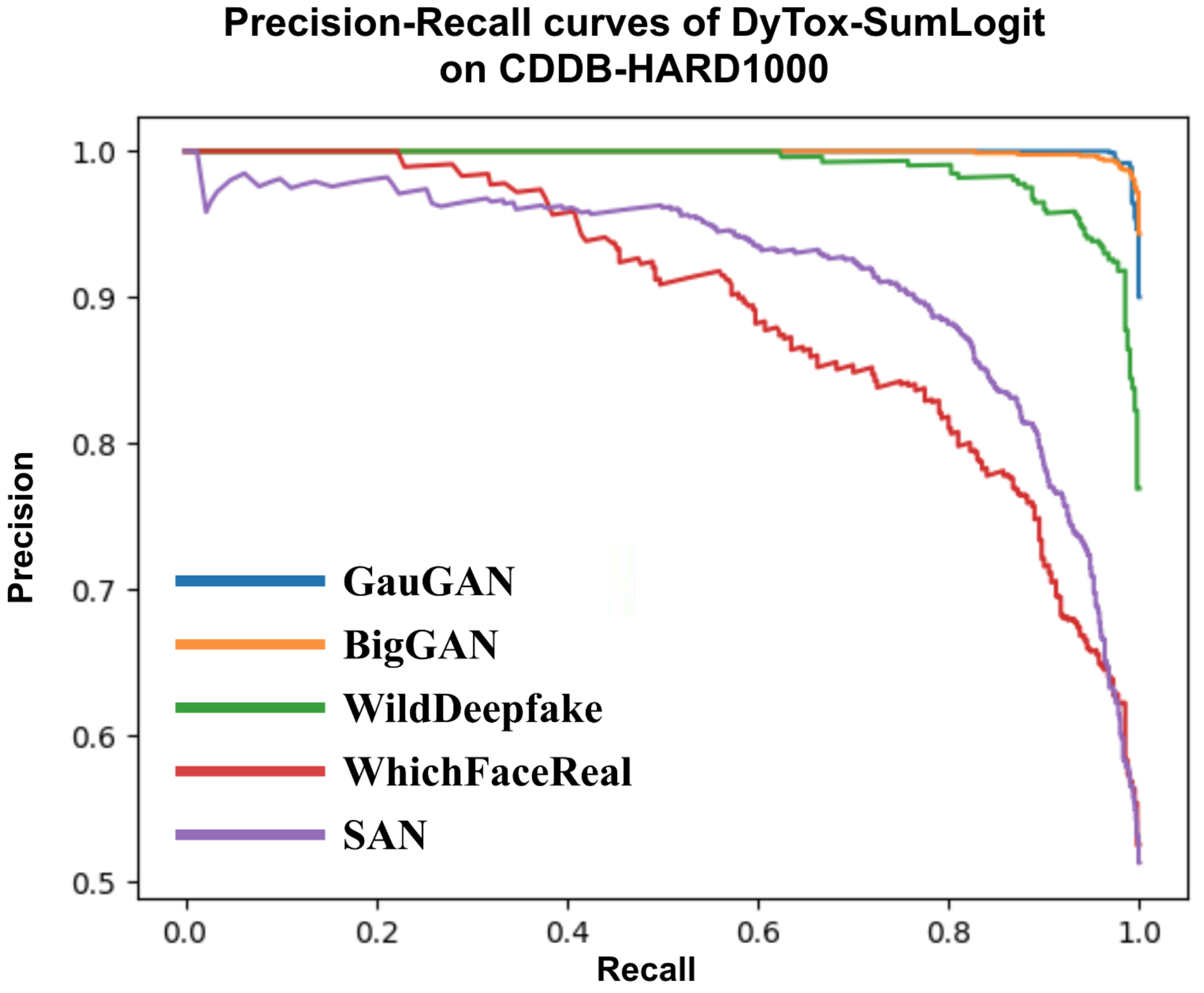} \\
    \includegraphics[width=0.49\linewidth]{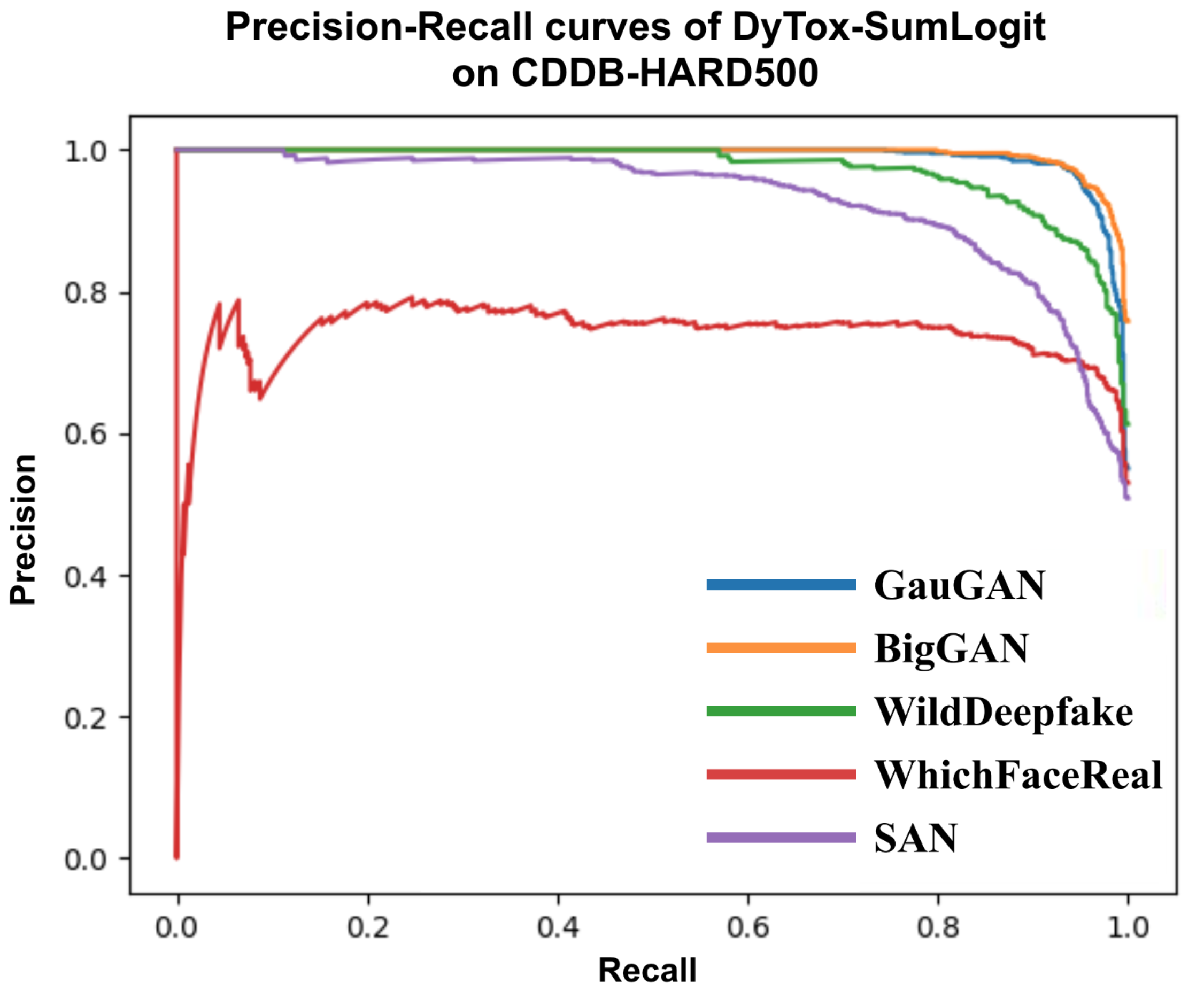} 
    \includegraphics[width=0.49\linewidth]{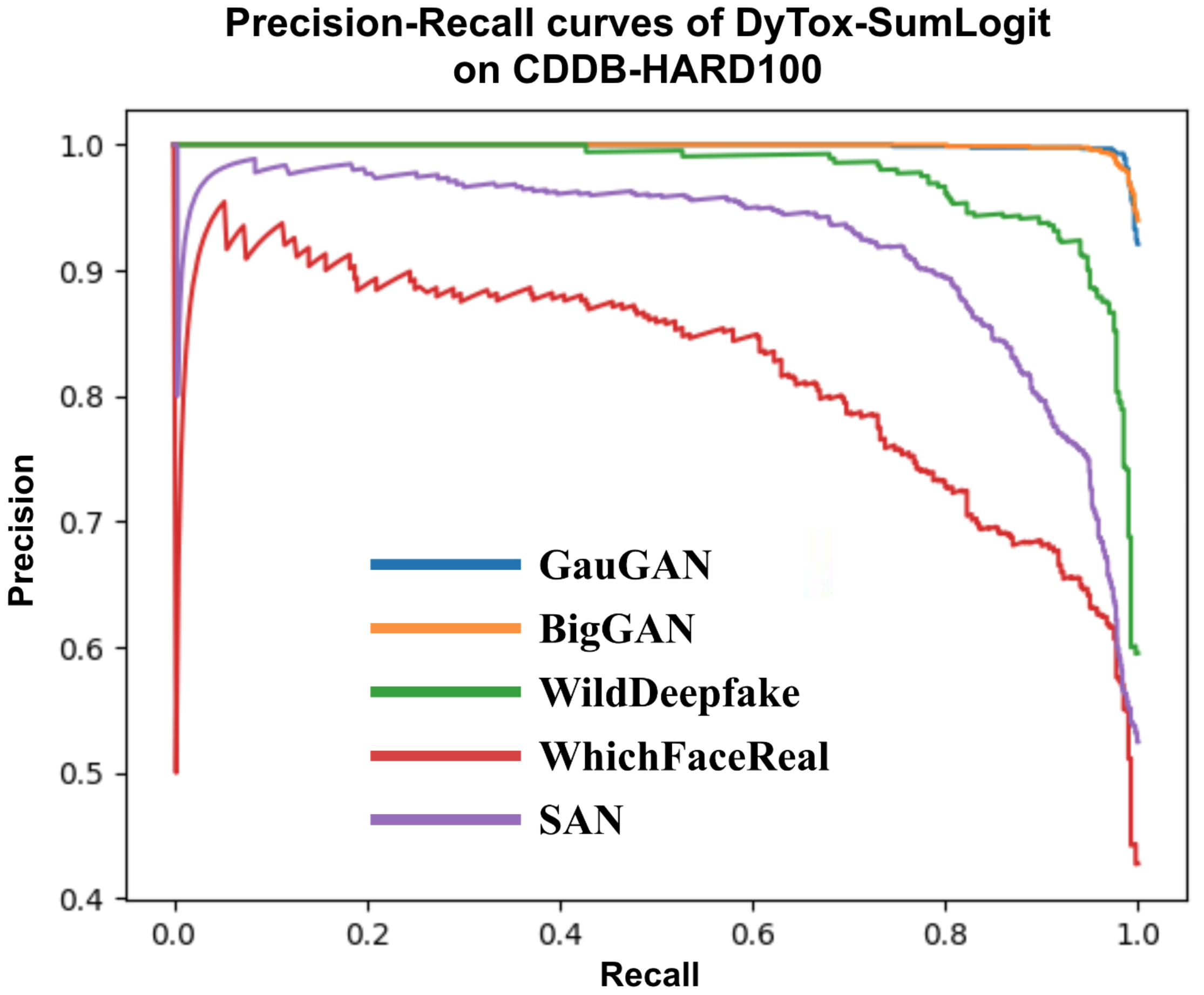}
  \caption{\scriptsize{Precision-Recall (PR) curves of DyTox-SumLogit with the highest mAP for the five CDDB evaluations \emph{EASY1500}, \emph{LONG1500}, \emph{HARD1500}, \emph{HARD1000}, and \emph{HARD500}, where 1500, 1000, 500 are memory budgets, and mAP is mean Average Precision, i.e., the mean of areas under all the PR curves.}}
  \label{fig:pr_curve5_dytox}
    \vspace{-0.2cm}
\end{figure}

\subsection{More Overviews of Five CDDB Evaluations}

Fig.\ref{fig:radar_plot2} presents radar plots on the four evaluation metrics, i.e., AAs, AFs, mAPs and AA-Ms, of the evaluated methods for \emph{EASY1500}, \emph{LONG1500}, \emph{HARD1500}, \emph{HARD1000}, and \emph{HARD500}, where 1500, 1000, 500 are three studied memory budgets, AA, AF, AA-M, mAP are Average Accuracy for deepfake detection, Average Forgetting degree, Average Accuracy for deepfake recognition, and mean Average Precision respectively. The mAP is calculated by the mean of areas under the precision-recall (PR) curves. For those multi-class and multi-task learning methods, we feed the PR calculation function with the normalized value of the maximum $\varphi(\cdot)$ activation output on fakes. Formally, the normalized value is $p_F = \frac{M_F}{M_F+M_R}$, where $M_R$ and $M_F$ are the maximum $\varphi(\cdot)$ activation values for reals and fakes respectively.

From the overall results in Fig.\ref{fig:radar_plot2}, we can see that LUCIR-MT (multi-task learning) generally works the best than the other evaluated methods in terms of AA, AF, AA-M and mAP for the five evaluations. By comparing the five different evaluations, HARD500's AA, AF, AA-M, mAP scores are clearly worse than the other four evaluations. This implies that the detection over HARD500 is the most challenging. In particular, it suffers from much more serious catastrophic forgetting due to the smaller memory budget. Fig.\ref{fig:pr_curve5} and Fig.\ref{fig:pr_curve5_dytox} present the PR curves of the LUCIR method and the DyTox method  for the five evaluations \emph{EASY1500}, \emph{LONG1500}, \emph{HARD1500}, \emph{HARD1000}, and \emph{HARD500}. The PR curves demonstrate that the most difficult tasks are generally the ones on SAN \cite{dai2019second}, WildDeepfake \cite{zi2020wilddeepfake}, and WhichFaceReal \cite{whichface} for the different evaluations. This is because SAN is a small data set, and WildDeepfake/WhichFaceReal consists of unknown deepfakes in the wild scenes.

%% file: main.bbl
\begin{thebibliography}{10}\itemsep=-1pt

\bibitem{faceswap2009}
Deepfakes faceswap github repository.
\newblock {\em https://github.com/MarekKowalski/FaceSwap}.

\bibitem{deepfakesgit}
Deepfakes github.
\newblock {\em https://github.com/deepfakes/faceswap}.

\bibitem{whichface}
Which face is real?
\newblock {\em http://www.whichfaceisreal.com}.

\bibitem{afchar2018mesonet}
Darius Afchar, Vincent Nozick, Junichi Yamagishi, and Isao Echizen.
\newblock Mesonet: a compact facial video forgery detection network.
\newblock In {\em WIFS}. IEEE, 2018.

\bibitem{agarwal2019protecting}
Shruti Agarwal, Hany Farid, Yuming Gu, Mingming He, Koki Nagano, and Hao Li.
\newblock Protecting world leaders against deep fakes.
\newblock In {\em CVPR workshops}, 2019.

\bibitem{aljundi2018memory}
Rahaf Aljundi, Francesca Babiloni, Mohamed Elhoseiny, Marcus Rohrbach, and
  Tinne Tuytelaars.
\newblock Memory aware synapses: Learning what (not) to forget.
\newblock In {\em ECCV}, 2018.

\bibitem{brock2018large}
Andrew Brock, Jeff Donahue, and Karen Simonyan.
\newblock Large scale {GAN} training for high fidelity natural image synthesis.
\newblock 2019.

\bibitem{castro2018end}
Francisco~M Castro, Manuel~J Mar{\'\i}n-Jim{\'e}nez, Nicol{\'a}s Guil, Cordelia
  Schmid, and Karteek Alahari.
\newblock End-to-end incremental learning.
\newblock In {\em ECCV}, 2018.

\bibitem{chaudhry2018efficient}
Arslan Chaudhry, Marc'Aurelio Ranzato, Marcus Rohrbach, and Mohamed Elhoseiny.
\newblock Efficient lifelong learning with a-gem.
\newblock {\em arXiv preprint arXiv:1812.00420}, 2018.

\bibitem{chaudhry2019continual}
Arslan Chaudhry, Marcus Rohrbach, Mohamed Elhoseiny, Thalaiyasingam Ajanthan,
  Puneet~Kumar Dokania, Philip~HS Torr, and Marc'Aurelio Ranzato.
\newblock Continual learning with tiny episodic memories.
\newblock {\em arXiv preprint arXiv:1902.10486}, 2019.

\bibitem{chen2017photographic}
Qifeng Chen and Vladlen Koltun.
\newblock Photographic image synthesis with cascaded refinement networks.
\newblock In {\em ICCV}, 2017.

\bibitem{chesney2019deep}
Bobby Chesney and Danielle Citron.
\newblock Deep fakes: A looming challenge for privacy, democracy, and national
  security.
\newblock {\em Calif. L. Rev.}, 107:1753, 2019.

\bibitem{choi2018stargan}
Yunjey Choi, Minje Choi, Munyoung Kim, Jung-Woo Ha, Sunghun Kim, and Jaegul
  Choo.
\newblock Stargan: Unified generative adversarial networks for multi-domain
  image-to-image translation.
\newblock In {\em CVPR}, 2018.

\bibitem{chollet2017xception}
Fran{\c{c}}ois Chollet.
\newblock Xception: Deep learning with depthwise separable convolutions.
\newblock In {\em CVPR}, 2017.

\bibitem{dai2019second}
Tao Dai, Jianrui Cai, Yongbing Zhang, Shu-Tao Xia, and Lei Zhang.
\newblock Second-order attention network for single image super-resolution.
\newblock In {\em CVPR}, 2019.

\bibitem{deng2009imagenet}
Jia Deng, Wei Dong, Richard Socher, Li-Jia Li, Kai Li, and Li Fei-Fei.
\newblock Imagenet: A large-scale hierarchical image database.
\newblock In {\em CVPR}, 2009.

\bibitem{deng2022incremental}
Jieren Deng, Jianhua Hu, Haojian Zhang, and Yunkuan Wang.
\newblock Incremental prototype prompt-tuning with pre-trained representation
  for class incremental learning.
\newblock {\em arXiv preprint arXiv:2204.03410}, 2022.

\bibitem{dinh2017density}
Laurent Dinh, Jascha Sohl-Dickstein, and Samy Bengio.
\newblock Density estimation using real {NVP}.
\newblock In {\em ICLR}, 2017.

\bibitem{dolhansky2020deepfake}
Brian Dolhansky, Joanna Bitton, Ben Pflaum, Jikuo Lu, Russ Howes, Menglin Wang,
  and Cristian~Canton Ferrer.
\newblock The deepfake detection challenge (dfdc) dataset.
\newblock {\em arXiv preprint arXiv:2006.07397}, 2020.

\bibitem{douillard2022dytox}
Arthur Douillard, Alexandre Ram{\'e}, Guillaume Couairon, and Matthieu Cord.
\newblock Dytox: Transformers for continual learning with dynamic token
  expansion.
\newblock In {\em CVPR}, 2022.

\bibitem{dufour2019contributing}
Nick Dufour and Andrew Gully.
\newblock Contributing data to deepfake detection research.
\newblock {\em Google AI Blog}, 2019.

\bibitem{d2021convit}
St{\'e}phane d’Ascoli, Hugo Touvron, Matthew~L Leavitt, Ari~S Morcos, Giulio
  Biroli, and Levent Sagun.
\newblock Convit: Improving vision transformers with soft convolutional
  inductive biases.
\newblock In {\em ICLR}, 2021.

\bibitem{farajtabar2020orthogonal}
Mehrdad Farajtabar, Navid Azizan, Alex Mott, and Ang Li.
\newblock Orthogonal gradient descent for continual learning.
\newblock In {\em AISTATS}, 2020.

\bibitem{gerstner2022detecting}
Candice~R Gerstner and Hany Farid.
\newblock Detecting real-time deep-fake videos using active illumination.
\newblock In {\em CVPR}, 2022.

\bibitem{goodfellow2014generative}
Ian~J Goodfellow, Jean Pouget-Abadie, Mehdi Mirza, Bing Xu, David Warde-Farley,
  Sherjil Ozair, Aaron Courville, and Yoshua Bengio.
\newblock Generative adversarial networks.
\newblock In {\em NeurIPS}, 2014.

\bibitem{he2016deep}
Kaiming He, Xiangyu Zhang, Shaoqing Ren, and Jian Sun.
\newblock Deep residual learning for image recognition.
\newblock In {\em CVPR}, 2016.

\bibitem{hinton2015distilling}
Geoffrey Hinton, Oriol Vinyals, Jeff Dean, et~al.
\newblock Distilling the knowledge in a neural network.
\newblock {\em arXiv preprint arXiv:1503.02531}, 2015.

\bibitem{hou2019learning}
Saihui Hou, Xinyu Pan, Chen~Change Loy, Zilei Wang, and Dahua Lin.
\newblock Learning a unified classifier incrementally via rebalancing.
\newblock In {\em CVPR}, 2019.

\bibitem{hu2018overcoming}
Wenpeng Hu, Zhou Lin, Bing Liu, Chongyang Tao, Zhengwei Tao, Jinwen Ma, Dongyan
  Zhao, and Rui Yan.
\newblock Overcoming catastrophic forgetting for continual learning via model
  adaptation.
\newblock In {\em ICLR}, 2018.

\bibitem{hung2019compacting}
Ching-Yi Hung, Cheng-Hao Tu, Cheng-En Wu, Chien-Hung Chen, Yi-Ming Chan, and
  Chu-Song Chen.
\newblock Compacting, picking and growing for unforgetting continual learning.
\newblock {\em NeurIPS}, 2019.

\bibitem{jiang2020deeperforensics}
Liming Jiang, Ren Li, Wayne Wu, Chen Qian, and Chen~Change Loy.
\newblock Deeperforensics-1.0: A large-scale dataset for real-world face
  forgery detection.
\newblock In {\em CVPR}, 2020.

\bibitem{kahardipraja2021towards}
Patrick Kahardipraja, Brielen Madureira, and David Schlangen.
\newblock Towards incremental transformers: An empirical analysis of
  transformer models for incremental nlu.
\newblock {\em arXiv preprint arXiv:2109.07364}, 2021.

\bibitem{karras2018progressive}
Tero Karras, Timo Aila, Samuli Laine, and Jaakko Lehtinen.
\newblock Progressive growing of {GANs} for improved quality, stability, and
  variation.
\newblock In {\em ICLR}, 2018.

\bibitem{karras2019style}
Tero Karras, Samuli Laine, and Timo Aila.
\newblock A style-based generator architecture for generative adversarial
  networks.
\newblock In {\em CVPR}, 2019.

\bibitem{kemker2018fearnet}
Ronald Kemker and Christopher Kanan.
\newblock Fearnet: Brain-inspired model for incremental learning.
\newblock In {\em ICLR}, 2018.

\bibitem{kim2021cored}
Minha Kim, Shahroz Tariq, and Simon~S Woo.
\newblock Cored: Generalizing fake media detection with continual
  representation using distillation.
\newblock In {\em ACMMM}, 2021.

\bibitem{kingma2018glow}
Durk~P Kingma and Prafulla Dhariwal.
\newblock Glow: Generative flow with invertible 1x1 convolutions.
\newblock {\em NeurIPS}, 2018.

\bibitem{kingma2013auto}
Diederik~P Kingma and Max Welling.
\newblock Auto-encoding variational bayes.
\newblock In {\em ICLR}, 2014.

\bibitem{kirkpatrick2017overcoming}
James Kirkpatrick, Razvan Pascanu, Neil Rabinowitz, Joel Veness, Guillaume
  Desjardins, Andrei~A Rusu, Kieran Milan, John Quan, Tiago Ramalho, Agnieszka
  Grabska-Barwinska, et~al.
\newblock Overcoming catastrophic forgetting in neural networks.
\newblock {\em Proceedings of the national academy of sciences},
  114(13):3521--3526, 2017.

\bibitem{korshunov2018deepfakes}
Pavel Korshunov and S{\'e}bastien Marcel.
\newblock Deepfakes: a new threat to face recognition? assessment and
  detection.
\newblock {\em arXiv preprint arXiv:1812.08685}, 2018.

\bibitem{lee2017overcoming}
Sang-Woo Lee, Jin-Hwa Kim, Jaehyun Jun, Jung-Woo Ha, and Byoung-Tak Zhang.
\newblock Overcoming catastrophic forgetting by incremental moment matching.
\newblock {\em NeurIPS}, 30, 2017.

\bibitem{li2022technical}
Duo Li, Guimei Cao, Yunlu Xu, Zhanzhan Cheng, and Yi Niu.
\newblock Technical report for iccv 2021 challenge sslad-track3b: Transformers
  are better continual learners.
\newblock {\em arXiv preprint arXiv:2201.04924}, 2022.

\bibitem{li2019diverse}
Ke Li, Tianhao Zhang, and Jitendra Malik.
\newblock Diverse image synthesis from semantic layouts via conditional imle.
\newblock In {\em ICCV}, 2019.

\bibitem{li2018ictu}
Yuezun Li, Ming-Ching Chang, and Siwei Lyu.
\newblock In ictu oculi: Exposing ai created fake videos by detecting eye
  blinking.
\newblock In {\em WIFS}, 2018.

\bibitem{li2018exposing}
Yuezun Li and Siwei Lyu.
\newblock Exposing deepfake videos by detecting face warping artifacts.
\newblock {\em arXiv preprint arXiv:1811.00656}, 2018.

\bibitem{li2019celeb}
Yuezun Li, Xin Yang, Pu Sun, Honggang Qi, and Siwei Lyu.
\newblock Celeb-df: A new dataset for deepfake forensics.
\newblock {\em onikle.com}, 2019.

\bibitem{li2020celeb}
Yuezun Li, Xin Yang, Pu Sun, Honggang Qi, and Siwei Lyu.
\newblock Celeb-df: A large-scale challenging dataset for deepfake forensics.
\newblock In {\em CVPR}, 2020.

\bibitem{li2017learning}
Zhizhong Li and Derek Hoiem.
\newblock Learning without forgetting.
\newblock {\em IEEE transactions on pattern analysis and machine intelligence},
  40(12):2935--2947, 2017.

\bibitem{lin2014microsoft}
Tsung-Yi Lin, Michael Maire, Serge Belongie, James Hays, Pietro Perona, Deva
  Ramanan, Piotr Doll{\'a}r, and C~Lawrence Zitnick.
\newblock Microsoft coco: Common objects in context.
\newblock In {\em ECCV}, 2014.

\bibitem{liu2020mnemonics}
Yaoyao Liu, Yuting Su, An-An Liu, Bernt Schiele, and Qianru Sun.
\newblock Mnemonics training: Multi-class incremental learning without
  forgetting.
\newblock In {\em CVPR}, 2020.

\bibitem{liu2015deep}
Ziwei Liu, Ping Luo, Xiaogang Wang, and Xiaoou Tang.
\newblock Deep learning face attributes in the wild.
\newblock In {\em ICCV}, 2015.

\bibitem{lopez2017gradient}
David Lopez-Paz and Marc'Aurelio Ranzato.
\newblock Gradient episodic memory for continual learning.
\newblock {\em NIPS}, 2017.

\bibitem{marra2019incremental}
Francesco Marra, Cristiano Saltori, Giulia Boato, and Luisa Verdoliva.
\newblock Incremental learning for the detection and classification of
  gan-generated images.
\newblock In {\em WIFS}, 2019.

\bibitem{matern2019exploiting}
Falko Matern, Christian Riess, and Marc Stamminger.
\newblock Exploiting visual artifacts to expose deepfakes and face
  manipulations.
\newblock In {\em WACV Workshops}, 2019.

\bibitem{mittal2021essentials}
Sudhanshu Mittal, Silvio Galesso, and Thomas Brox.
\newblock Essentials for class incremental learning.
\newblock In {\em CVPR}, 2021.

\bibitem{nguyen2019multi}
Huy~H Nguyen, Fuming Fang, Junichi Yamagishi, and Isao Echizen.
\newblock Multi-task learning for detecting and segmenting manipulated facial
  images and videos.
\newblock {\em arXiv preprint arXiv:1906.06876}, 2019.

\bibitem{nguyen2019use}
Huy~H Nguyen, Junichi Yamagishi, and Isao Echizen.
\newblock Use of a capsule network to detect fake images and videos.
\newblock {\em arXiv preprint arXiv:1910.12467}, 2019.

\bibitem{park2019SPADE}
Taesung Park, Ming-Yu Liu, Ting-Chun Wang, and Jun-Yan Zhu.
\newblock Semantic image synthesis with spatially-adaptive normalization.
\newblock In {\em CVPR}, 2019.

\bibitem{pellegrini2020latent}
Lorenzo Pellegrini, Gabriele Graffieti, Vincenzo Lomonaco, and Davide Maltoni.
\newblock Latent replay for real-time continual learning.
\newblock In {\em IROS}, 2020.

\bibitem{rebuffi2017icarl}
Sylvestre-Alvise Rebuffi, Alexander Kolesnikov, Georg Sperl, and Christoph~H
  Lampert.
\newblock icarl: Incremental classifier and representation learning.
\newblock In {\em CVPR}, 2017.

\bibitem{richter2016playing}
Stephan~R Richter, Vibhav Vineet, Stefan Roth, and Vladlen Koltun.
\newblock Playing for data: Ground truth from computer games.
\newblock In {\em ECCV}, 2016.

\bibitem{riemer2018learning}
Matthew Riemer, Ignacio Cases, Robert Ajemian, Miao Liu, Irina Rish, Yuhai Tu,
  and Gerald Tesauro.
\newblock Learning to learn without forgetting by maximizing transfer and
  minimizing interference.
\newblock {\em arXiv preprint arXiv:1810.11910}, 2018.

\bibitem{robins1995catastrophic}
Anthony Robins.
\newblock Catastrophic forgetting, rehearsal and pseudorehearsal.
\newblock {\em Connection Science}, 7(2):123--146, 1995.

\bibitem{rossler2019faceforensics++}
Andreas Rossler, Davide Cozzolino, Luisa Verdoliva, Christian Riess, Justus
  Thies, and Matthias Nie{\ss}ner.
\newblock Faceforensics++: Learning to detect manipulated facial images.
\newblock In {\em ICCV}, 2019.

\bibitem{saha2020gradient}
Gobinda Saha, Isha Garg, and Kaushik Roy.
\newblock Gradient projection memory for continual learning.
\newblock In {\em ICLR}, 2020.

\bibitem{Sanderson2009lncs}
C. Sanderson and B.C. Lovell.
\newblock Multi-region probabilistic histograms for robust and scalable
  identity inference.
\newblock {\em Lecture Notes in Computer Science}, 2009.

\bibitem{shahbazi2021efficient}
Mohamad Shahbazi, Zhiwu Huang, Danda~Pani Paudel, Ajad Chhatkuli, and Luc
  Van~Gool.
\newblock Efficient conditional gan transfer with knowledge propagation across
  classes.
\newblock In {\em CVPR}, 2021.

\bibitem{shin2017continual}
Hanul Shin, Jung~Kwon Lee, Jaehong Kim, and Jiwon Kim.
\newblock Continual learning with deep generative replay.
\newblock {\em arXiv preprint arXiv:1705.08690}, 2017.

\bibitem{szegedy2016rethinking}
Christian Szegedy, Vincent Vanhoucke, Sergey Ioffe, Jon Shlens, and Zbigniew
  Wojna.
\newblock Rethinking the inception architecture for computer vision.
\newblock In {\em CVPR}, 2016.

\bibitem{tang2021layerwise}
Shixiang Tang, Dapeng Chen, Jinguo Zhu, Shijie Yu, and Wanli Ouyang.
\newblock Layerwise optimization by gradient decomposition for continual
  learning.
\newblock In {\em CVPR}, 2021.

\bibitem{tao2020topology}
Xiaoyu Tao, Xinyuan Chang, Xiaopeng Hong, Xing Wei, and Yihong Gong.
\newblock Topology-preserving class-incremental learning.
\newblock In {\em ECCV}, 2020.

\bibitem{thies2019deferred}
Justus Thies, Michael Zollh{\"o}fer, and Matthias Nie{\ss}ner.
\newblock Deferred neural rendering: Image synthesis using neural textures.
\newblock {\em ACM Transactions on Graphics (TOG)}, 38(4):1--12, 2019.

\bibitem{thies2016face2face}
Justus Thies, Michael Zollhofer, Marc Stamminger, Christian Theobalt, and
  Matthias Nie{\ss}ner.
\newblock Face2face: Real-time face capture and reenactment of rgb videos.
\newblock In {\em CVPR}, 2016.

\bibitem{van2021class}
Gido~M van~de Ven, Zhe Li, and Andreas~S Tolias.
\newblock Class-incremental learning with generative classifiers.
\newblock In {\em CVPR}, 2021.

\bibitem{wang2021training}
Shipeng Wang, Xiaorong Li, Jian Sun, and Zongben Xu.
\newblock Training networks in null space of feature covariance for continual
  learning.
\newblock In {\em CVPR}, 2021.

\bibitem{wang2020cnn}
Sheng-Yu Wang, Oliver Wang, Richard Zhang, Andrew Owens, and Alexei~A Efros.
\newblock {CNN}-generated images are surprisingly easy to spot... for now.
\newblock In {\em CVPR}, 2020.

\bibitem{wang2022deepfake}
Xueyu Wang, Jiajun Huang, Siqi Ma, Surya Nepal, and Chang Xu.
\newblock Deepfake disrupter: The detector of deepfake is my friend.
\newblock In {\em CVPR}, 2022.

\bibitem{wang2021learning}
Zifeng Wang, Zizhao Zhang, Chen-Yu Lee, Han Zhang, Ruoxi Sun, Xiaoqi Ren,
  Guolong Su, Vincent Perot, Jennifer Dy, and Tomas Pfister.
\newblock Learning to prompt for continual learning.
\newblock In {\em CVPR}, 2022.

\bibitem{wu2019large}
Yue Wu, Yinpeng Chen, Lijuan Wang, Yuancheng Ye, Zicheng Liu, Yandong Guo, and
  Yun Fu.
\newblock Large scale incremental learning.
\newblock In {\em CVPR}, 2019.

\bibitem{yang2019exposing}
Xin Yang, Yuezun Li, and Siwei Lyu.
\newblock Exposing deep fakes using inconsistent head poses.
\newblock In {\em ICASSP}, 2019.

\bibitem{yoon2017lifelong}
Jaehong Yoon, Eunho Yang, Jeongtae Lee, and Sung~Ju Hwang.
\newblock Lifelong learning with dynamically expandable networks.
\newblock {\em arXiv preprint arXiv:1708.01547}, 2017.

\bibitem{yu2015lsun}
Fisher Yu, Ari Seff, Yinda Zhang, Shuran Song, Thomas Funkhouser, and Jianxiong
  Xiao.
\newblock Lsun: Construction of a large-scale image dataset using deep learning
  with humans in the loop.
\newblock {\em arXiv preprint arXiv:1506.03365}, 2015.

\bibitem{yu2020semantic}
Lu Yu, Bartlomiej Twardowski, Xialei Liu, Luis Herranz, Kai Wang, Yongmei
  Cheng, Shangling Jui, and Joost van~de Weijer.
\newblock Semantic drift compensation for class-incremental learning.
\newblock In {\em CVPR}, 2020.

\bibitem{yu2021improving}
Pei Yu, Yinpeng Chen, Ying Jin, and Zicheng Liu.
\newblock Improving vision transformers for incremental learning.
\newblock {\em arXiv preprint arXiv:2112.06103}, 2021.

\bibitem{zeng2019continual}
Guanxiong Zeng, Yang Chen, Bo Cui, and Shan Yu.
\newblock Continual learning of context-dependent processing in neural
  networks.
\newblock {\em Nature Machine Intelligence}, 1(8):364--372, 2019.

\bibitem{zenke2017continual}
Friedemann Zenke, Ben Poole, and Surya Ganguli.
\newblock Continual learning through synaptic intelligence.
\newblock In {\em ICML}, 2017.

\bibitem{zhou2017two}
Peng Zhou, Xintong Han, Vlad~I Morariu, and Larry~S Davis.
\newblock Two-stream neural networks for tampered face detection.
\newblock In {\em CVPR Workshops}, 2017.

\bibitem{zhu2017unpaired}
Jun-Yan Zhu, Taesung Park, Phillip Isola, and Alexei~A Efros.
\newblock Unpaired image-to-image translation using cycle-consistent
  adversarial networks.
\newblock In {\em ICCV}, 2017.

\bibitem{zi2020wilddeepfake}
Bojia Zi, Minghao Chang, Jingjing Chen, Xingjun Ma, and Yu-Gang Jiang.
\newblock Wilddeepfake: A challenging real-world dataset for deepfake
  detection.
\newblock In {\em ACMMM}, 2020.

\end{thebibliography}
